\documentclass[11pt, a4paper, logo, copyright, numbering]{googledeepmind}

\usepackage[authoryear, sort&compress, round]{natbib}
\usepackage{xcolor}
\usepackage{appendix}
\usepackage{subcaption}  \usepackage{mwe}  \usepackage{graphicx}
\usepackage{xr}
\usepackage{marvosym}

\title{Neural Compression of Atmospheric States}

\correspondingauthor{Piotr Mirowski piotrmirowski@deepmind.com, Shakir Mohamed shakir@deepmind.com}


\reportnumber{} 

\renewcommand{\today}{}

\author[1,*]{Piotr Mirowski}
\author[1,*]{David Warde-Farley}
\author[1]{Mihaela Rosca}
\author[1]{Matthew Koichi Grimes}
\author[1]{Yana Hasson}
\author[1]{Hyunjik Kim}
\author[1]{Mélanie Rey}
\author[1]{Simon Osindero}
\author[1,\Cross]{Suman Ravuri}
\author[1,\Cross]{Shakir Mohamed}

\affil[*]{Equal contributions}
\affil[1]{Google DeepMind}
\affil[\Cross]{Senior authors}

\begin{abstract}
Atmospheric states derived from reanalysis comprise a substantial portion of weather and climate simulation outputs. Many stakeholders --- such as researchers, policy makers, and insurers --- use this data to better understand the earth system and guide policy decisions. Atmospheric states have also received increased interest as machine learning approaches to weather prediction have shown promising results. A key issue for all audiences is that dense time series of these high-dimensional states comprise an enormous amount of data, precluding all but the most well resourced groups from accessing and using historical data and future projections. To address this problem, we propose a method for compressing atmospheric states using methods from the neural network literature, adapting spherical data to processing by conventional neural architectures through the use of the area-preserving HEALPix projection.
We investigate two model classes for building neural compressors: the hyperprior model from the neural image compression literature and recent vector-quantised models.
We show that both families of models satisfy the desiderata of small average error, a small number of high-error reconstructed pixels, faithful reproduction of extreme events such as hurricanes and heatwaves, preservation of the spectral power distribution across spatial scales.
We demonstrate compression ratios in excess of $1000\times$, with compression and decompression at a rate of approximately one second per global atmospheric state.

 \end{abstract}

\begin{document}

\maketitle

\section{Introduction}
\label{sx:introduction}

This paper presents a family of neural network compression methods of simulated atmospheric states, with the aim of reducing the currently immense storage requirements of such data from cloud scale (petabytes) to desktop scale (terabytes). 
This need for compression has come about over past 50~years, characterized by a steady push to increase the resolution of atmospheric simulations, increasing the size and storage demands of the resulting datasets (e.g., \citet{neumann2019assessing}, \citet{schneider2023harnessing}, \citet{essd-16-2113-2024}), while atmospheric simulation has come to play an increasingly critical role in scientific, industrial and policy-level pursuits.
Higher spatial resolutions unlock the ability of simulators to deliver more accurate predictions and resolve ever more atmospheric phenomena.
For example, while current models often operate at 25 - 50~km resolution, resolving storms requires 1~km resolution~\citep{stevens2020added}, while resolving the motion of (and radiative effects due to) low clouds require 100~m resolution~\citep{schneider2017earth,satoh2019global}.
Machine learning models for weather prediction also face opportunities and challenges with higher resolution: while additional granularity may afford better modeling opportunities, even the present size of atmospheric states poses a significant bottleneck for loading training data and serving model outputs~\citep{chantry2021opportunities}.

To put the data storage problem in perspective, storing 40 years of reanalysis data from the ECMWF Reanalysis v5 dataset (ERA5, \citet{hersbach2020era5}) at full spatial and temporal resolution (i.e.\ without subsampling)
requires 181~TB of storage per atmospheric variable.
For 6 such variables, a single trajectory exceeds 1~PB, to say nothing of an ensemble of tens or hundreds of trajectories as required for model intercomparison or for predicting distributions of outcomes~\citep{eyring2016overview}. As \citet{sz3_framework} write, ``recent climate research generates 260 TB of data every 16s''. Such conditions create considerable difficulties, prompting researchers to discard data or decrease temporal and/or spatial resolution. 
Having to decimate the dataset to satisfy financial and engineering constraints is especially unfortunate for machine learning models, whose performance is often bounded by the size of the training dataset,
while also making fair and systematic comparisons to results from traditional numerical weather prediction (NWP) more challenging.
Perhaps most importantly, the size of the data precludes access to the full dataset for all but the most well-resourced groups. Priced at the time of writing, cloud storage for 100 PB costs on the order of \$1M per month\footnote{This value was obtained by looking up the publicly available rate for cloud storage fees on 28 June 2024 from two of the large cloud storage providers. Costs depend on data replication and access requirements.}.

Data compression is a well-studied problem with many off-the-shelf algorithms available. Unfortunately, for continuous-valued data such as atmospheric states, modern lossless compression methods tailored to such data offer only modest savings, reducing the size by only a factor of 2  (c.f. Figure \ref{fig:reprojection_error_vs_cr_bounded_compression}). Existing lossy compression algorithms pose their own problems, as they are usually designed to compress while optimizing mean squared error~(MSE)~\citep{sz3_algo}.
In the weather and climate domain, the MSE loss in particular can attenuate high spatial frequencies and thereby remove physically important spatial discontinuities~\citep{ravuri2021nowcasting}. Na\"ive pursuit of reduced MSE can result in significant, scientifically problematic distortions, including outright erasure of entire hurricanes as noted in \cite{huang2022compressing}. This motivates the development of bespoke compression methods evaluated with an eye towards not just MSE but also other physical metrics of scientific interest, such as power spectrum distortion and the preservation of extrema.

\begin{figure}
\begin{flushright}
    \includegraphics[width=\textwidth, trim = 6cm 2.5cm 5cm 4.5cm, clip]{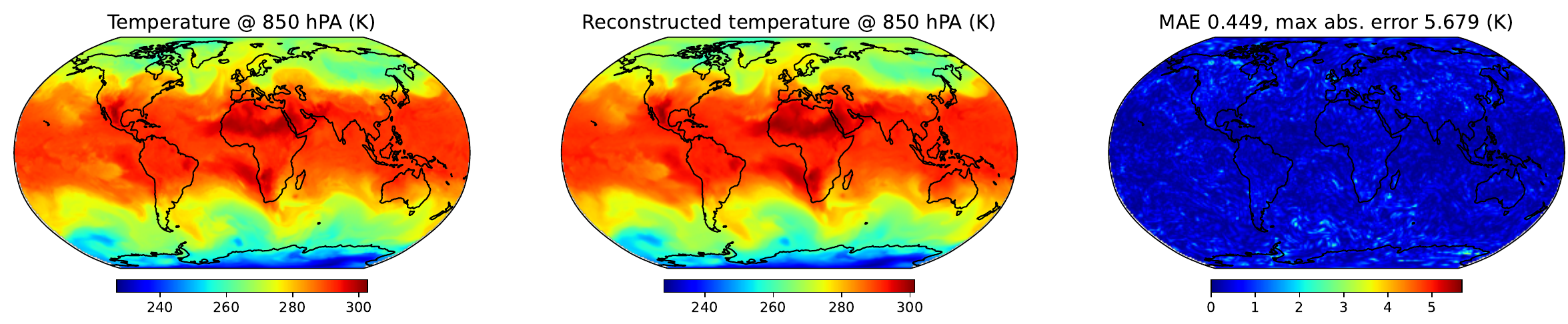}
    \includegraphics[width=\textwidth, trim = 6cm 2.5cm 5cm 4.5cm, clip]{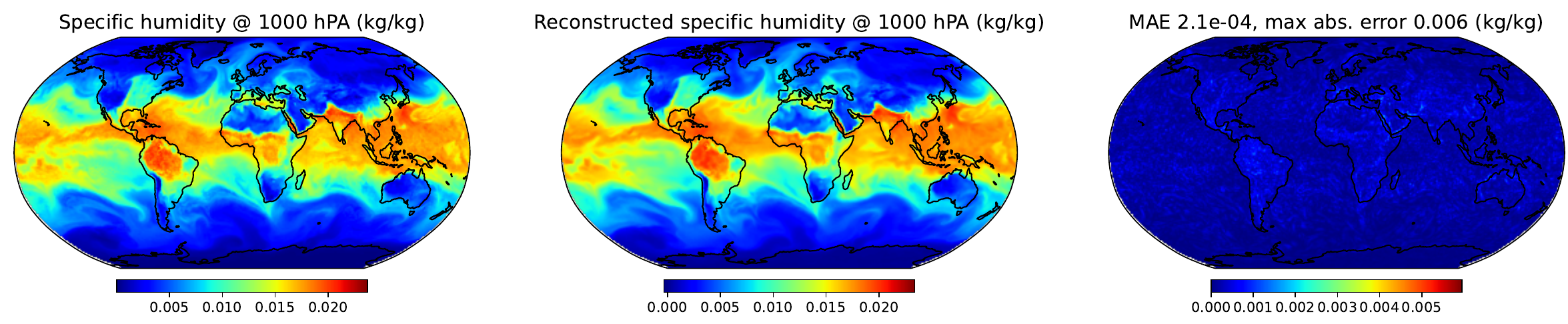}
    \includegraphics[width=\textwidth, trim = 6cm 2.5cm 5cm 4.5cm, clip]{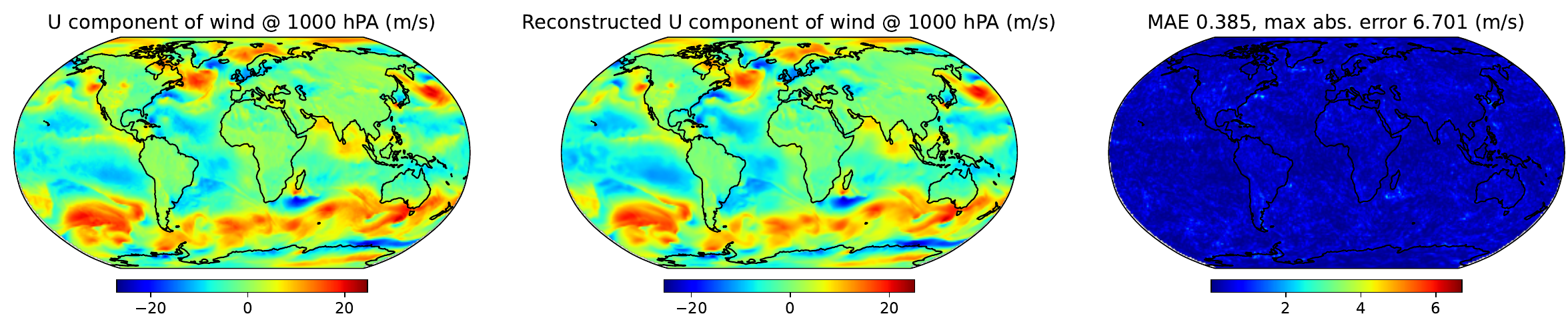}
    \includegraphics[width=\textwidth, trim = 6cm 3.8cm 5cm 2.8cm, clip]{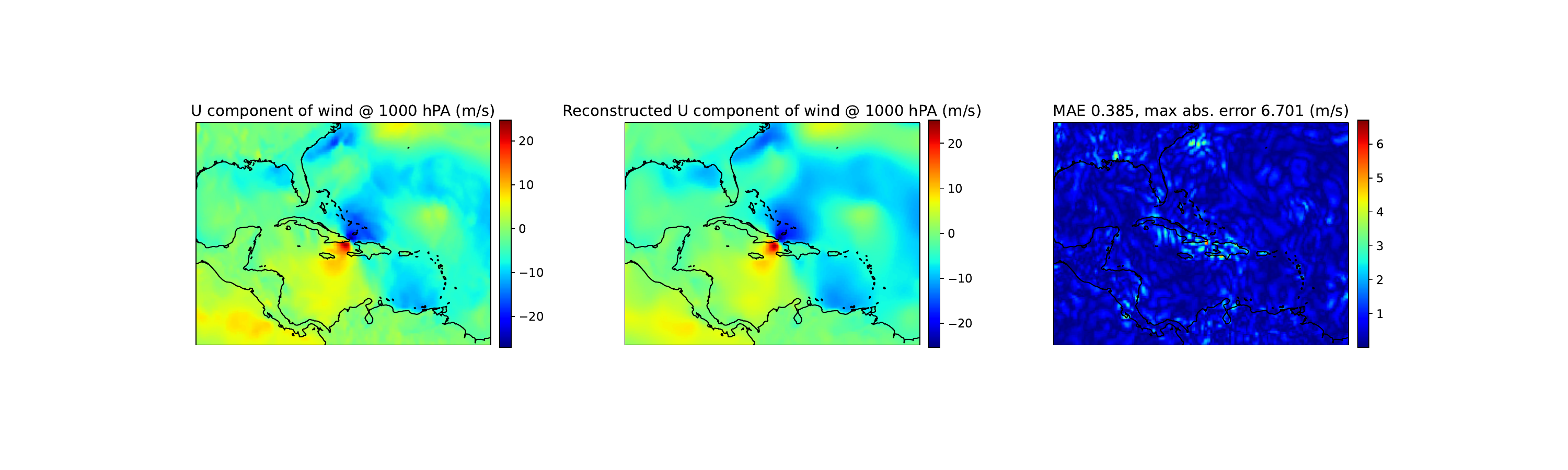}
\includegraphics[width=\textwidth, trim = 6cm 3.8cm 5cm 2.8cm, clip]{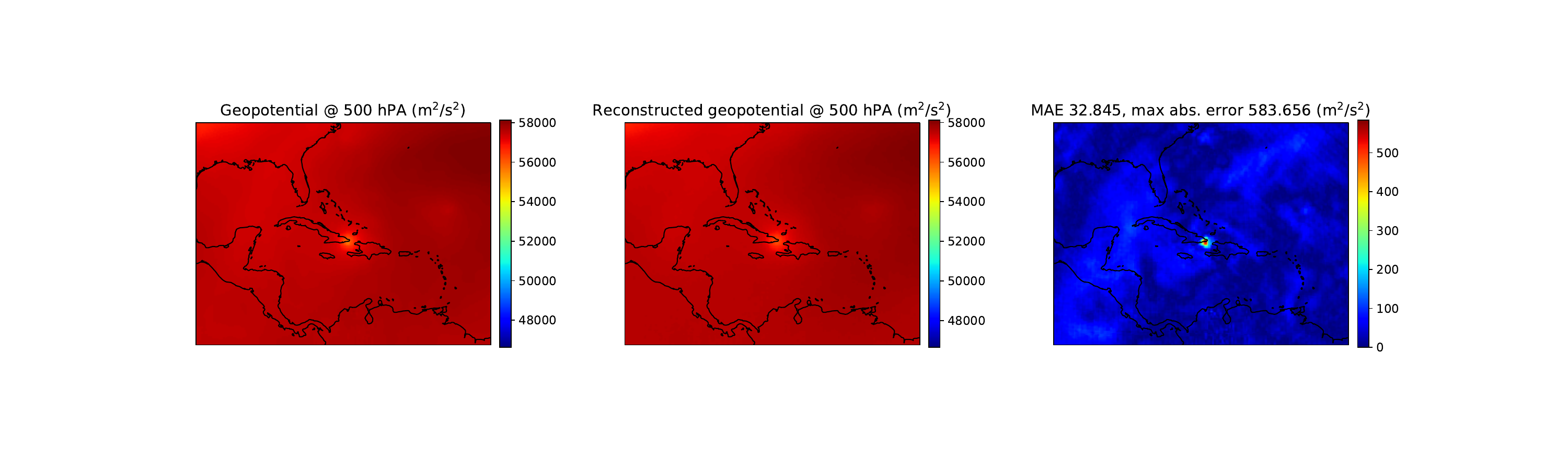}
    \includegraphics[width=0.65\textwidth, trim = 20cm 3.8cm 5cm 2.8cm, clip]{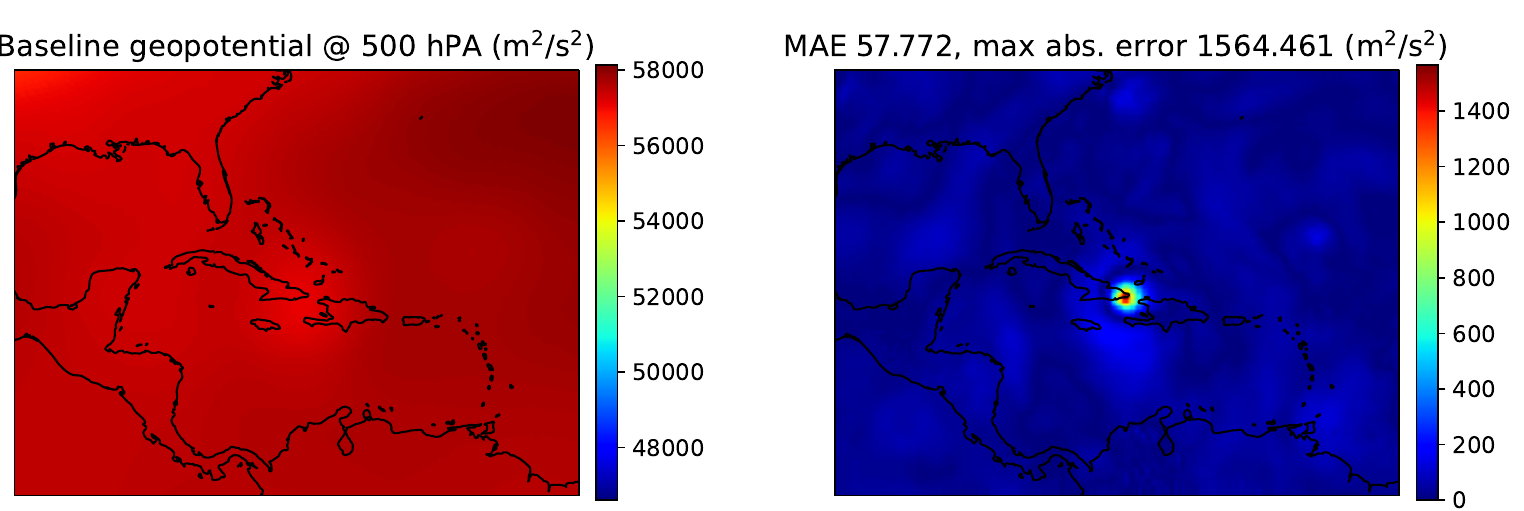}
\caption{Hyperprior reconstructions of a global frame on 2016/10/5 at 0 UTC, compressed $1000\times$. From top to bottom: temperature at 850 hPa, specific humidity at 1000 hPa, the zonal component of wind at 1000 hPa (and a zoom over the Caribbean), and geopotential at 500 hPa over the Caribbean. Last row shows reconstruction using \citet{huang2022compressing} compressed $1150\times$ as a baseline. Columns show the ground truth, the reconstruction and the residual along mean absolute error (MAE). Note the improvement in Hurricane Matthew's reconstruction over the baseline.}
    \label{fig:hurricane_matthew_hyperprior}
    \end{flushright}
\end{figure}

Based on the scientific, financial and social needs described above, suitable candidate compressors will be those capable of high compression rates, low error (both on average and for extreme values) and fast execution; adequately meeting these needs will enable researchers to use atmospheric datasets in full without need of subsampling, increase the speed of data dissemination at all levels, and reduce storage costs.
In pursuit of this goal, we present a family of compression methods based on autoencoder neural networks that are adapted and trained for atmospheric data, and examine the inherent tradeoffs within this design space.
These compression methods illustrate and exploit the informational redundancy of atmospheric states.
In Section \ref{sx:learning2compress} we discuss existing compression methods and insight available in this area, and describe our own technical groundwork.
Within the family of methods we will describe, we will put forward a system based on the hyperprior model~\citep{balle2018hp} as the candidate neural compressor which most fully satisfies the requirements of atmospheric state compression.
As a case study, we showcase in Figure \ref{fig:hurricane_matthew_hyperprior} the compression results for hurricane Matthew, which shows low errors across physical variables of interest, as well as the ability to preserve extreme values and events.  

\begin{figure}
    \centering
    \includegraphics[width=\textwidth]{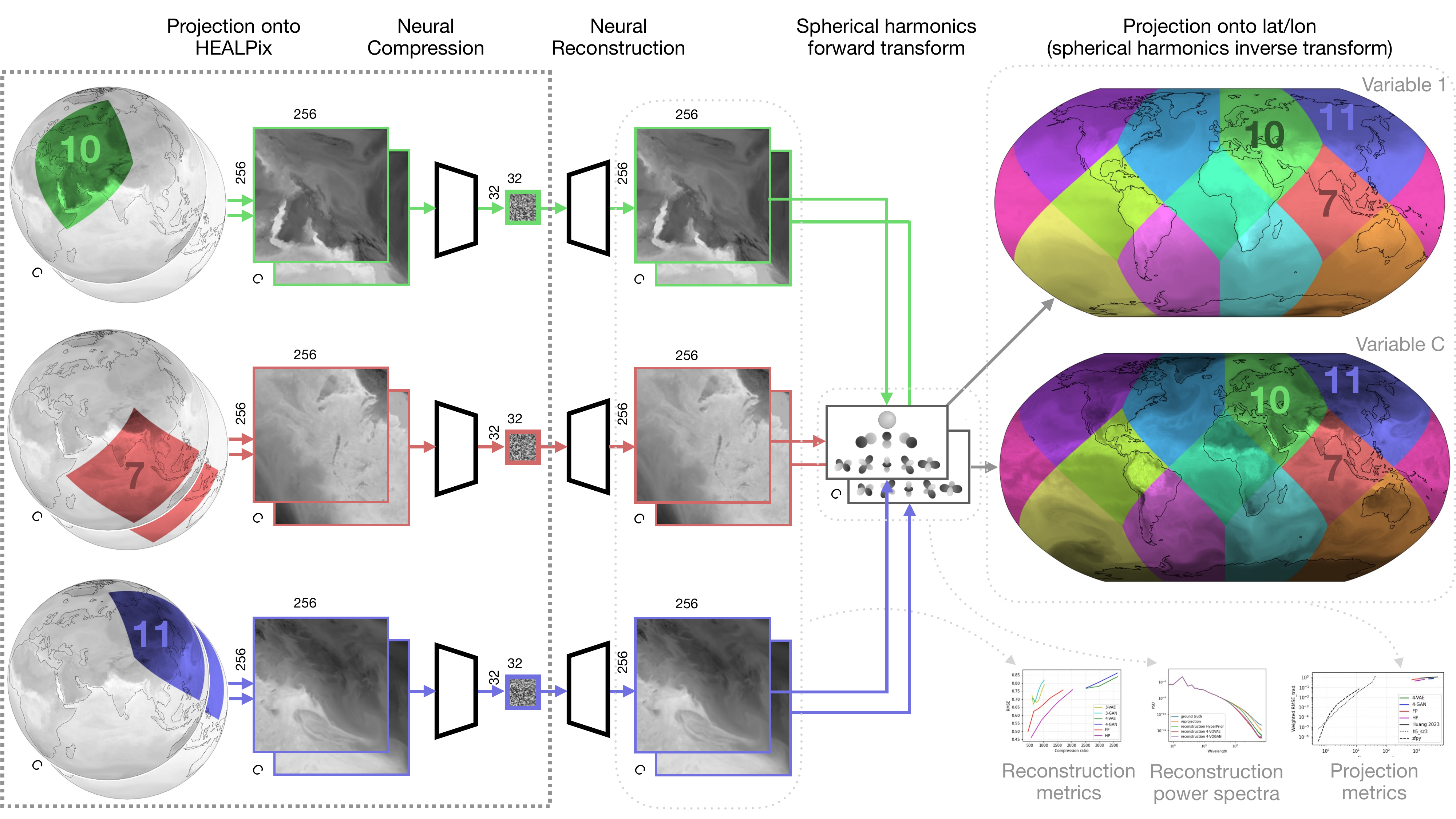}
    \hfill
    \caption{Overview of our neural data compression system. Global surface and atmospheric data are projected onto the 12 HEALPix square base pixels. On the left, the figure illustrates 3 such squares, numbered 7 (in red, over the Indian Ocean), 10 (in green, over continental Europe and Arabic peninsula) and 11 (in blue, over East Asia), for $C$ variables (which may include different physical variables at one or multiple pressure levels), collected from a single atmospheric state at a given timestamp. We jointly compress $C$ separate channels for every $256 \times 256$ square into a quantized representation (here, we illustrate a $32 \times 32$ map of discrete VQ indices corresponding to a 4-downsampling-block VQ-VAE or VQ-GAN). During training and evaluation, we reconstruct the HEALPix squares back from compressed, quantized representations. After reconstruction, spherical harmonics coefficients are calculated via forward transform of the HEALPix reconstructions. These coefficients are used to synthesize re-projected images via inverse transform in latitude/longitude coordinates. On the right, the figure illustrates two re-projected reconstructions for variables 1 and $C$, with areas corresponding to different HEALPix squares coded in different colours. Metrics are computed during reconstruction (comparison of reconstruction vs. ground truth in HEALPix projection), after spherical harmonics forward transform (power spectrum) and after re-projection back onto latitude/longitude representations (comparison of re-projected reconstruction vs. ground truth in latitude/longitude coordinates).}
    \label{fig:system}
\end{figure}

Our overall approach comprises four stages, depicted in Figure \ref{fig:system}: reprojecting the atmospheric state to a square format better suited to ML accelerators using the HEALPix projection~\citep{gorski1999healpix}, employing a neural encoder to map these projections into a discrete representation (which can be losslessly compressed with standard techniques), reconstructing HEALPix projections using a neural decoder, and finally reprojecting back onto the equirectangular latitude/longitude grid using the spherical harmonics transform.

We will show in Section~\ref{sx:experiments} that our approach is able to compress these data at low error by approximately three orders of magnitude, and thus demonstrate bespoke high-compression methods applicable to multi-decadal data.
We obtain a mean absolute error (MAE) of approximately $0.4^\circ$~K for temperature, approximately $0.5~\text{m}/\text{s}$ for zonal and meridional wind, below 1 hPa for surface pressure and approximately $40~\text{m}^2/\text{s}^2$ for geopotential, with less than 0.5\% of HEALPix pixels beyond an error of $1^\circ$~K (temperature), $1~\text{m}/\text{s}$ (zonal and meridional wind) or less than 0.05\% of pixels beyond an error of $100~\text{m}^2/\text{s}^2$ (geopotential) while preserving spectral shape, and approximately one second encoding/decoding time per global state. 
We show that high-error pixels are rare enough that their values can be stored in a lookup table while keeping the overall compression ratio above 1000.
Because sharp features and rare events can be the first victim of aggressive compression, we conduct an analysis of weather events of interest such as hurricanes and heat waves to demonstrate that these are not distorted.
We provide a discussion of the different variations in use and further contextualisation of the results in Section \ref{sx:discussion}, and discuss limitations and conclude in Section \ref{sx:conclusion}.

\section{Learning to Compress}
\label{sx:learning2compress}

\subsection{Lossless and Lossy Compression}
\label{sx:methods:lossless-lossy}
Compression methods can be divided into \emph{lossless} and \emph{lossy} categories.
Lossless compression aims to transform data in such a way that the original can be reinstantiated exactly.
As digital information is ultimately discrete in nature, lossless compression methods typically proceed from an assumption that their input is a sequence of discrete tokens.
While continuous-valued datasets are still discrete in the sense of being encoded as finite, binary strings, general purpose lossless compression methods often perform poorly due to the lack of readily discoverable statistical structure in this binary space.
Empirically,~\citet{sdrbench_2020} find compression ratios (the ratio of the original data size to the compressed size) of less than 3 for a range of lossless compression algorithms~\citep{zstd, zfpy_2014, fpc_2009, blosc, fpzip_2006} applied to continuous-valued scientific datasets. For many research budgets, compressing a 100 PB atmospheric trajectory ensemble to 33 PB would not meaningfully address data storage challenges.

For the output physical simulations,~\cite{simfs_girolamo_2019} propose to store trajectories as a sparse set of checkpoint frames spaced at regular time intervals, and re-simulate the discarded frames from the nearest preceding checkpoint. The compression ratio is dependent on the frame spacing, and re-simulation may incur significant computational cost. The methods we describe do not assume the existence of, or access to, the original data simulator.

Lossy compression methods, by contrast, allow for some distortion in the recovered signal after decompression.
There exists an inherent tension between the amount of compression (the obtained code's \emph{rate}) and the degree of distortion incurred, the \emph{rate-distortion trade-off}.
The kind and degree of distortion acceptable depends heavily on the application domain, and the quantification and mitigation of application-relevant distortion is central to the design of compression methods.
Domain knowledge also allows the algorithm designer to exploit known statistical regularities in the data to achieve higher compression at a fixed distortion level.

Several error-bounded lossy compression algorithms have been designed for floating-point arrays \citep{sz3_algo,sz3_framework,zfpy_2014,tthresh_2020,sperr_2023}. Among those, we evaluate SZ3~\citep{sz3_algo,sz3_framework} and ZFP~\citep{zfpy_2014} and show in Section \ref{sx:discuss:error-bound} that these methods only achieve approximately $50\times$ compression.
\citet{klower2021compressing}~proposes using lower-precision numbers to represent atmospheric states, effectively discretizing floating-point numbers by rounding them. This achieves compression ratios of up to 155 for temperature, at $\pm 0.4^\circ$~C median error. Neural network-based approaches for compression, that we describe further in the next section, have also recently been introduced by \citet{huang2022compressing} and \citet{han2024cra5}.

Our overall approach is best understood within the paradigm of \emph{transform coding}~\citep{goyal2001theoretical}, which underpins most modern methods for the compression of natural signals such as images, audio and video.
In transform coding, the original signal is invertibly mapped into an alternative domain judged more suitable for lossy scalar quantization (for example, certain elements of the transformed representation are known to be less crucial to perceived quality for the reconstruction~\citep{wallace1992jpeg}), and a stream of bits representing indices in the reduced set of quantized scalars is losslessly compressed via entropy coding.
The invertible transform is typically linear (e.g., the familiar JPEG format uses a discrete cosine transform on small square sub-regions), and designed to produce a collection of relatively uncorrelated scalars.
Traditionally, the transform, quantization and entropy coding steps form wholly separate subsystems of a compression pipeline.
While this modularity affords implementation benefits, recent advances in machine learning have demonstrated that quantization can be directly and fruitfully incorporated into scalable representation learning systems.
In this work, we explore two such families of methods, the latter of which also employs a relaxation of the entropy coding objective to directly parameterize the rate-distortion trade-off.

\subsection{Neural Compression}
\label{sx:methods:networks}

The methods explored in this work build upon the \emph{autoencoder} family~\citep{ackley1985learning,hinton1990connectionist} wherein a parameterized non-linear mapping (i.e.\ neural network, best thought of as the composition of an ``encoder'' and a ``decoder'' function) is fit to data, with the goal of the system reproducing its own input from an alternative representation; Figure \ref{fig:system} depicts the overall setup.
Given that the identity function is otherwise trivially representable, constraints imposed on the functional form of an autoencoder (such as a low-dimensional ``bottleneck'' between encoder and decoder, or penalties imposed during optimization) guide the learning procedure in the direction of descriptive parsimony~\citep{elman1988learning,Cottrell1987}.
Probabilistically motivated variants of autoencoders have seen wide use in learning low-dimensional representations of data, and have an established connection to compression theory, formalized in several ways through the minimum description length principle and bits-back-encoding~\citep{hinton1993autoencoders, grunwald2007minimum}, variational inference~\citep{rezende2014reparameterizationtrick,kingma2014vae}.

In contrast to autoencoders, a recent line of work uses neural networks to represent continuous data directly, mapping spatial coordinates to their corresponding data values~\citep{dupont2022functa, xie2022neural_fields_beyond}. For example, the pixels of a single photograph can be used as a ``dataset'' from which to train a network to map pixel coordinates $(i, j)$ to the corresponding pixel color $(r, g, b)$. Once trained, the network's parameters serve as the compressed representation, and can be significantly more compact than the raw array of data they encode.
\citet{huang2022compressing} demonstrate this on atmospheric data on a wide range of compression ratios, but this approach has the disadvantage that it requires training or fine-tuning the model for each new data point (although we note that there are now methods that use meta-learning to address this limitation \citep{coinplusplus_2022}).

In an autoencoder, the encoder maps rich high-dimensional signals to lower-dimensional codes more suitable for storage or transmission, which can be lossily reconstructed by applying the decoder.
The encoder-decoder framework has several advantages. Once trained, the encoder can be applied to data beyond its training set. The compressed representation can be designed to have the same spatial layout and topology as the input data, aiding interpretability. Of particular interest are strategies for learning \emph{discrete} representations with autoencoders, which lossless compression, and may facilitate direct, downstream use with sequence models such as transformers~\citep{vaswani2017transformer, nguyen2023climax}.
We next describe several families of such autoencoders which achieve discrete representations using two different strategies.

\subsubsection{VQ-VAE and VQ-GAN}

The vector-quantized variational autoencoder, or VQ-VAE~\citep{oord2017vqvae} is a neural network that maps input data to a set of discrete indices into a learned codebook of vector-valued codes.
Taking inspiration from \cite{kingma2014vae}, a continuous neural encoder maps the input $x$ to a continuous code layer. While the original VAE relied on amortized inference (by the encoder) of sufficient statistics of continuous random variables and a ``reparameterization trick'' to obtain differentiable samples~\citep{williams1992simple,rezende2014reparameterizationtrick}, the VQ-VAE defines a degenerate variational posterior in which (components of) the latent representation adopt the closest (in Euclidean distance) of $N$ distinct values from a learned codebook~$D$.
In the case of images, convolutional architectures~\citep{lecun1989generalization} are employed for the encoder and decoder, and a vector $z_{ij}$ at each spatial location in the lower dimensional, final layer spatial map is quantized to values from a codebook shared across all spatial locations. The encoder and decoder are trained end-to-end by minimizing the L2 norm between the input data and their reconstructions. During this optimization, the gradients through the non-differentiable quantizing operation are estimated by simply ignoring it and substituting the identity function~\citep{hinton2012coursera,bengio2013estimating,theis2017lossy}.
Additional loss functions steer the codebook entries toward the output of the encoder while also encouraging the encoder to ``pick'' between codebook entries rather than predict convex combinations thereof (see \citet{oord2017vqvae} for details).

\begin{figure}
    \centering
    \includegraphics[width=0.7\textwidth]{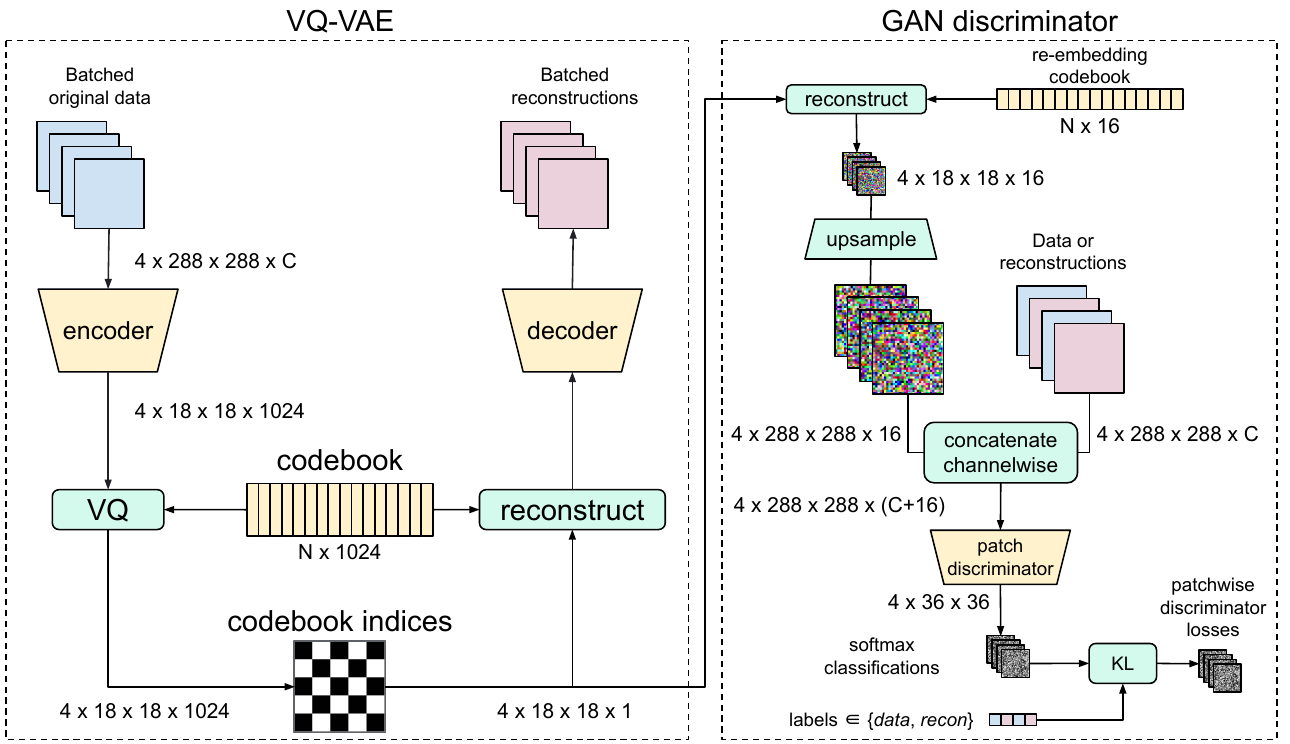}
    \caption{The VQ-GAN network, which contains a VQ-VAE network and an auxiliary discriminator. The encoder and decoder are residual convnets with attention, which reduce the input data into an 18x18 grid of vectors of size 1024. These are then quantized to their nearest match in the learned codebook of N vectors. See Appendix~\ref{sx:appendix:models} for details.}
    \label{fig:vqvae}
\end{figure}

Once quantized, an encoded value can be parsimoniously represented as an integer index $i \in [1, N]$ such that $z \approx D_i$.
Beyond the compression that results from replacing a multi-channel image with a much smaller array of integers, the resulting map can be further compressed with standard entropy-coding techniques, potentially exploiting both non-uniformity of codebook usage and spatial correlations in the resulting low-dimensional maps. Figure~\ref{fig:vqvae} illustrates the VQ-VAE network, depicting the encoder/decoder and the codebook reconstruction.

As noted in section~\ref{sx:introduction}, na\"ively optimizing the mean squared error can have unacceptable consequences for atmospheric states, as its tendency to wash out sharp features can erase the very events climate scientists wish to measure, such as hurricanes.
We therefore also investigate VQ-GANs~\citep{esser2021taming}, which augment a VQ-VAE by adding a patch-wise discriminator~\citep{isola2017image} and are trained with an auxiliary adversarial loss term~\citep{goodfellow2014gan}. During training, the discriminator receives the quantized feature map indices, channel-wise concatenated (after upsampling and continuous embedding) with either the original atmospheric data from which it was computed or its corresponding reconstruction. The discriminator is trained to maximize its classification accuracy on discriminating patches of real data from reconstruction patches, while the encoder-decoder pipeline is trained to reconstruct its input well while minimizing this accuracy.
Adversarial losses were similarly used for restoring high spatial frequencies for precipitation nowcasting in~\cite{ravuri2021nowcasting}.
For both VQ-VAE and VQ-GAN experiments, we base our neural network architectures on those presented in \cite{esser2021taming}.

\subsubsection{Factorized prior and Hyperprior Models}

Whereas VQ models discretize a latent vector $z$ to its nearest neighbor in a learned codebook, the factorized prior model~\citep{balle2017fp} discretizes individual elements of $z$ by rounding to integers at test time.
This is approximated during training as a zero-mean scalar noising function in the same spirit as the ``reparameterization trick'' described above.
Unlike the VQ models described earlier, the factorized prior's objective function is explicitly derived with data compression in mind, and a continuous relaxation of the Shannon entropy of the resulting discrete code comprises an additional loss term.
Probabilistically speaking, the encoder-decoder pair are trained to optimize a variational lower bound on the expected log-likelihood of the data modeled by latent variables with a prior consisting of independent uniform random variables (hence, ``factorized'') and an isotropic Gaussian likelihood.

\begin{figure}
    \centering
    \includegraphics[width=0.7\textwidth]{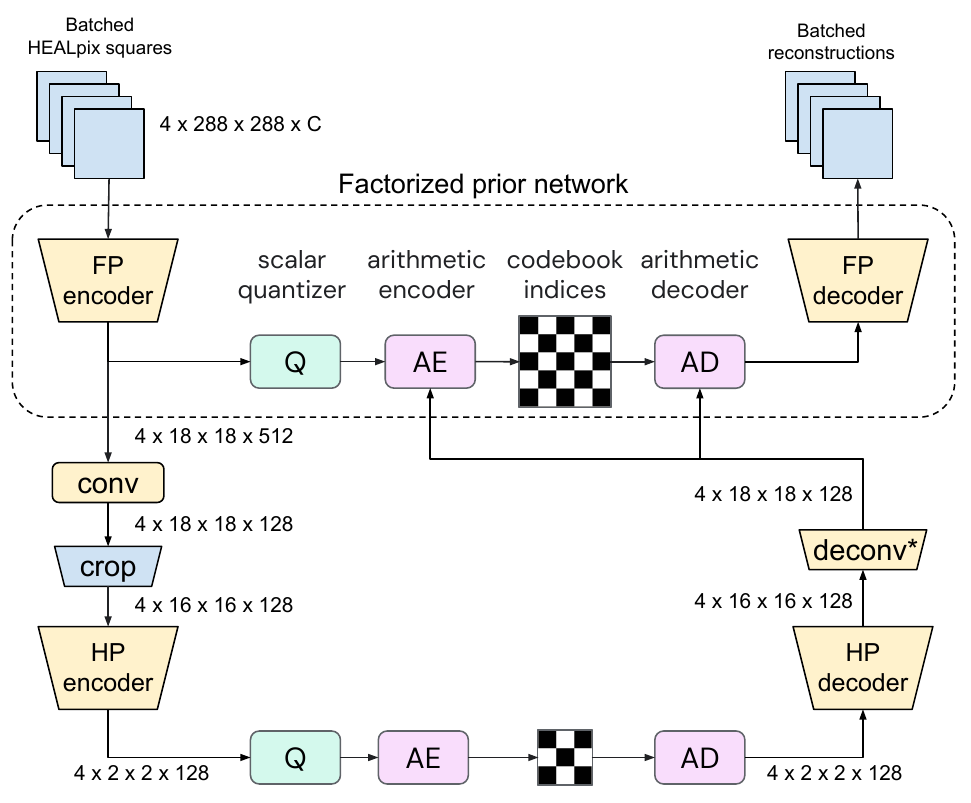}
    \caption{Hyperprior network. This augments the factorized prior network (dashed rectangle) by adding an additional encoder-decoder pathway that encodes the FP network's latents, and decodes them to a grid of variances.}
    \label{fig:hyperprior}
\end{figure}

The hyperprior model~\citep{balle2018hp} is an extension which, for training purposes, composes two factorized prior autoencoders.
The first block's encoder produces a feature map~$z$, which is quantized to integers~$\hat z$.
As is common in convolutional feature maps, the pixels of~$\hat{z}$ retain local spatial correlations.
The second autoencoder implements, with the same variational approach (though a different likelihood function) a probabilistic generative model which captures these correlations, introducing a new, smaller set of latent variables $w$\footnote{\citet{balle2018hp} uses the notation $y$ and $z$, rather than $z$ and $w$, for first and second level latent variables, respectively.}.
The probabilistic model defined by this secondary autoencoder is used to more effectively entropy code the first model's quantized maps $\hat{z}$ by predicting the scale (i.e.\ modeling the variance) of each element of $\hat{z}$ conditioned on the encoded, quantized secondary latent variables $\hat{w}$, which are themselves entropy-coded and stored alongside the  resulting encodings of $\hat{z}$.
Figure~\ref{fig:hyperprior} shows the network schematic for the hyperprior model, while Figure~\ref{fig:balle} gives a detailed picture including layer sizes.
We recommend the hyperprior model as the most suitable neural compressor that performs well across all the key requirements considered in the introduction.

\subsubsection{Effective compression ratio}

The na\"ive compression ratio for the VQ-VAE or VQ-GAN model would consist in dividing the storage space needed for representing input data (e.g., 5 vertical variables times $256 \times 256$ pixels of 32-bit floating point data) vs. the storage space needed by the compressed vector-quantized maps (e.g., for a VQ-VAE with 3 downsampling blocks, $32 \times 32$ indices of a 13-bit dictionary with $2^{13} = 8192$ elements)\footnote{Given that we use overset HEALPix grids detailed in Appendix \ref{sx:appendix:overset_healpix}, the compression is computed as a ratio between $288 \times 288$ 5-channel 32-bit images and $n$-bit $36 \times 36$ maps.}. For this specific example, the compression ratio is about 788. In entropy coding can further reduce the storage footprint, we estimate the compression ratio to be approximately $1100\times$ for that model architecture when compressing test data.

For the purpose of this work, we estimate storage cost in bits as the empirical Shannon entropy of codebook indices per grid, $H = -\sum_i f_i \log_2 f_i$, where $f_i$ is the observed frequency of codebook element $i$ in a given grid. 
This notably ignores expected spatial correlations in the encoded representation (for which the hyperprior model explicitly accounts). The obtained compression ratios for VQ models could thus be improved by fitting a secondary autoregressive model on the codebook index maps\footnote{While not chiefly concerned with compression, \cite{oord2017vqvae} employ autoregressive PixelCNN models of codebook indices to lower-bound the log likelihood of test data as well as for the generation of \emph{de novo} samples.}. The compression ratios obtained with this simplistic entropy computation thus serve as a conservative lower bound.

\subsection{HEALPix map projection}

The representation and projection of the initial data that is to be compressed is a key consideration in the design of neural compression methods, and it is this first step depicted in Figure \ref{fig:system} that we now describe. 
Atmospheric data lies on a sphere, whereas numerical programming natively operates on rectilinear arrays. 
The equirectangular latitude/longitude projection suffers from fundamental limitations in that pixels do not represent consistent units of area across the map, ranging from $28~\text{km} \times 28~\text{km}$ at the equator to negligible area at either pole (where the grid heavily oversamples).
To address these limitations, there exist a multitude of alternative gridding conventions, with the most common reprojections being octahedral~\citep{smolarkiewicz2016fvm} and icosahedral~\citep{zangl2015icon} sampling, cubed spheres~\citep{lin1997cubedsphere,ronchi1996cubed}, dual lat-lon / ``Yin-Yang''~\citep{cote1998gem,ohno2009visualization} maps, and locally linear projections~\citep{metoffice2016data}.

The Hierarchical Equal Area iso-Latitude Pixelation (HEALPix) projection~\citep{gorski1999healpix} defines a curvilinear partitioning of the sphere into 12 equal-area base pixels, further subdivided by powers of 2 into a local square coordinate system, which can be sized to the needs of the application.
The pixels of a HEALPix grid, when traversed diagonally, lie along lines of constant latitude, making conversion to and from spherical harmonics especially efficient\footnote{We have explored the use of cubed sphere \citep{ronchi1996cubed,lin1997cubedsphere} and Yin-Yang \citep{cote1998gem,ohno2009visualization} projections, that have the advantage of \emph{overset}, i.e., overlapping grids, but where grid points do not lie on equally spaced equilatitude rings that prove useful for the computation of spherical harmonics.}.
This ease of conversion to the lingua franca of spherically embedded continuous data makes HEALPix a convenient grid format for conversion to any other grid format.
HEALPix is widely used in astronomy for data situated on the celestial sphere, and is seeing increasing use in atmospheric science in classical NWP models~\citep{karlbauer2023advancing} and in deep learning-based methods~\citep{ramavajjala2024heal, chang2023seamless}. 

Figure \ref{fig:system} illustrates the projection of atmospheric data onto HEALPix, and specifically shows 3 such HEALPix base pixels (base pixel 7 is equatorial whereas base pixels 10 and 11 are polar). It also illustrates the forward transform of the HEALPix representations to compute spherical harmonics coefficients (analysis), and the inverse transform of spherical harmonics coefficients to compute latitude/longitude projections (synthesis). To avoid losing high-frequency information, we choose maximum wavenumber $l_{max}=721$ (equal to the number of latitudes on the latitude/longitude grid and half the number of longitudes during synthesis of the reprojection image) for the spherical harmonics transform; see Appendix~\ref{sx:appendix:spherical_harmonics} for more details. One important consideration when using HEALPix is that the round-trip from latitude/longitude representations to HEALPix projections at our chosen resolution to spherical harmonic coefficients back to latitude/longitude reprojection introduces small interpolation errors, which remove some high-frequency information (as visible in Figure~\ref{fig:spectra}, illustrating the comparison between power spectra of the ground truth and power spectra of the reprojection).

In the experimental analysis that follows, we will use HEALPix with a $256 \times 256$ pixel grid within each of the 12 top-level base pixels, preserving a spatial resolution of around 0.25 degrees per pixel. For reference, the equirectangular projection in which ERA5 data are distributed contains arrays of 721 latitudes by 1440 longitudes, i.e., $1,038,240$ pixels (with significant oversampling at the poles) whereas the HEALPix projection at $256 \times 256$ contains $786,432$ pixels; we show in Appendix \ref{sx:appendix:overset_healpix} that these HEALPix pixels cover smaller areas than lat/lon cells at the equator.
Similarly to \cite{karlbauer2023advancing}, we also extend HEALPix to support overset grids, or grids with overlapping domains at the edges. On each side of each $256 \times 256$ grid, we add a margin of 16 pixels, whose values are interpolated from the neighboring grids with which they overlap. This enables us to remove discontinuities between neighboring grids in the decompressed data, by linearly blending between the two grids' values where they overlap. See Appendix~\ref{sx:appendix:overset_healpix} for details.

\section{Results}
\label{sx:experiments}

\subsection{Data and Standardization}
We study the ECMWF ERA5 dataset~\citep{era5} of atmospheric states. This dataset is the result of reanalysis, or fitting a physics simulation to sparse historical data, thereby generating a "movie" of atmospheric state, with values sampled in a dense regular grid over space and time. The frames are sampled hourly from 1 January 1959 to 10 January 2023, resulting in 561,264 frames.

From this dataset we extracted two subsets for study: a surface level dataset and a vertical levels dataset which spans a range of elevations from the surface up to the lower stratosphere. The surface dataset contains four variables: temperature at 2~m, pressure at surface, and the 2D (zonal and meridional) wind velocity at 10~m. The vertical levels are in fact pressure levels, or isosurfaces of constant pressure, and contain five variables for each level: temperature, specific humidity, geopotential, and the 2D wind velocities. 
Of ERA5's full complement of 137 pressure levels, we selected 13, at 50, 100, 150, 200, 250, 300, 400, 500, 600, 700, 850, 925, and 1000~hPa. When represented in uncompressed 32-bit floating point, these two datasets are 6 TB and 97 TB respectively.
We standardize all atmospheric variables to zero mean, unit standard deviation per pressure level based on the mean and standard deviation computed across all frames of the training dataset.

\subsection{Train/test split and generalisation}

In order to assess the generalization capabilities of our models, we split the data into a training set spanning 1 January 1959 through 31 December 2009, representing 447,072 frames over 51 years, and a test set spanning 1 January 2010 through 10 January 2023, representing 114,192 frames over slightly more than 13 years. Section \ref{sx:discuss:generalisation} investigates if and how reconstruction error deteriorates as a function of time across the test set.
When comparing to the baseline method \citep{huang2022compressing}, we use the dates and timestamps of their ``Dataset 2'', namely 366 frames at midnight of each day in 2016.

\subsubsection{Conditioning on side information}

As terrain features have significant impact on the evolution of atmospheric state, we hypothesized that performance could be improved by augmenting the model with orographic information.
In addition to the atmospheric variables being compressed, the encoder is provided with 6 additional channels of information, corresponding to: the logarithm of the elevation in meters plus 1~m (where elevation had been cropped to be between 0 and 8850~m); the binary land/sea mask, the cosine and sine of the colatitude in radians; the cosine and sine of the longitude in radians. We obtained elevation data from the Google Earth Engine, accessing the GTOPO30 Global 30 Arc-Second Elevation from USGS (at about 1km resolution)\footnote{\url{https://developers.google.com/earth-engine/datasets/catalog/USGS_GTOPO30}}; NaN elevation values correspond to seas: we replace them by 0 for the elevation channel, while also using them to create the second channel with land-sea mask.

Vertical models are trained to reconstruct one pressure level at a time, conditioned on the target vertical level.
We incorporate this information using FiLM layers (feature-wise linear modulation, \citet{perez2018film}) throughout the encoder and decoder.
For simplicity, in these experiments we represent the pressure level as one-hot vector with 13 elements; FiLM at a given layer then corresponds to incorporating a learnable scale and offset per pressure level.

\subsection{Compression Results}

We present experimental support for the following claims:
\begin{enumerate}
    \item We achieve compression ratios of $1000\times$ and above while keeping the mean absolute error (MAE) low, around 0.4$^\circ$~K (for temperature), around 0.5 $\text{m}/\text{s}$ (for zonal and meridional wind), below 1 hPa for surface pressure and around 50 $\text{m}^2/\text{s}^2$ for geopotential.
    \label{claim:cr_and_mae}
    \item Only a small number of reconstructed pixels exhibit errors that exceed given thresholds: about 0.5\% pixels beyond 1$^\circ$~K error in temperature, about 0.5\% pixels beyond 1 $\text{m}/\text{s}$ error in zonal and meridional wind, about 0.15\% pixels beyond 1 hPa error in surface pressure, and about 0.05\% pixels beyond 100 $\text{m}^2/\text{s}^2$ error in geopotential. \label{claim:few_bad_pixels}
    \item Our spectral error is kept low, preserving most high-frequency features in the reconstruction images.
    \item We preserve rare and extreme events such as hurricanes and heat waves.
\end{enumerate}
Below we evaluate a number of models, described in section~\ref{sx:methods:networks}. Of these, the hyperprior model trained with rate-distortion penalty coefficient 0.025 achieved a compression ratio of around $1000 \times$ while satisfying all of the above claims, in particular preserving the spectral error over most spatial scales. For comparison, we also exhibit the VQ-VAE with 3 downsampling blocks, which outputs $36 \times 36$ VQ maps of 13-bit (or 8192-element) indices into a learned codebook of quantized vectors; that VQ-VAE model achieves a compression ratio $1100 \times$. We observe that the VQ-GAN architecture with similar VQ code dictionary size and number of layers performed similarly to VQ-VAE, which is why we focus only on the VQ-VAE in the analysis.
Finally, we compare these models with a VQ-VAE with 4 downsampling blocks trained exclusively to reconstruct one single variable (geopotential), with a compression ratio of $1800 \times$, and we discuss the advantages and drawbacks of single-channel compressors.

\subsection{High compression ratio with low RMSE \& MAE}

We first define the metrics and notation used in the results. Following the notation of \cite{lam2023graphcast}, we define $D_{\text{eval}}$ as the set of evaluation dates and timestamps, $G_h$ as the grid of $256 \times 256$ pixels in HEALPix tile $h$, while $x_{j,i}$ and $\hat x_{j,i}$ respectively are the ground truth and the reconstruction at the $i$-th pixel of a grid at the $j$-th variable and pressure level. Equations \ref{eq:mae} and \ref{eq:rmse} describe the MAE and RMSE metrics we use for HEALPix projection results.

\begin{equation}
    \text{MAE} (j) = \frac{1}{12 \times | D_{\text{eval}} |} \sum _{d \in D_{\text{eval}}, 0 \leq h \leq 11} \frac{1}{| G_h |} \sum _{i \in G_h} | x_{j,i} - \hat x_{j,i} |
    \label{eq:mae}
\end{equation}

\begin{equation}
    \text{RMSE} (j) = \frac{1}{12 \times | D_{\text{eval}} |} \sum _{d \in D_{\text{eval}}, 0 \leq h \leq 11} \sqrt{\frac{1}{| G_h |} \sum _{i \in G_h} || x_{j,i} - \hat x_{j,i} ||_2^2}
    \label{eq:rmse}
\end{equation}

On the re-projected latitude/longitude grid $G_{0.25^\circ}$ of cells spaced by $0.25^\circ$, we define weighted MAE (Eq. \ref{eq:weighted_mae}) and weighted RMSE (Eq. \ref{eq:weighted_rmse}) as weighted by coefficient $a_i$ which is proportional to the area of the latitude/longitude cell (i.e., to the cosine of the colatitude of the $i$-th cell of latitude/longitude grid) and normalized to sum to 1 over the grid. We also define the traditional weighted RMSE, used by \citet{huang2022compressing}, in Equation \ref{eq:weighted_rmse_trad}.

\begin{equation}
    \text{wMAE} (j) = \frac{1}{| D_{\text{eval}} |} \sum _{d \in D_{\text{eval}}} \frac{1}{| G_h |} \sum _{i \in G_{0.25^\circ}} a_i | x_{j,i} - \hat x_{j,i} |
    \label{eq:weighted_mae}
\end{equation}

\begin{equation}
    \text{wRMSE} (j) = \frac{1}{| D_{\text{eval}} |} \sum _{d \in D_{\text{eval}}} \sqrt{\frac{1}{| G_{0.25^\circ} |} \sum _{i \in G_h} a_i || x_{j,i} - \hat x_{j,i} ||_2^2}
    \label{eq:weighted_rmse}
\end{equation}

\begin{equation}
    \text{wRMSE}_{\text{trad}} (j) = \sqrt{\frac{1}{| D_{\text{eval}} |} \sum _{d \in D_{\text{eval}}} \frac{1}{| G_{0.25^\circ} |} \sum _{i \in G_h} a_i || x_{j,i} - \hat x_{j,i} ||_2^2}
    \label{eq:weighted_rmse_trad}
\end{equation}

Figure~\ref{fig:reprojection_error_vs_cr} shows the reconstruction results evaluated in latitude/longitude projection, namely the RMSE, the MAE and the 0.99999 quantile of error for several models. For our target compression ratio of $1000 \times$, we find the hyperprior model to be the strongest performer. Additionally, Figures \ref{fig:error_vs_cr} in the Appendix provides results in HEALPix projection, comparing the various models both on surface and vertical data reconstruction at multiple levels.   

\subsection{Small number of erroneous pixels in reconstructions}
\label{sx:experiments:bad-pixels}

\begin{figure}[t]
    \centering
    \includegraphics[width=0.32\textwidth, trim = 0cm 0cm 1cm 0cm, clip]{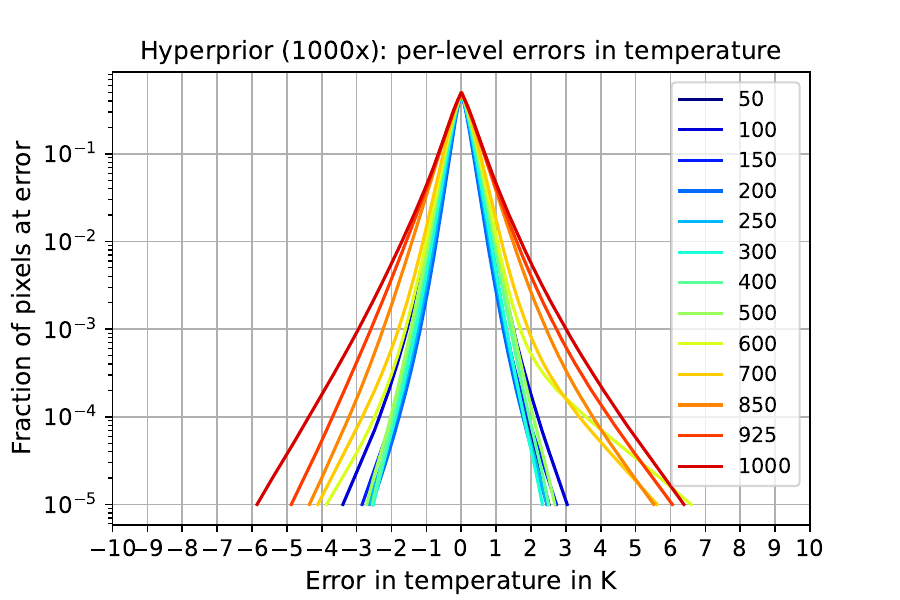}
    \includegraphics[width=0.32\textwidth, trim = 0cm 0cm 1cm 0cm, clip]{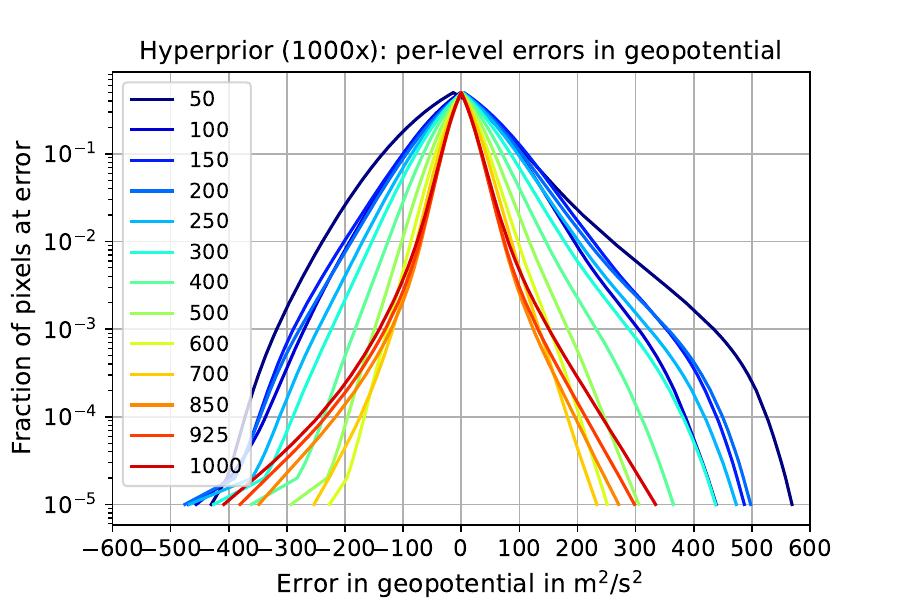}
    \includegraphics[width=0.32\textwidth, trim = 0cm 0cm 1cm 0cm, clip]{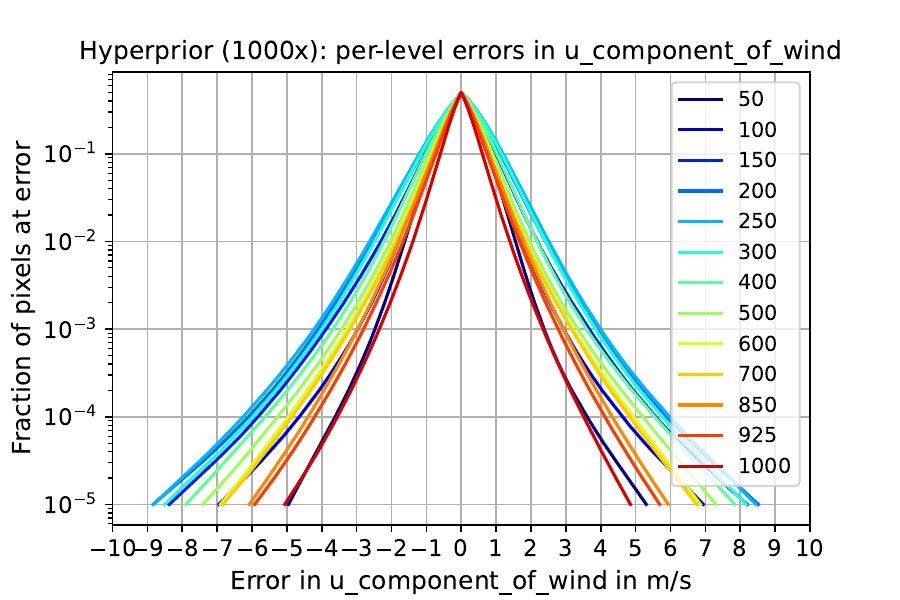}
    \includegraphics[width=0.32\textwidth, trim = 0cm 0cm 1cm 0cm, clip]{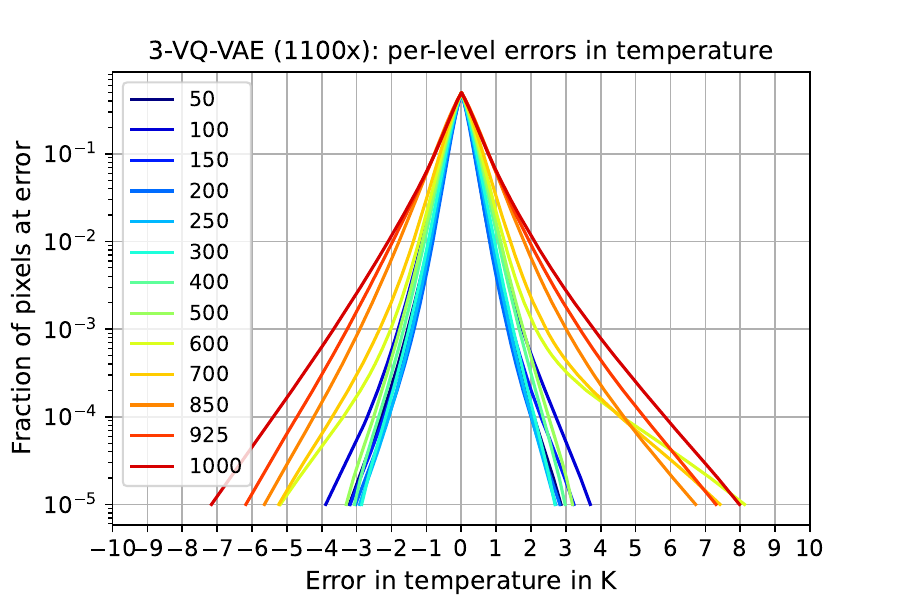}
    \includegraphics[width=0.32\textwidth, trim = 0cm 0cm 1cm 0cm, clip]{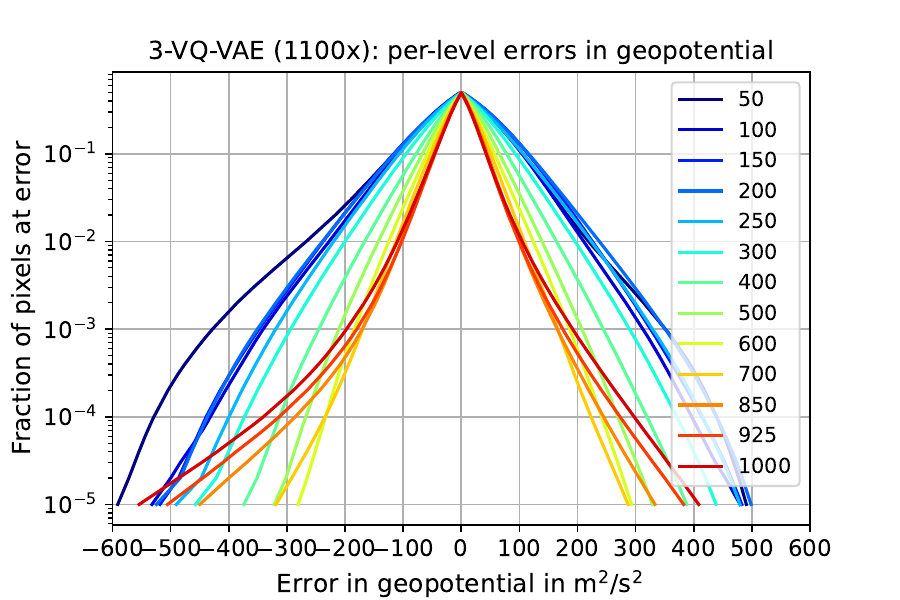}
    \includegraphics[width=0.32\textwidth, trim = 0cm 0cm 1cm 0cm, clip]{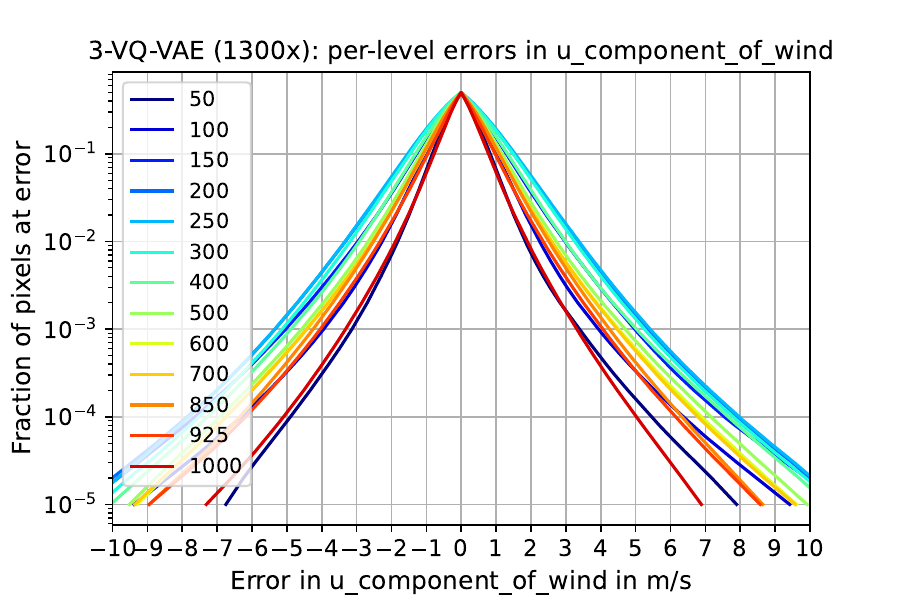}
    \caption{Distribution of signed errors, with the error in the variable units on the X axis and the fraction of pixels in the evaluation set (2010-2023) at that error levels. Top row shows hyperprior results, bottom row shows 3-block VQ-VAE results, with columns corresponding to temperature, geopotential and zonal wind speed. On each plot we show coloured curves for each of the 13 levels in the vertical dataset.}
    \label{fig:error_histograms}
\end{figure}

Figure~\ref{fig:error_histograms} presents a histogram with the distribution of absolute pixelwise reconstruction errors over the entire evaluation set, for three of the vertical variables (temperature, geopotential and zonal wind) and for each elevation level (colour-coded from 50 hPa in blue to 1000 hPa in red). On the top row, corresponding to the hyperprior model, the distributions show sparse tails, with few pixels (0.01\%) reaching high errors of $\pm 4^\circ$~K, $\pm 400~\text{m}^2/\text{s}^2$ or $\pm 6~\text{m}/\text{s}$.

Tables~\ref{tab:percent_bad_pixels_hyper} (for the hyperprior model) and Table \ref{tab:percent_bad_cells_vqvae} (for the VQ-VAE with 3 downsampling blocks) quantify these ``bad pixels'', showing that only 0.5\% pixels at most have absolute errors exceeding 1$^\circ$~K (temperature) or 1 $\text{m}/\text{s}$ (zonal and meridional wind speeds) across all pressure levels.

Because only a few pixels exceed this maximum error threshold, we can trivially cap the maximum error by storing the original uncompressed values of these ``bad pixels'' in a list, to be stored alongside the compressed representation. When reconstructing the image, we overwrite the bad pixels using their original values, reducing their reconstruction error to zero, and capping the maximum error over all reconstructions. The sparsity of bad pixels limits the size of this list, keeping the overall compression ratio low even after accounting for the list's storage cost. The trade-offs of the different strategies for storing this auxiliary information, in terms of compression ratio and distortion characteristics, require further study.

\begin{table}[]
    \centering
    \caption{Fraction of pixels with absolute error exceeding given error thresholds, for the hyperprior model with about $1000 \times$ compression ratio.}
    \begin{tabular}{c|c|c|c}
        Variable & Error thresh. &\% bad pixels & Compression ratio \\
        \hline
Temperature (all levels) & 1$^\circ$~K & 0.5\% & $1059 \times$ \\
        Zonal wind (all levels) & 1 $\text{m}/\text{s}$ & 0.5\% & $1059 \times$ \\
        Meridional wind (all levels) & 1 $\text{m}/\text{s}$ & 0.5\% & $1059 \times$ \\
        Geopotential (all levels) & 100 $\text{m}^2/\text{s}^2$ & 0.05\% & $1059 \times$ \\
    \end{tabular}
    
    \label{tab:percent_bad_pixels_hyper}
\end{table}

\begin{table}[]
    \centering
    \caption{Fraction of pixels with absolute error exceeding given error thresholds, for the 3-block VQ-VAE model with $1100 \times$ compression ratio, and for the 4-block VQ-VAE model with $1800 \times$ compression ratio trained only on geopotential (last row).}
    \begin{tabular}{c|c|c|c}
        Variable & Error thresh. & \% bad pixels & Compression ratio \\
        \hline
Temperature (all levels) & 1$^\circ$~K & 0.52\% & $1110 \times$ \\
        Zonal wind (all levels) & 1 $\text{m}/\text{s}$ & 0.51\% & $1110 \times$ \\
        Meridional wind (all levels) & 1 $\text{m}/\text{s}$ & 0.51\% & $1110 \times$ \\
        Geopotential (all levels) & 100 $\text{m}^2/\text{s}^2$ & 0.063\% & $1110 \times$ \\
        \hline
        Geopotential$^*$ (all levels) & 100 $\text{m}^2/\text{s}^2$ & 0.012\% & $1796 \times$ \\
    \end{tabular}
    
    \label{tab:percent_bad_cells_vqvae}
\end{table}

\subsection{Spectral properties of the reconstructions}

\begin{figure}
    \centering
    \includegraphics[width=\textwidth, trim = 5cm 0cm 5cm 0cm, clip]{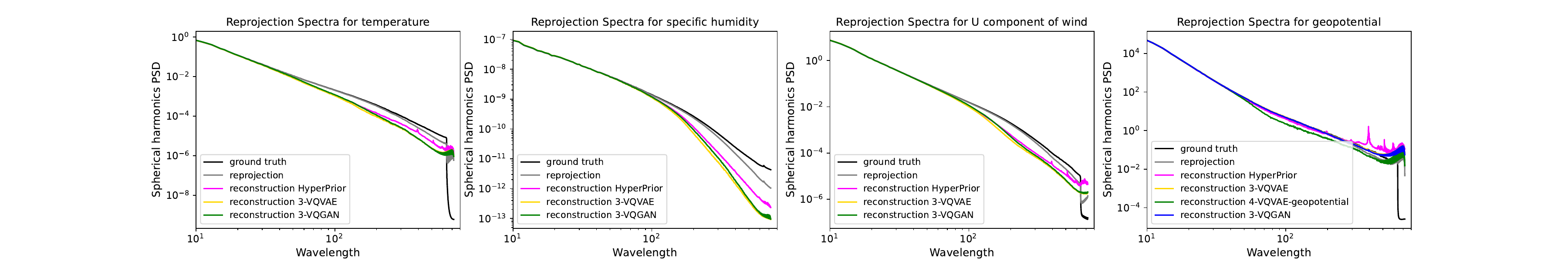}
    \hfill
    \caption{Power spectrum density for temperature, specific humidity, zonal wind and geopotential, computed using forward spherical harmonics transform. Results are computed on the daily 2016 baseline dataset.}
    \label{fig:spectra}
\end{figure}

Figure~\ref{fig:spectra} shows the power spectrum density for 4 of the reconstructed variables (temperature, specific humidity, zonal wind and geopotential), with spectra averaged over all equirectangular latitude/longitude frames and over all vertical levels. We notice that some high-frequency information is automatically lost after a round-trip consisting of HEALPix projection from latitude/longitude, spherical harmonic analysis of HEALPix tiles and spherical harmonic synthesis from the spherical harmonics coefficients.

Figure~\ref{fig:spectra} shows that the spectrum of hyperprior reconstructions is closest to the spectrum of ground truth than that of 3-block VQ-VAEs for all variables.
Appendix~\ref{sx:appendix:spherical_harmonics} details the computation of spherical harmonics on HEALPix tiles, and the extraction of the power spectral density from spherical harmonics coefficients.

\subsection{Extreme values}

\begin{figure}
    \centering
    \includegraphics[width=0.9\textwidth, trim = 6cm 2.5cm 5cm 2cm, clip]{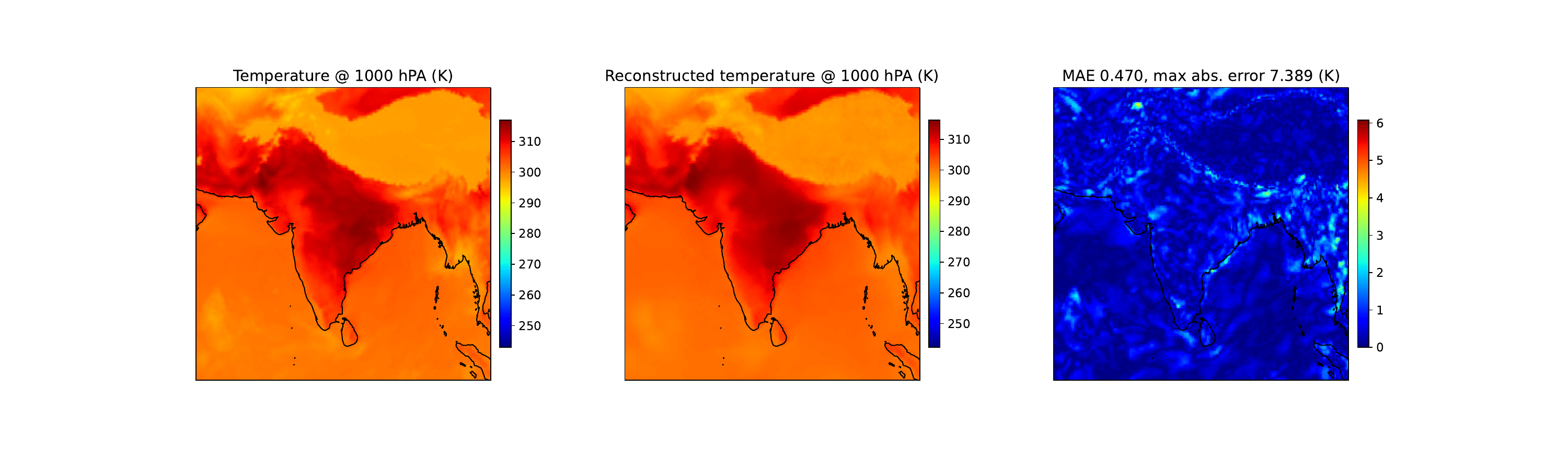}
    \includegraphics[width=0.9\textwidth, trim = 6cm 2.5cm 5cm 2cm, clip]{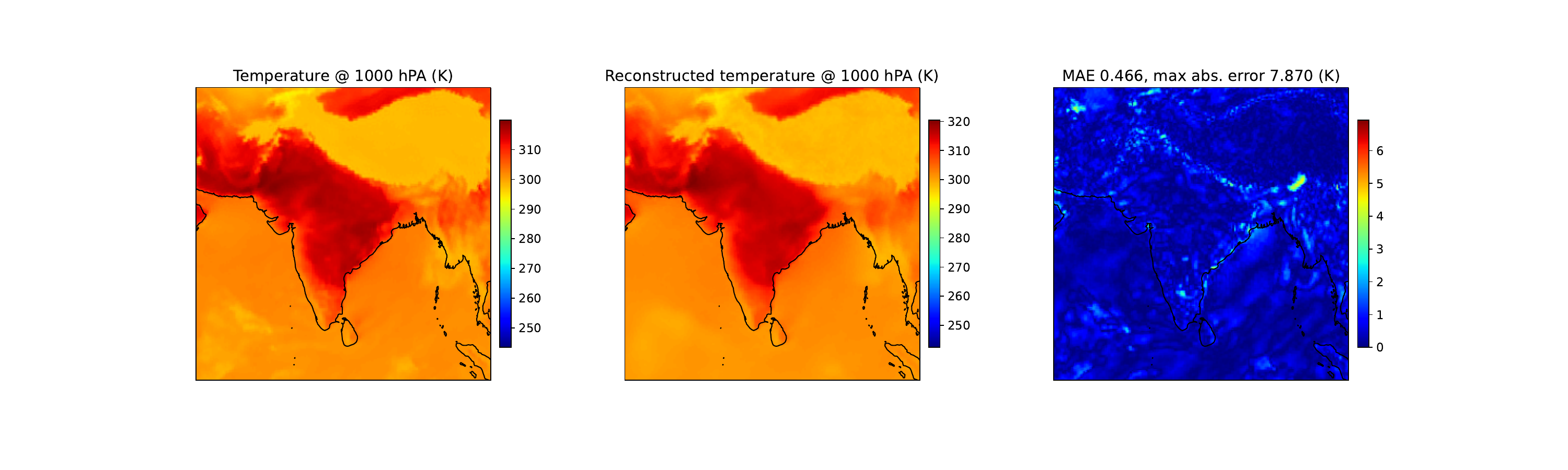}
    \includegraphics[width=0.9\textwidth, trim = 6cm 2.5cm 5cm 2cm, clip]{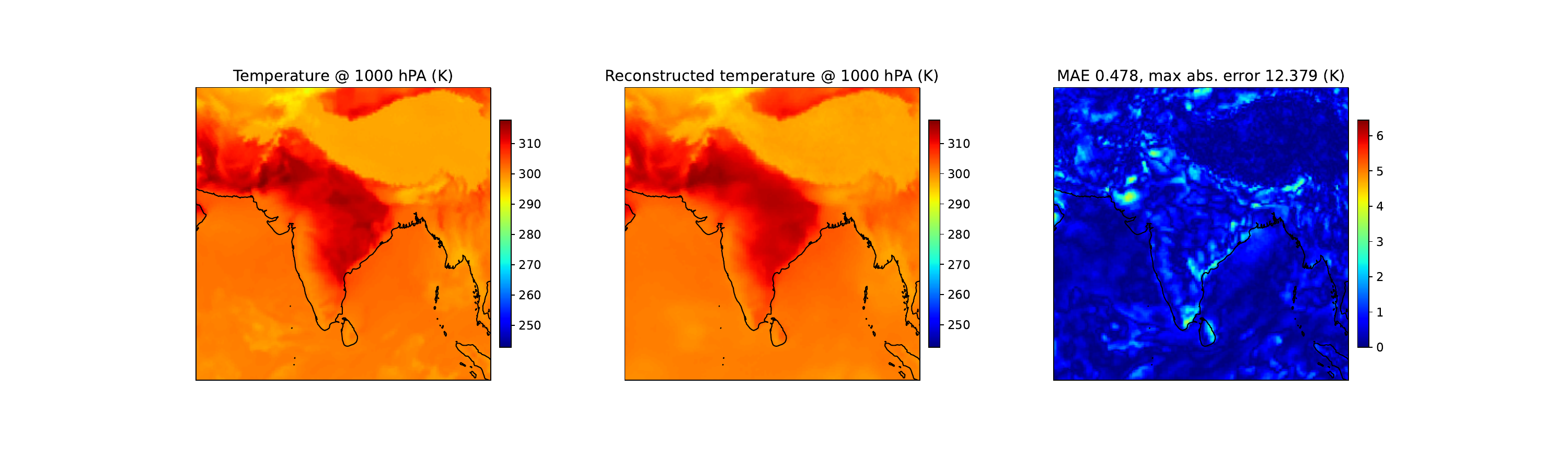}
    \includegraphics[width=0.9\textwidth, trim = 6cm 2.5cm 5cm 2cm, clip]{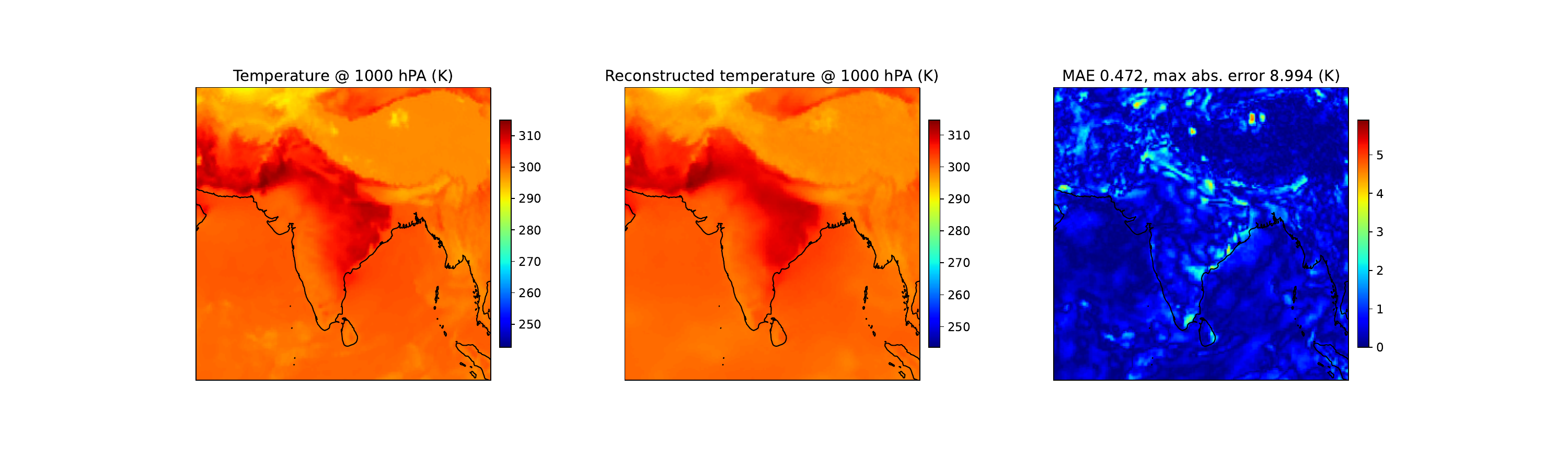}
    \hfill
    \caption{Reconstructions of temperature at 1000 hPa from 2015/5/23 at 6, 12, 18 and 24 UTC (during the 2015 Indian heatwave). Globally, MAE is about $0.47-0.48^\circ~\text{K}$ and most the area of interest is under $1^\circ~\text{K}$ error. The residual show that a few scattered areas near Bhutan or near Karachi have reconstruction over-estimated by about $2-3~^\circ~\text{K}$.}
    \label{fig:heatwave}
\end{figure}

In order to more fully understand the behaviour of these models, we manually select dates corresponding to extrema in temperature, such as Indian heatwave of 2015, and to hurricanes, such as Hurricane Harvey in September 2017 and Hurricane Matthew in October 2016. Figures \ref{fig:hurricane_matthew_hyperprior} and \ref{fig:hurricane_matthew_vqvae} illustrate the performance of the hyperprior and 3-block VQ-VAE models on Hurricane Matthew, and show that geopotential and wind speed information corresponding to the eye of the cyclone is preserved after compression and reconstruction. Figure~\ref{fig:heatwave} illustrates the reconstruction of the temperature at 1000 hPa over 24 hours on 23 May during the 2015 Indian heatwave, and shows that temperature extrema up to around $45^\circ~\text{C}$ are correctly represented.

Appendix~\ref{sx:appendix:hist2d-err-value} contains additional analysis of the reconstruction errors, which suggest that the hyperprior compression model errors are roughly uniformly distributed over the target values, and that the model tends to overestimate higher temperature and wind speed and underestimate lower temperature or wind speed.

\subsection{Comparison with baseline method}

To be able to compare results with \cite{huang2022compressing}, we use the same dataset (described in \cite{ashkboos2022ens}) which contains 11 pressure levels, including 10 of ours and one extra pressure level at 10 hPa. Appendix~\ref{sx:appendix:huangiclr2023} explains how we run their code and establish parity between evaluating on their 11 pressure levels and our 13 pressure levels.

We also define the same subset of data (corresponding to Dataset 2 in \citet{huang2022compressing}), namely 366 daily frames (at 0 UTC) in 2016\footnote{Note that results on Figs. \ref{fig:spectra}, \ref{fig:error_vs_cr} and \ref{fig:reprojection_error_vs_cr_bounded_compression} are computed on the reduced subset of 366 daily frames (at 0 UTC) in 2016.}.

In Figure~\ref{fig:error_vs_cr}, our VQ-VAE, VQ-GAN and hyperprior models achieve significantly lower temperature errors (e.g., $\text{MAE}=0.3^\circ$~K for the $1000 \times$ hyperprior model) than the baseline method. For geopotential, our compression results are marginally better than the baseline ($\text{MAE}=45~\text{m}^2/\text{s}^2$ for the $1000 \times$ hyperprior model, vs. $60~\text{m}^2/\text{s}^2$ for the $1300 \times$ baseline).

Two main observations arise: 1) it is worth investigating single-variable compression of geopotential, similarly in the baseline models (see additional results in Section \ref{sx:discuss:geopotential}), and 2) the baseline method can achieve similar if not better error maxima. We discuss error maxima and provide detailed analysis of the possible origin of the errors in Appendix~\ref{sx:appendix:analysis}.

In concurrent work,~\citet{han2024cra5} demonstrate a neural compressor applied to ERA5 in lat-lon format, with a number of important differences from our own work. While we explore both VQ-VAE and hyperprior networks as candidate models, \citet{han2024cra5} propose a stagewise training method based on a single VQ-VAE network incorporating a window-based attention mechanism to encode the data to latent variables, and another to serve as a hyperprior block to encode first-level variances.
They validate and test on 2022 and 2023 data respectively, whereas we train on data up to 2010 and evaluate long-term generalisation on 2010-2023.
They report comparable RMSE to what we report in this work, while that work reports compression ratios of $229-323\times$ compared to the ratios of $1000-3000\times$ that we report.

\begin{figure}
    \centering
    \includegraphics[width=1\textwidth, trim = 6cm 0cm 6.5cm 0cm, clip]{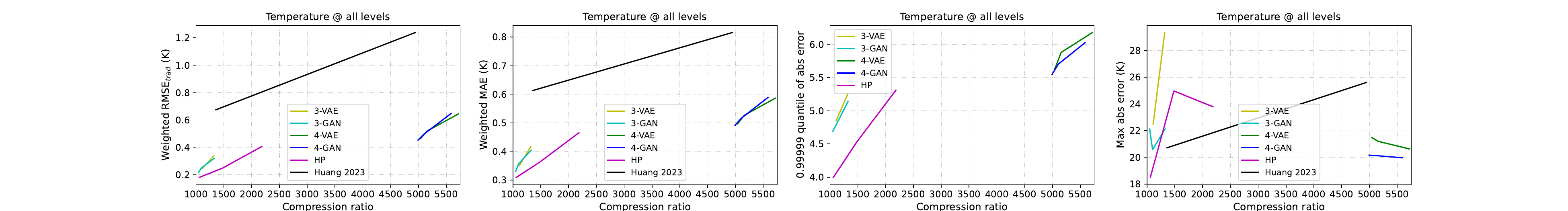}
    \includegraphics[width=1\textwidth, trim = 6cm 0cm 6.5cm 0cm, clip]{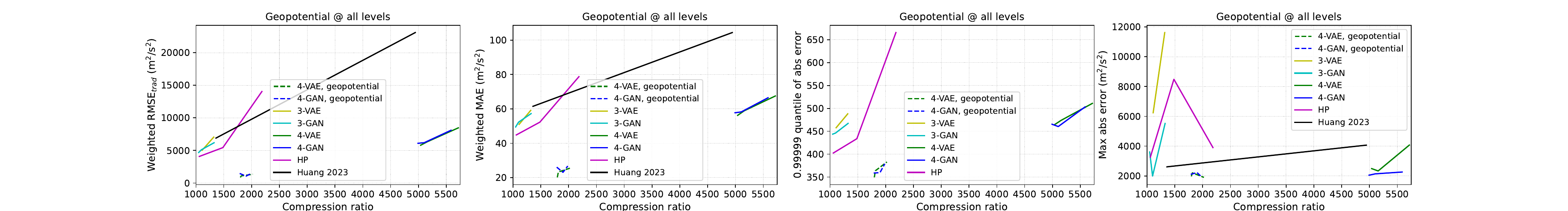}
    \includegraphics[width=1\textwidth, trim = 6cm 0cm 6.5cm 0cm, clip]{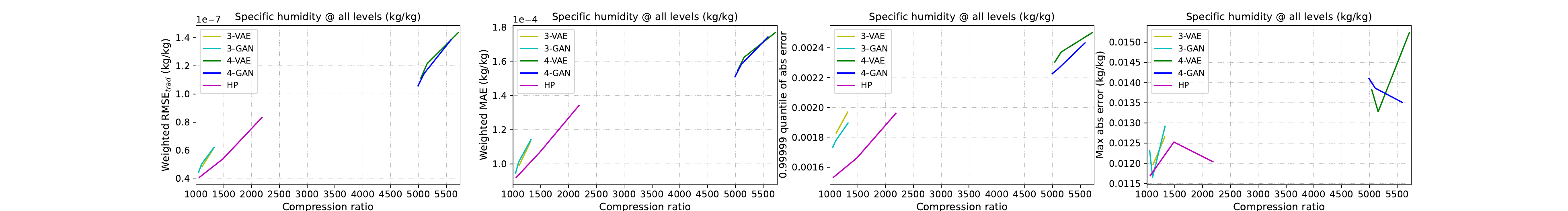}
    \includegraphics[width=1\textwidth, trim = 6cm 0cm 6.5cm 0cm, clip]{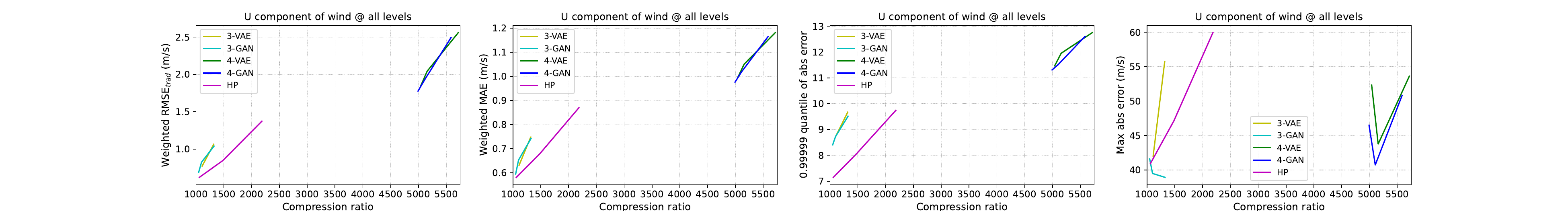}
    \hfill
    \caption{Reconstruction and reprojection error vs. compression ratio, for vertical data, in equirectangular projection. Columns: metrics: respectively root mean squared error (RMSE), mean absolute error (MAE), 0.99999 quantile of error and maximum error over the daily 2016 baseline dataset. Lower errors and higher compression rates are better (bottom right of each plot).}
    \label{fig:reprojection_error_vs_cr}
\end{figure}

\section{Analysis and Discussion}
\label{sx:discussion}

Next, we motivate the use of neural compression methods by comparison with error-bound compression, discuss the possibility of compressing high-error residuals (Section \ref{sx:discuss:error-bound}), and present results for single-variable models (Section \ref{sx:discuss:geopotential}). A principal limitation of  neural compression methods for scientific data is that they have unbounded (and potentially high) maximum element-wise reconstruction errors. We investigate and explain a cause of this high error for the ERA5 dataset as being due to artefacts present in the dataset as well as joint reconstruction of multiple variables (Section \ref{sx:discuss:limitation-artefacts}). Finally, we evaluate performance of generalization over time in Section \ref{sx:discuss:generalisation}).

\subsection{Comparison with error-bounded compression}
\label{sx:discuss:error-bound}

\begin{figure}
    \centering
    \includegraphics[width=1\textwidth, trim = 3cm 0cm 4cm 0cm, clip]{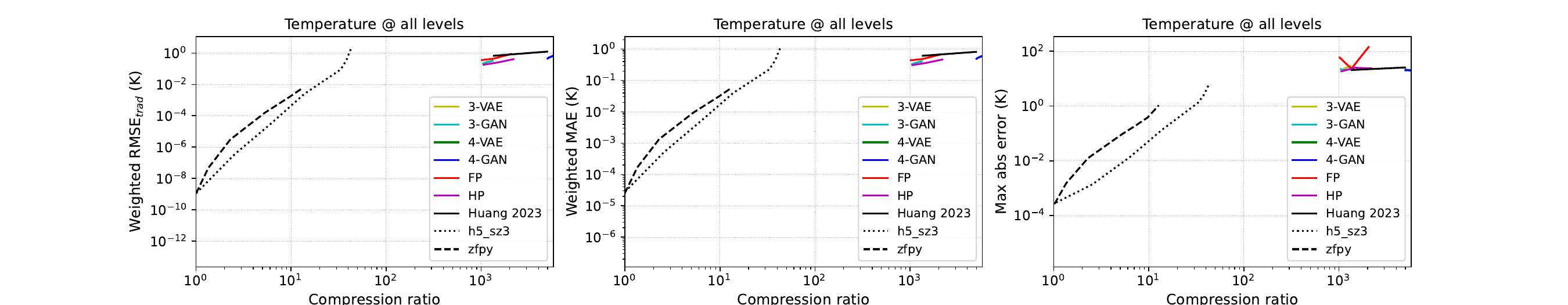}
    \hfill
    \caption{Reconstruction and reprojection error vs. compression ratio, for vertical data, in equirectangular projection. Columns: metrics: respectively root mean squared error (RMSE), mean absolute error (MAE) and maximum error. Results are computed on the daily 2016 baseline dataset. Lower errors and higher compression rates are better (bottom right of each plot).}
    \label{fig:reprojection_error_vs_cr_bounded_compression}
\end{figure}

Our compression techniques achieve strong compression ratios for comparable mean reconstruction errors compared to conventional and competing methods.
However, they lack the precise error bound guarantees provided by off-the-shelf compression tools. 
To address this, we investigated compressing the residuals post-neural compression using both~\citet{sz3_algo} and~\citet{zfpy_2014} to ensure both compression performance and error control. As shown in Figure \ref{fig:reprojection_error_vs_cr_bounded_compression}, bounded compression methods such as SZ3 achieve far lower compression ratios of temperature (i.e., less than $50\times$) for similar errors (namely, weighted RMSE and weighted MAE of the order of $1^\circ$~K) than the neural compression models.

We compared further compressing the residuals of neural compression (e.g., the hyperprior reconstructions in HEALPix projection minus the ground truth in HEALPix projections) and observed similar compression ratios at given error bounds; in other words, compressing the residual was as expensive as compressing the original data using bounded compression methods. We note that SZ3 operates on 1D data, and as such does not exploit the spatial correlations in the data.

From the results in section~\ref{sx:experiments:bad-pixels}, about $0.5\%$ of reconstructed HEALPix pixels present errors higher than the error threshold of $\pm 1^\circ$~K. We could consider compressing the values of those pixels using the SZ3 bounded compression method at a compression ratio of the order of $10\times$, which would result in the order of $1000\times$ compression ratio. However, we would still need to store the byte position (row and column) of those high-error pixels in the $256 \times 256$ tile, or preferably, develop a storage strategy that exploits local spatial correlations of those erroneous pixels. It is also not obvious that such a method would not create additional discontinuities (and as a consequence, high-frequency artefacts) in the reconstructed images. 

\subsection{Comparison with single-variable models}
\label{sx:discuss:geopotential}

The approach we present is developed for multivariate compression, i.e. compression of multiple atmospheric variables using a single latent code.
It is also possible to develop compression models for individual variables.
Figure \ref{fig:reprojection_error_vs_cr} shows that the 4-block VQ-VAE model trained only on geopotential data manages to reconstruct geopotential with $\text{MAE} = 0.2~\text{m}^2/\text{s}^2$ at all levels latitude/longitudewhile maintaining an effective compression ratio of $1800\times$. The last row of Table \ref{tab:percent_bad_cells_vqvae} highlights that very few (0.012\%) pixels cross the 100 $\text{m}^2/\text{s}^2$ error threshold. The right-most column of Figure \ref{fig:spectra} shows that the power spectrum of that model's reconstructions of geopotential are the closest to the spectrum of the reprojection.

\subsection{Difficulty in learning and reconstructing specific variables and pressure levels}
\label{sx:discuss:limitation-artefacts}

\begin{figure}
    \centering
    \includegraphics[width=0.45\textwidth, trim = 0cm 25.5cm 2cm 3.5cm, clip]{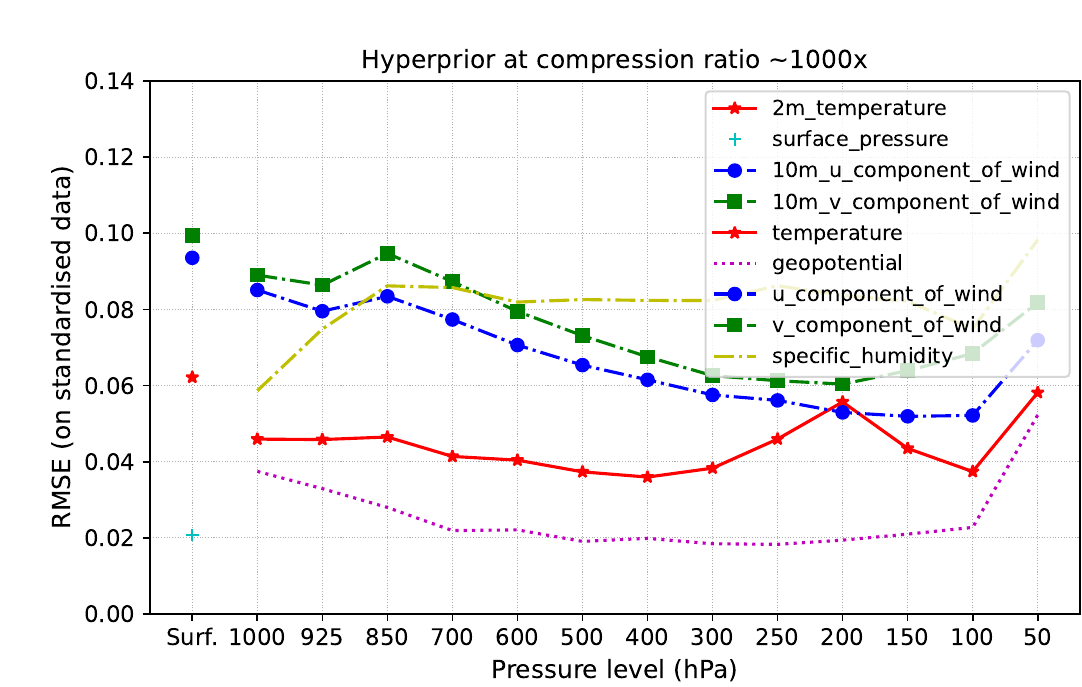}
    \includegraphics[width=0.45\textwidth, trim = 0cm 25.5cm 2cm 3.5cm, clip]{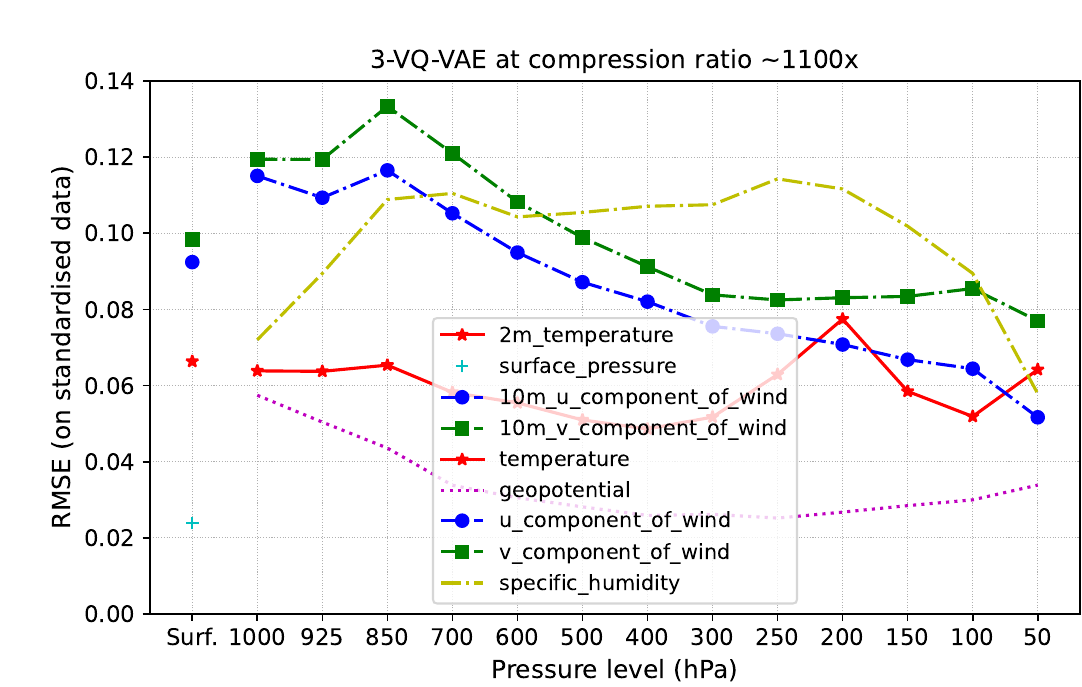}
    \hfill
    \caption{Standardised RMSE of different variables and atmospheric levels for a single hyperprior model (left) and a 3-block VQ-VAE model (right) at about $1000\times$ compression ratios.}
    \label{fig:rmse_comparison_variables}
\end{figure}

We investigate which variables and pressure levels pose particular challenges for our models.
Recall that prior to training, individual variables are standardised to zero mean, unit standard deviation per pressure level, with means and standard deviations computed across all frames of the training dataset.
While it is important to ground most analyses in ``unstandardised'' computations such that deviations between reconstruction and ground truth are reported the correct units for each variable, comparing the characteristics of individual reconstructed variables before the standardisation is undone can provide insight into the information being preserved by the encoded representation.

Figure \ref{fig:rmse_comparison_variables} depicts the \emph{standardized} average root mean square error computed separately for different variables and different pressure levels, for both the hyperprior model at around $1000\times$ compression rate and the 3-block VQ-VAE model at $1100\times$ compression rate. We first notice that the hyperprior model achieves lower RMSE across all variables than the 3-block VQ-VAE model. We also observe that specific humidity consistently has the lowest RMSE overall among vertical data models, and that surface pressure has the lowest RMSE among surface models; then comes temperature, and then wind speed and geopotential. We observe interesting patterns persistent across models, such as wind speeds at 850 hPa having the highest RMSE.

We notice that the main difference between the RMSE profile for the hyperprior model and the VQ-VAE model is the higher RMSE at 50 hPa. In fact, other deep learning models seem to have faced relatively poor performance at the 50 hPa level when training on the same dataset, as evidenced by the per-pressure level scorecards for the GraphCast model~\citep{lam2023graphcast} (in particular, see Figs. 2 and 29, where the prediction skills is significantly lower than at other pressure levels). We surmise that these reconstruction or prediction errors are due to out-of-sample, low-value anomalies in the specific humidity at 50 hPa, at the limit of the upper part of the troposphere, and present a detailed analysis of the data artefacts in Appendix \ref{sx:appendix:artefacts}.
A potential remedy is to simply extend the training set: while autoencoders can be applied to data beyond the training set, it is entirely valid to train a bespoke compressor and decompressor on an entire target dataset if the performance requirements justify the additional computation.

Figure \ref{fig:error_histograms} seems to suggests that error maxima should be bounded, as the $10^{-5}$ fraction of errors 
corresponds to about $\pm 5^\circ$~K (temperature), $\pm 400~\text{m}^2/\text{s}^2$ (geopotential) or $\pm 8~\text{m}/\text{s}$ (zonal wind speed). However, looking at the error maxima on Figure \ref{fig:reprojection_error_vs_cr} shows that the max errors (over 366 frames of the daily 2016 dataset subset) of the hyperprior are as high as $\pm 18^\circ$~K, $\pm 4000~\text{m}^2/\text{s}^2$ and $\pm 40~\text{m}/\text{s}$. Higher error maxima appear on the full test dataset covering 2010-2023.

Our error analysis exposes three potential shortcomings of autoencoder methods. The first is that these models lack error-bound guarantees.
Second, these methods seem to poorly handle out-of-sample artefacts in the data, which may cause them to produce reconstructions with unbounded errors (see analysis of artefacts in Appendix \ref{sx:appendix:artefacts}). Lastly, while compressing variables jointly fruitfully exploits statistical redundancy across variables, this results in brittleness, where an artefact present in the values of one variable may contaminate the reconstruction of another variable.

\subsection{Generalisation of compression over time}
\label{sx:discuss:generalisation}

Unlike methods where the neural network is trained to be a functional over the training data, such as our baseline data compression method \citep{huang2022compressing}, our autoencoder-based approach allows for generalization to unseen data frames; this is a significant advantage as we do not need to retrain the compression model to handle new data frames.

Given that weather and climate processes are not stationary, a pressing question is whether, and how quickly, the compression model deteriorates as a function of time. Our dataset split with training on data covering 1959 through 2009, and evaluation on data covering 2010 through 10 January 2023, allows us to test that hypothesis over 13 years. 

\begin{figure}[tbp]
    \centering
    \includegraphics[width=\textwidth, trim = 5.5cm 0cm 5cm 0cm, clip]{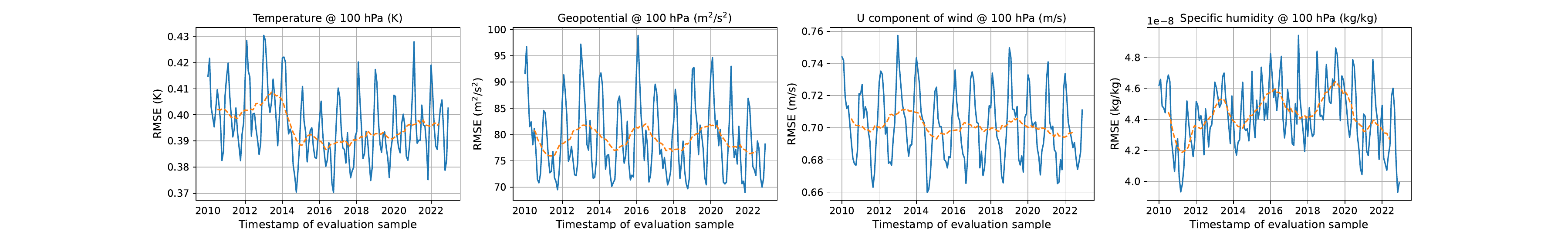}
    \includegraphics[width=\textwidth, trim = 5.5cm 0cm 5cm 0cm, clip]{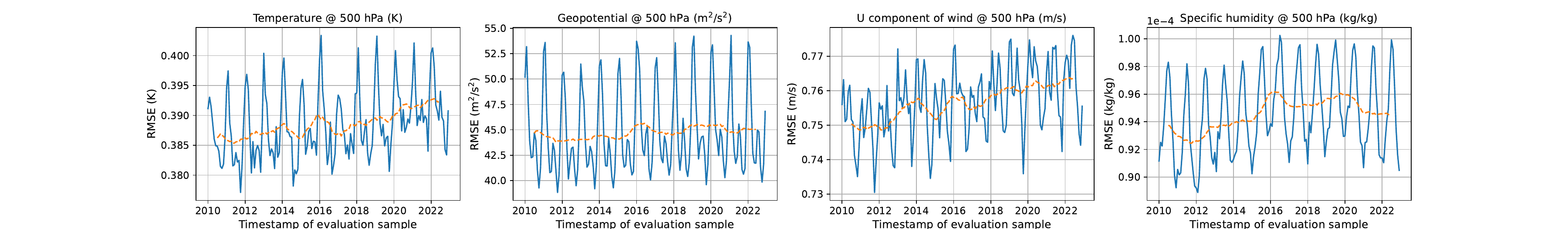}
    \includegraphics[width=\textwidth, trim = 5.5cm 0cm 5cm 0cm, clip]{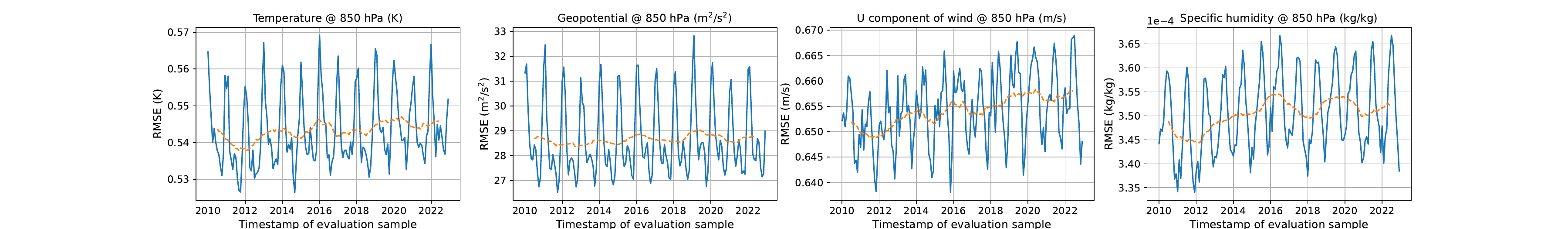}
    \includegraphics[width=\textwidth, trim = 5.5cm 0cm 5cm 0cm, clip]{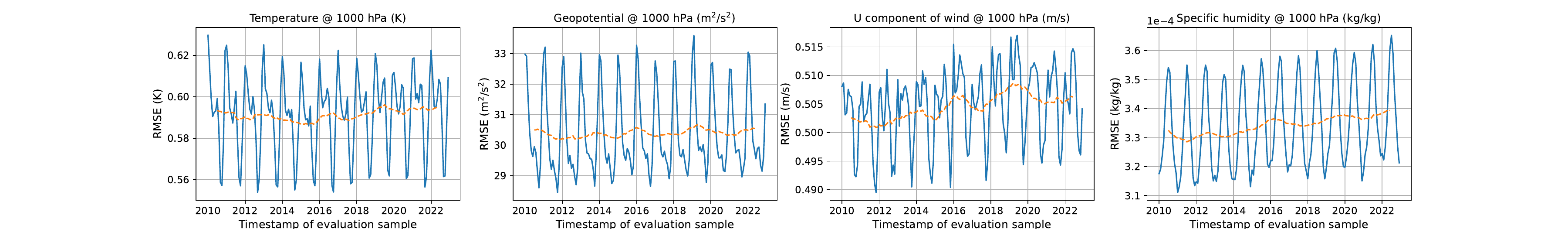}
    \hfill
    \caption{Plot of the average RMSE over 13 years of the test set, from 1/1/2010 to 31/12/2022, using the hyperprior with about $1000\times$ compression rate. Rows from top to bottom show results at 50 hPa, 100 hPa, 150 hPa, 500 hPa, 850 hPa and 1000 hPa, and columns show temperature, geopotential, zonal wind speed and specific humidity. The blue curve shows the monthly average (and exhibits strong seasonal tendencies); the red dashed curve shows the yearly rolling average (and suggests an upward trend over the years, the further away the test sample from the training data cutoff date of 31 December 2009).}
    \label{fig:generalisation-time}
\end{figure}

Figure \ref{fig:generalisation-time} illustrates the absolute value RMSE (averaged over HEALPix squares, hours of the days and days of the month) for 4 variables and 5 different pressure levels, as well as a yearly rolling average. The yearly average of temperature, wind speed and specific humidity at 500 hPa, 850 hPa and 1000 hPa seem to have an upward trend over the years. The simplest explanation for such a trend would be that the model is expected to perform worse on test samples that are further away from the training data cutoff date (in our case, 31 December 2009).

However, the monthly average clearly exhibits strong seasonal tendencies that dominate the variability of the yearly rolling average. The standardized RMSE has similar seasonalities, as does the RMSE of the 3-block VQ-VAE model. We also know that because of the unequal land mass distribution between the northern and southern hemisphere, weather variables such as global temperature exhibit seasonality.

\section{Conclusion}
\label{sx:conclusion}

We propose a method for the neural compression of atmospheric states using the area-preserving HEALPix projection to simplify the processing, by conventional neural networks, of data situated on the sphere. We highlight that this choice of projection allows for efficient computation of spherical harmonics, thereby simplifying both the reprojection of decompressed atmospheric states to latitude/longitude coordinates and straightforward interrogation of the spectral properties of the reconstructions.
We conduct a detailed analysis of the performance characteristics of several neural compression methods, demonstrating very high compression ratios in excess of $1000\times$ with minimal distortion, while also drawing attention to several shortcomings.
We find that the hyperprior model, even with a relatively simple encoder and decoder, appears best suited to the task among the models studied.
Notably, the hyperprior model largely preserves the shape of the power spectrum, despite being trained only via mean squared error and a compressibility penalty.
We examine performance of the studied compressors on states representing extreme atmospheric events and find that these, too, are well-preserved.
These results make a compelling case for the viability of neural compression methods in this application domain, while our analysis elucidates trade-offs inherent in our approach, as well as practical considerations surrounding the design of error-bounded compression systems atop these methods.

\newpage
\section*{Acknowledgements}
The authors wish to thank Tom Andersson and Peter Battaglia for crucial feedback on the paper, David Ding and Maja Trębacz for authoring the reference implementations of VQ-VAE and VQ-GAN which served as a starting point for ours, and Jacklynn Stott and Jon Small for management and legal support. We are grateful to Langwen Huang for help in reproducting results from \citep{huang2022compressing}. We also thank Stephan Hoyer, Dmitrii Kochkov, Janni Yuval, Veronika Eyring and Pierre Gentine for helpful discussions.

\section*{Author Contributions}
PM, MKG, YH, DWF and MR planned and ran the main experiments discussed in the paper.
PM and YH analysed the experimental results.
MiR, DWF, SR and MeR wrote code and ran preliminary experiments on VQ models.
HK implemented code and ran preliminary experiments on factorized prior and hyperprior models.
DWF, PM and MiR wrote the results analysis pipeline.
SR, PM, DWF and MeR wrote the ERA5 data ingestion and projection.
DWF and MiR optimised the code and data storage infrastructure and enabled large-scale experiments.
DWF managed data governance and compliance.
YH reproduced baseline results and ran the bounded compression experiments.
MKG implemented code and ran experiments on functional priors (not reported in the paper).
PM and MeR implemented and ran experiments on additional projections (not reported in the paper).
MKG, PM, SM, YH, DWF and SR wrote the paper.
SM, SR, PM, and SO advised the project. 
SM, SR, and PM originated and managed the project.

\newpage
\bibliography{climate_compression}

\begin{thebibliography}{69}
\providecommand{\natexlab}[1]{#1}
\providecommand{\url}[1]{\texttt{#1}}
\expandafter\ifx\csname urlstyle\endcsname\relax
  \providecommand{\doi}[1]{doi: #1}\else
  \providecommand{\doi}{doi: \begingroup \urlstyle{rm}\Url}\fi

\bibitem[Ackley et~al.(1985)Ackley, Hinton, and Sejnowski]{ackley1985learning}
D.~H. Ackley, G.~E. Hinton, and T.~J. Sejnowski.
\newblock A learning algorithm for boltzmann machines.
\newblock \emph{Cognitive science}, 9\penalty0 (1):\penalty0 147--169, 1985.

\bibitem[Ashkboos et~al.(2022)Ashkboos, Huang, Dryden, Ben-Nun, Dueben,
  Gianinazzi, Kummer, and Hoefler]{ashkboos2022ens}
S.~Ashkboos, L.~Huang, N.~Dryden, T.~Ben-Nun, P.~Dueben, L.~Gianinazzi,
  L.~Kummer, and T.~Hoefler.
\newblock Ens-10: A dataset for post-processing ensemble weather forecasts.
\newblock \emph{Advances in Neural Information Processing Systems},
  35:\penalty0 21974--21987, 2022.

\bibitem[Ball\'{e} et~al.(2017)Ball\'{e}, Laparra, and Simoncelli]{balle2017fp}
J.~Ball\'{e}, V.~Laparra, and E.~P. Simoncelli.
\newblock End-to-end optimized image compression.
\newblock In \emph{Int'l Conf on Learning Representations (ICLR)}, Toulon,
  France, April 2017.
\newblock URL \url{https://arxiv.org/abs/1611.01704}.
\newblock Available at http://arxiv.org/abs/1611.01704.

\bibitem[Ball{\'{e}} et~al.(2018)Ball{\'{e}}, Minnen, Singh, Hwang, and
  Johnston]{balle2018hp}
J.~Ball{\'{e}}, D.~Minnen, S.~Singh, S.~J. Hwang, and N.~Johnston.
\newblock Variational image compression with a scale hyperprior.
\newblock In \emph{6th International Conference on Learning Representations,
  {ICLR} 2018, Vancouver, BC, Canada, April 30 - May 3, 2018, Conference Track
  Proceedings}. OpenReview.net, 2018.
\newblock URL \url{https://openreview.net/forum?id=rkcQFMZRb}.

\bibitem[Ballester-Ripoll et~al.(2020)Ballester-Ripoll, Lindstrom, and
  Pajarola]{tthresh_2020}
R.~Ballester-Ripoll, P.~Lindstrom, and R.~Pajarola.
\newblock {TTHRESH}: Tensor compression for multidimensional visual data.
\newblock \emph{IEEE Transactions on Visualization and Computer Graphics},
  2020.

\bibitem[Bengio et~al.(2013)Bengio, L{\'e}onard, and
  Courville]{bengio2013estimating}
Y.~Bengio, N.~L{\'e}onard, and A.~Courville.
\newblock Estimating or propagating gradients through stochastic neurons for
  conditional computation.
\newblock \emph{arXiv preprint arXiv:1308.3432}, 2013.

\bibitem[{Blosc Development Team}(2009-2023)]{blosc}
{Blosc Development Team}.
\newblock {A fast, compressed and persistent data store library}, 2009-2023.
\newblock URL \url{https://blosc.org}.

\bibitem[Burtscher and Ratanaworabhan(2009)]{fpc_2009}
M.~Burtscher and P.~Ratanaworabhan.
\newblock Fpc: A high-speed compressor for double-precision floating-point
  data.
\newblock \emph{IEEE Transactions on Computers}, 2009.

\bibitem[Chang et~al.(2023)Chang, Lee, Fu, and Tang]{chang2023seamless}
A.~Chang, H.~Lee, R.~Fu, and Q.~Tang.
\newblock A seamless approach for evaluating climate models across spatial
  scales.
\newblock \emph{Frontiers in Earth Science}, 11, 2023.
\newblock ISSN 2296-6463.
\newblock \doi{10.3389/feart.2023.1245815}.
\newblock URL
  \url{https://www.frontiersin.org/journals/earth-science/articles/10.3389/feart.2023.1245815}.

\bibitem[Chantry et~al.(2021)Chantry, Christensen, Dueben, and
  Palmer]{chantry2021opportunities}
M.~Chantry, H.~Christensen, P.~Dueben, and T.~Palmer.
\newblock Opportunities and challenges for machine learning in weather and
  climate modelling: hard, medium and soft ai.
\newblock \emph{Philosophical Transactions of the Royal Society A},
  379\penalty0 (2194):\penalty0 20200083, 2021.

\bibitem[Collet et~al.(2015)Collet, Terrell, Skibinski, Handte, Huang, Cruz,
  and {other contributors}]{zstd}
Y.~Collet, N.~Terrell, P.~Skibinski, F.~Handte, S.~Huang, P.~Cruz, and {other
  contributors}.
\newblock Zstandard, zstd - fast lossless compression algorithm, 2015.
\newblock URL \url{https://github.com/facebook/zstd}.

\bibitem[Cottrell et~al.(1987)Cottrell, Munro, and Zipser]{Cottrell1987}
G.~W. Cottrell, P.~Munro, and D.~Zipser.
\newblock Learning internal representations from gray-scale images: An example
  of extensional programming.
\newblock In \emph{Proceedings of the Ninth Annual Conference of the Cognitive
  Science Society}, pages 461--473, Seattle, WA, 1987.

\bibitem[Côté et~al.(1998)Côté, Gravel, Méthot, Patoine, Roch, and
  Staniforth]{cote1998gem}
J.~Côté, S.~Gravel, A.~Méthot, A.~Patoine, M.~Roch, and A.~Staniforth.
\newblock The operational cmc–mrb global environmental multiscale (gem)
  model. part i: Design considerations and formulation.
\newblock \emph{Monthly Weather Review}, 126\penalty0 (6):\penalty0 1373 --
  1395, 1998.
\newblock \doi{10.1175/1520-0493(1998)126<1373:TOCMGE>2.0.CO;2}.
\newblock URL
  \url{https://journals.ametsoc.org/view/journals/mwre/126/6/1520-0493_1998_126_1373_tocmge_2.0.co_2.xml}.

\bibitem[Di~Girolamo et~al.(2019)Di~Girolamo, Schmid, Schulthess, and
  Hoefler]{simfs_girolamo_2019}
S.~Di~Girolamo, P.~Schmid, T.~C. Schulthess, and T.~Hoefler.
\newblock Simfs: A simulation data virtualizing file system interface.
\newblock \emph{IEEE International Parallel and Distributed Processing
  Symposium}, 2019.

\bibitem[Dupont et~al.(2022{\natexlab{a}})Dupont, Kim, Eslami, Rezende, and
  Rosenbaum]{dupont2022functa}
E.~Dupont, H.~Kim, S.~M.~A. Eslami, D.~Rezende, and D.~Rosenbaum.
\newblock From data to functa: Your data point is a function and you can treat
  it like one, 2022{\natexlab{a}}.

\bibitem[Dupont et~al.(2022{\natexlab{b}})Dupont, Loya, Alizadeh, Golinski,
  Teh, and Doucet]{coinplusplus_2022}
E.~Dupont, H.~Loya, M.~Alizadeh, A.~Golinski, Y.~W. Teh, and A.~Doucet.
\newblock {COIN++:} data agnostic neural compression.
\newblock \emph{Transactions on Machine Learning Research}, 2022{\natexlab{b}}.
\newblock URL \url{https://arxiv.org/abs/2201.12904}.

\bibitem[Elman and Zipser(1988)]{elman1988learning}
J.~L. Elman and D.~Zipser.
\newblock Learning the hidden structure of speech.
\newblock \emph{The Journal of the Acoustical Society of America}, 83\penalty0
  (4):\penalty0 1615--1626, 1988.

\bibitem[Esser et~al.(2021)Esser, Rombach, and Ommer]{esser2021taming}
P.~Esser, R.~Rombach, and B.~Ommer.
\newblock Taming transformers for high-resolution image synthesis.
\newblock In \emph{Proceedings of the IEEE/CVF conference on computer vision
  and pattern recognition}, pages 12873--12883, 2021.

\bibitem[Eyring et~al.(2016)Eyring, Bony, Meehl, Senior, Stevens, Stouffer, and
  Taylor]{eyring2016overview}
V.~Eyring, S.~Bony, G.~A. Meehl, C.~A. Senior, B.~Stevens, R.~J. Stouffer, and
  K.~E. Taylor.
\newblock Overview of the coupled model intercomparison project phase 6 (cmip6)
  experimental design and organization.
\newblock \emph{Geoscientific Model Development}, 9\penalty0 (5):\penalty0
  1937--1958, 2016.

\bibitem[Goodfellow et~al.(2014)Goodfellow, Pouget-Abadie, Mirza, Xu,
  Warde-Farley, Ozair, Courville, and Bengio]{goodfellow2014gan}
I.~Goodfellow, J.~Pouget-Abadie, M.~Mirza, B.~Xu, D.~Warde-Farley, S.~Ozair,
  A.~Courville, and Y.~Bengio.
\newblock Generative adversarial nets.
\newblock In Z.~Ghahramani, M.~Welling, C.~Cortes, N.~Lawrence, and
  K.~Weinberger, editors, \emph{Advances in Neural Information Processing
  Systems}, volume~27. Curran Associates, Inc., 2014.
\newblock URL
  \url{https://proceedings.neurips.cc/paper_files/paper/2014/file/5ca3e9b122f61f8f06494c97b1afccf3-Paper.pdf}.

\bibitem[Gorski et~al.(1999)Gorski, Wandelt, Hansen, Hivon, and
  Banday]{gorski1999healpix}
K.~M. Gorski, B.~D. Wandelt, F.~K. Hansen, E.~Hivon, and A.~J. Banday.
\newblock The healpix primer, 1999.

\bibitem[Goyal(2001)]{goyal2001theoretical}
V.~K. Goyal.
\newblock Theoretical foundations of transform coding.
\newblock \emph{IEEE Signal Processing Magazine}, 18\penalty0 (5):\penalty0
  9--21, 2001.

\bibitem[Gr{\"u}nwald(2007)]{grunwald2007minimum}
P.~D. Gr{\"u}nwald.
\newblock \emph{The minimum description length principle}.
\newblock MIT press, 2007.

\bibitem[Han et~al.(2024)Han, Guo, Xu, Bai, et~al.]{han2024cra5}
T.~Han, S.~Guo, W.~Xu, L.~Bai, et~al.
\newblock Cra5: Extreme compression of era5 for portable global climate and
  weather research via an efficient variational transformer.
\newblock \emph{arXiv preprint arXiv:2405.03376}, 2024.

\bibitem[Hersbach et~al.(2018)Hersbach, Bell, Berrisford, Biavati, Horányi,
  J., J., C., R., I., D., A., C., D., and J-N]{era5}
H.~Hersbach, B.~Bell, P.~Berrisford, G.~Biavati, A.~Horányi, M.~S. J., N.~J.,
  P.~C., R.~R., R.~I., S.~D., S.~A., S.~C., D.~D., and T.~J-N.
\newblock Era5 hourly data on single levels from 1940 to present, 2018.

\bibitem[Hersbach et~al.(2020)Hersbach, Bell, Berrisford, Hirahara,
  Hor{\'a}nyi, Mu{\~n}oz-Sabater, Nicolas, Peubey, Radu, Schepers,
  et~al.]{hersbach2020era5}
H.~Hersbach, B.~Bell, P.~Berrisford, S.~Hirahara, A.~Hor{\'a}nyi,
  J.~Mu{\~n}oz-Sabater, J.~Nicolas, C.~Peubey, R.~Radu, D.~Schepers, et~al.
\newblock The era5 global reanalysis.
\newblock \emph{Quarterly Journal of the Royal Meteorological Society},
  146\penalty0 (730):\penalty0 1999--2049, 2020.

\bibitem[Hinton et~al.(2012)Hinton, Srivastava, Swersky, Tieleman, and
  Mohamed]{hinton2012coursera}
G.~Hinton, N.~Srivastava, K.~Swersky, T.~Tieleman, and A.~Mohamed.
\newblock Coursera: Neural networks for machine learning.
\newblock \emph{Lecture 9c: Using noise as a regularizer}, 2012.

\bibitem[Hinton(1990)]{hinton1990connectionist}
G.~E. Hinton.
\newblock Connectionist learning procedures.
\newblock In \emph{Machine learning}, pages 555--610. Elsevier, 1990.

\bibitem[Hinton and Zemel(1993)]{hinton1993autoencoders}
G.~E. Hinton and R.~Zemel.
\newblock Autoencoders, minimum description length and helmholtz free energy.
\newblock \emph{Advances in neural information processing systems}, 6, 1993.

\bibitem[Hoyer et~al.(2023)Hoyer, Yuval, Kochkov, Langmore, Norgaard, Mooers,
  and Brenner]{hoyer2023neural}
S.~Hoyer, J.~Yuval, D.~Kochkov, I.~Langmore, P.~Norgaard, G.~Mooers, and M.~P.
  Brenner.
\newblock Neural general circulation models for weather and climate.
\newblock \emph{AGU23}, 2023.

\bibitem[Huang and Hoefler(2023)]{huang2022compressing}
L.~Huang and T.~Hoefler.
\newblock Compressing multidimensional weather and climate data into neural
  networks.
\newblock \emph{ICLR}, 2023.

\bibitem[Isola et~al.(2017)Isola, Zhu, Zhou, and Efros]{isola2017image}
P.~Isola, J.-Y. Zhu, T.~Zhou, and A.~A. Efros.
\newblock Image-to-image translation with conditional adversarial networks.
\newblock In \emph{Proceedings of the IEEE conference on computer vision and
  pattern recognition}, pages 1125--1134, 2017.

\bibitem[Karlbauer et~al.(2023)Karlbauer, Cresswell-Clay, Durran, Moreno,
  Kurth, and Butz]{karlbauer2023advancing}
M.~Karlbauer, N.~Cresswell-Clay, D.~R. Durran, R.~A. Moreno, T.~Kurth, and
  M.~V. Butz.
\newblock Advancing parsimonious deep learning weather prediction using the
  healpix mes.
\newblock \emph{Authorea Preprints}, 2023.

\bibitem[Kingma and Welling(2014)]{kingma2014vae}
D.~P. Kingma and M.~Welling.
\newblock Auto-encoding variational bayes, 2014.

\bibitem[Kl{\"o}wer et~al.(2021)Kl{\"o}wer, Razinger, Dominguez, D{\"u}ben, and
  Palmer]{klower2021compressing}
M.~Kl{\"o}wer, M.~Razinger, J.~J. Dominguez, P.~D. D{\"u}ben, and T.~N. Palmer.
\newblock Compressing atmospheric data into its real information content.
\newblock \emph{Nature Computational Science}, 1\penalty0 (11):\penalty0
  713--724, 2021.

\bibitem[Lam et~al.(2023)Lam, Sanchez-Gonzalez, Willson, Wirnsberger,
  Fortunato, Alet, Ravuri, Ewalds, Eaton-Rosen, Hu, Merose, Hoyer, Holland,
  Vinyals, Stott, Pritzel, Mohamed, and Battaglia]{lam2023graphcast}
R.~Lam, A.~Sanchez-Gonzalez, M.~Willson, P.~Wirnsberger, M.~Fortunato, F.~Alet,
  S.~Ravuri, T.~Ewalds, Z.~Eaton-Rosen, W.~Hu, A.~Merose, S.~Hoyer, G.~Holland,
  O.~Vinyals, J.~Stott, A.~Pritzel, S.~Mohamed, and P.~Battaglia.
\newblock Learning skillful medium-range global weather forecasting.
\newblock \emph{Science}, 382\penalty0 (6677):\penalty0 1416--1421, 2023.
\newblock \doi{10.1126/science.adi2336}.
\newblock URL \url{https://www.science.org/doi/abs/10.1126/science.adi2336}.

\bibitem[LeCun et~al.(1989)]{lecun1989generalization}
Y.~LeCun et~al.
\newblock Generalization and network design strategies.
\newblock \emph{Connectionism in perspective}, 19\penalty0 (143-155):\penalty0
  18, 1989.

\bibitem[Li et~al.(2023)Li, Lindstrom, and Clyne]{sperr_2023}
S.~Li, P.~Lindstrom, and J.~Clyne.
\newblock Lossy scientific data compression with sperr.
\newblock In \emph{2023 IEEE International Parallel and Distributed Processing
  Symposium (IPDPS)}, 2023.

\bibitem[Liang et~al.(2022)Liang, Zhao, Di, Li, Underwood, Gok, Tian, Deng,
  Calhoun, Tao, Chen, and Cappello]{sz3_framework}
X.~Liang, K.~Zhao, S.~Di, S.~Li, R.~Underwood, A.~M. Gok, J.~Tian, J.~Deng,
  J.~C. Calhoun, D.~Tao, Z.~Chen, and F.~Cappello.
\newblock Sz3: A modular framework for composing prediction-based error-bounded
  lossy compressors.
\newblock \emph{IEEE Transactions on Big Data}, 2022.

\bibitem[Lin(1997)]{lin1997cubedsphere}
S.-J. Lin.
\newblock A finite-volume integration method for computing pressure gradient
  force in general vertical coordinates.
\newblock \emph{Quarterly Journal of the Royal Meteorological Society},
  123\penalty0 (542):\penalty0 1749--1762, 1997.
\newblock \doi{https://doi.org/10.1002/qj.49712354214}.
\newblock URL
  \url{https://rmets.onlinelibrary.wiley.com/doi/abs/10.1002/qj.49712354214}.

\bibitem[Lindstrom(2014)]{zfpy_2014}
P.~Lindstrom.
\newblock Fixed-rate compressed floating-point arrays.
\newblock \emph{IEEE Transactions on Visualization and Computer Graphics},
  2014.

\bibitem[Lindstrom and Isenburg(2006)]{fpzip_2006}
P.~Lindstrom and M.~Isenburg.
\newblock Fast and efficient compression of floating-point data.
\newblock \emph{IEEE Transactions on Visualization and Computer Graphics},
  2006.

\bibitem[Neumann et~al.(2019)Neumann, D{\"u}ben, Adamidis, Bauer, Br{\"u}ck,
  Kornblueh, Klocke, Stevens, Wedi, and Biercamp]{neumann2019assessing}
P.~Neumann, P.~D{\"u}ben, P.~Adamidis, P.~Bauer, M.~Br{\"u}ck, L.~Kornblueh,
  D.~Klocke, B.~Stevens, N.~Wedi, and J.~Biercamp.
\newblock Assessing the scales in numerical weather and climate predictions:
  will exascale be the rescue?
\newblock \emph{Philosophical Transactions of the Royal Society A},
  377\penalty0 (2142):\penalty0 20180148, 2019.

\bibitem[Nguyen et~al.(2023)Nguyen, Brandstetter, Kapoor, Gupta, and
  Grover]{nguyen2023climax}
T.~Nguyen, J.~Brandstetter, A.~Kapoor, J.~K. Gupta, and A.~Grover.
\newblock Climax: A foundation model for weather and climate.
\newblock \emph{ICLR}, 2023.

\bibitem[Ohno and Kageyama(2009)]{ohno2009visualization}
N.~Ohno and A.~Kageyama.
\newblock Visualization of spherical data by yin--yang grid.
\newblock \emph{Computer Physics Communications}, 180\penalty0 (9):\penalty0
  1534--1538, 2009.

\bibitem[Perez et~al.(2018)Perez, Strub, De~Vries, Dumoulin, and
  Courville]{perez2018film}
E.~Perez, F.~Strub, H.~De~Vries, V.~Dumoulin, and A.~Courville.
\newblock Film: Visual reasoning with a general conditioning layer.
\newblock In \emph{Proceedings of the AAAI conference on artificial
  intelligence}, volume~32, 2018.

\bibitem[Ramavajjala(2024)]{ramavajjala2024heal}
V.~Ramavajjala.
\newblock Heal-vit: Vision transformers on a spherical mesh for medium-range
  weather forecasting.
\newblock \emph{arXiv preprint arXiv:2403.17016}, 2024.

\bibitem[Ravuri et~al.(2021)Ravuri, Lenc, Willson, Kangin, Lam, Mirowski,
  Fitzsimons, Athanassiadou, Kashem, Madge, Prudden, Mandhane, Clark, Brock,
  Simonyan, Hadsell, Robinson, Clancy, and Mohamed]{ravuri2021nowcasting}
S.~Ravuri, K.~Lenc, M.~Willson, D.~Kangin, R.~Lam, P.~Mirowski, M.~Fitzsimons,
  M.~Athanassiadou, S.~Kashem, S.~Madge, R.~Prudden, A.~Mandhane, A.~Clark,
  A.~Brock, K.~Simonyan, R.~Hadsell, N.~Robinson, E.~Clancy, and A.~A. .~S.
  Mohamed.
\newblock Skilful precipitation nowcasting using deep generative models of
  radar.
\newblock \emph{Nature}, 597:\penalty0 672 – 677, 2021.

\bibitem[Rezende et~al.(2014)Rezende, Mohamed, and
  Wierstra]{rezende2014reparameterizationtrick}
D.~J. Rezende, S.~Mohamed, and D.~Wierstra.
\newblock Stochastic backpropagation and approximate inference in deep
  generative models, 2014.

\bibitem[Ronchi et~al.(1996)Ronchi, Iacono, and Paolucci]{ronchi1996cubed}
C.~Ronchi, R.~Iacono, and P.~S. Paolucci.
\newblock The “cubed sphere”: A new method for the solution of partial
  differential equations in spherical geometry.
\newblock \emph{Journal of computational physics}, 124\penalty0 (1):\penalty0
  93--114, 1996.

\bibitem[Satoh et~al.(2019)Satoh, Stevens, Judt, Khairoutdinov, Lin, Putman,
  and D{\"u}ben]{satoh2019global}
M.~Satoh, B.~Stevens, F.~Judt, M.~Khairoutdinov, S.-J. Lin, W.~M. Putman, and
  P.~D{\"u}ben.
\newblock Global cloud-resolving models.
\newblock \emph{Current Climate Change Reports}, 5:\penalty0 172--184, 2019.

\bibitem[Schneider et~al.(2017)Schneider, Lan, Stuart, and
  Teixeira]{schneider2017earth}
T.~Schneider, S.~Lan, A.~Stuart, and J.~Teixeira.
\newblock Earth system modeling 2.0: A blueprint for models that learn from
  observations and targeted high-resolution simulations.
\newblock \emph{Geophysical Research Letters}, 44\penalty0 (24):\penalty0
  12--396, 2017.

\bibitem[Schneider et~al.(2023)Schneider, Behera, Boccaletti, Deser, Emanuel,
  Ferrari, Leung, Lin, M{\"u}ller, Navarra, et~al.]{schneider2023harnessing}
T.~Schneider, S.~Behera, G.~Boccaletti, C.~Deser, K.~Emanuel, R.~Ferrari, L.~R.
  Leung, N.~Lin, T.~M{\"u}ller, A.~Navarra, et~al.
\newblock Harnessing ai and computing to advance climate modelling and
  prediction.
\newblock \emph{nature climate change}, 13\penalty0 (9):\penalty0 887--889,
  2023.

\bibitem[Smolarkiewicz et~al.(2016)Smolarkiewicz, Deconinck, Hamrud, Kühnlein,
  Mozdzynski, Szmelter, and Wedi]{smolarkiewicz2016fvm}
P.~K. Smolarkiewicz, W.~Deconinck, M.~Hamrud, C.~Kühnlein, G.~Mozdzynski,
  J.~Szmelter, and N.~P. Wedi.
\newblock A finite-volume module for simulating global all-scale atmospheric
  flows.
\newblock \emph{Journal of Computational Physics}, 314:\penalty0 287--304,
  2016.
\newblock ISSN 0021-9991.
\newblock \doi{https://doi.org/10.1016/j.jcp.2016.03.015}.
\newblock URL
  \url{https://www.sciencedirect.com/science/article/pii/S0021999116001674}.

\bibitem[Stevens et~al.(2020)Stevens, Acquistapace, Hansen, Heinze, Klinger,
  Klocke, Rybka, Schubotz, Windmiller, Adamidis, et~al.]{stevens2020added}
B.~Stevens, C.~Acquistapace, A.~Hansen, R.~Heinze, C.~Klinger, D.~Klocke,
  H.~Rybka, W.~Schubotz, J.~Windmiller, P.~Adamidis, et~al.
\newblock The added value of large-eddy and storm-resolving models for
  simulating clouds and precipitation.
\newblock \emph{Journal of the Meteorological Society of Japan. Ser. II},
  98\penalty0 (2):\penalty0 395--435, 2020.

\bibitem[Stevens et~al.(2024)Stevens, Adami, Ali, Anzt, Aslan, Attinger,
  B\"ack, Baehr, Bauer, Bernier, Bishop, Bockelmann, Bony, Brasseur, Bresch,
  Breyer, Brunet, Buttigieg, Cao, Castet, Cheng, Dey~Choudhury, Coen, Crewell,
  Dabholkar, Dai, Doblas-Reyes, Durran, El~Gaidi, Ewen, Exarchou, Eyring,
  Falkinhoff, Farrell, Forster, Frassoni, Frauen, Fuhrer, Gani, Gerber,
  Goldfarb, Grieger, Gruber, Hazeleger, Herken, Hewitt, Hoefler, Hsu, Jacob,
  Jahn, Jakob, Jung, Kadow, Kang, Kang, Kashinath, Kleinen-von K\"onigsl\"ow,
  Klocke, Kloenne, Kl\"ower, Kodama, Kollet, K\"olling, Kontkanen, Kopp, Koran,
  Kulmala, Lappalainen, Latifi, Lawrence, Lee, Lejeun, Lessig, Li, Lippert,
  Luterbacher, Manninen, Marotzke, Matsouoka, Merchant, Messmer, Michel,
  Michielsen, Miyakawa, M\"uller, Munir, Narayanasetti, Ndiaye, Nobre, Oberg,
  Oki, \"Ozkan-Haller, Palmer, Posey, Prein, Primus, Pritchard, Pullen,
  Putrasahan, Quaas, Raghavan, Ramaswamy, Rapp, Rauser, Reichstein, Revi,
  Saluja, Satoh, Schemann, Schemm, Schnadt~Poberaj, Schulthess, Senior, Shukla,
  Singh, Slingo, Sobel, Solman, Spitzer, Stier, Stocker, Strock, Su, Taalas,
  Taylor, Tegtmeier, Teutsch, Tompkins, Ulbrich, Vidale, Wu, Xu, Zaki, Zanna,
  Zhou, and Ziemen]{essd-16-2113-2024}
B.~Stevens, S.~Adami, T.~Ali, H.~Anzt, Z.~Aslan, S.~Attinger, J.~B\"ack,
  J.~Baehr, P.~Bauer, N.~Bernier, B.~Bishop, H.~Bockelmann, S.~Bony,
  G.~Brasseur, D.~N. Bresch, S.~Breyer, G.~Brunet, P.~L. Buttigieg, J.~Cao,
  C.~Castet, Y.~Cheng, A.~Dey~Choudhury, D.~Coen, S.~Crewell, A.~Dabholkar,
  Q.~Dai, F.~Doblas-Reyes, D.~Durran, A.~El~Gaidi, C.~Ewen, E.~Exarchou,
  V.~Eyring, F.~Falkinhoff, D.~Farrell, P.~M. Forster, A.~Frassoni, C.~Frauen,
  O.~Fuhrer, S.~Gani, E.~Gerber, D.~Goldfarb, J.~Grieger, N.~Gruber,
  W.~Hazeleger, R.~Herken, C.~Hewitt, T.~Hoefler, H.-H. Hsu, D.~Jacob, A.~Jahn,
  C.~Jakob, T.~Jung, C.~Kadow, I.-S. Kang, S.~Kang, K.~Kashinath,
  K.~Kleinen-von K\"onigsl\"ow, D.~Klocke, U.~Kloenne, M.~Kl\"ower, C.~Kodama,
  S.~Kollet, T.~K\"olling, J.~Kontkanen, S.~Kopp, M.~Koran, M.~Kulmala,
  H.~Lappalainen, F.~Latifi, B.~Lawrence, J.~Y. Lee, Q.~Lejeun, C.~Lessig,
  C.~Li, T.~Lippert, J.~Luterbacher, P.~Manninen, J.~Marotzke, S.~Matsouoka,
  C.~Merchant, P.~Messmer, G.~Michel, K.~Michielsen, T.~Miyakawa, J.~M\"uller,
  R.~Munir, S.~Narayanasetti, O.~Ndiaye, C.~Nobre, A.~Oberg, R.~Oki,
  T.~\"Ozkan-Haller, T.~Palmer, S.~Posey, A.~Prein, O.~Primus, M.~Pritchard,
  J.~Pullen, D.~Putrasahan, J.~Quaas, K.~Raghavan, V.~Ramaswamy, M.~Rapp,
  F.~Rauser, M.~Reichstein, A.~Revi, S.~Saluja, M.~Satoh, V.~Schemann,
  S.~Schemm, C.~Schnadt~Poberaj, T.~Schulthess, C.~Senior, J.~Shukla, M.~Singh,
  J.~Slingo, A.~Sobel, S.~Solman, J.~Spitzer, P.~Stier, T.~Stocker, S.~Strock,
  H.~Su, P.~Taalas, J.~Taylor, S.~Tegtmeier, G.~Teutsch, A.~Tompkins,
  U.~Ulbrich, P.-L. Vidale, C.-M. Wu, H.~Xu, N.~Zaki, L.~Zanna, T.~Zhou, and
  F.~Ziemen.
\newblock Earth virtualization engines (eve).
\newblock \emph{Earth System Science Data}, 16\penalty0 (4):\penalty0
  2113--2122, 2024.
\newblock \doi{10.5194/essd-16-2113-2024}.
\newblock URL \url{https://essd.copernicus.org/articles/16/2113/2024/}.

\bibitem[Theis et~al.(2017)Theis, Shi, Cunningham, and
  Husz{\'a}r]{theis2017lossy}
L.~Theis, W.~Shi, A.~Cunningham, and F.~Husz{\'a}r.
\newblock Lossy image compression with compressive autoencoders.
\newblock In \emph{International conference on learning representations}, 2017.

\bibitem[{UK Met Office}(2016)]{metoffice2016data}
{UK Met Office}.
\newblock Nwp-ukv: Met office uk atmospheric high resolution model data, 2016.
\newblock URL
  \url{https://catalogue.ceda.ac.uk/uuid/f47bc62786394626b665e23b658d385f}.

\bibitem[van~den Oord et~al.(2017)van~den Oord, Vinyals, and
  Kavukcuoglu]{oord2017vqvae}
A.~van~den Oord, O.~Vinyals, and K.~Kavukcuoglu.
\newblock Neural discrete representation learning.
\newblock \emph{CoRR}, abs/1711.00937, 2017.
\newblock URL \url{http://arxiv.org/abs/1711.00937}.

\bibitem[Vaswani et~al.(2017)Vaswani, Shazeer, Parmar, Uszkoreit, Jones, Gomez,
  Kaiser, and Polosukhin]{vaswani2017transformer}
A.~Vaswani, N.~Shazeer, N.~Parmar, J.~Uszkoreit, L.~Jones, A.~N. Gomez, L.~u.
  Kaiser, and I.~Polosukhin.
\newblock Attention is all you need.
\newblock In I.~Guyon, U.~V. Luxburg, S.~Bengio, H.~Wallach, R.~Fergus,
  S.~Vishwanathan, and R.~Garnett, editors, \emph{Advances in Neural
  Information Processing Systems}, volume~30. Curran Associates, Inc., 2017.
\newblock URL
  \url{https://proceedings.neurips.cc/paper_files/paper/2017/file/3f5ee243547dee91fbd053c1c4a845aa-Paper.pdf}.

\bibitem[Wallace(1992)]{wallace1992jpeg}
G.~K. Wallace.
\newblock The jpeg still picture compression standard.
\newblock \emph{IEEE transactions on consumer electronics}, 38\penalty0
  (1):\penalty0 xviii--xxxiv, 1992.

\bibitem[Williams(1992)]{williams1992simple}
R.~J. Williams.
\newblock Simple statistical gradient-following algorithms for connectionist
  reinforcement learning.
\newblock \emph{Machine learning}, 8:\penalty0 229--256, 1992.

\bibitem[Willmert(2019)]{willmert2019constraining}
J.~Willmert.
\newblock \emph{Constraining inflationary B-modes with the BICEP/Keck array
  telescopes}.
\newblock PhD thesis, University of Minnesota, 2019.

\bibitem[Willmert(2020)]{willmert2020notes}
J.~Willmert.
\newblock Notes on spherical harmonics series, 2020.
\newblock URL
  \url{https://justinwillmert.com/articles/2020/notes-on-calculating-the-spherical-harmonics/}.

\bibitem[Willmert(2022)]{willmert2022ring}
J.~Willmert.
\newblock Spherical harmonic transforms on ring-based pixelizations, 2022.
\newblock URL
  \url{https://justinwillmert.com/articles/2022/spherical-harmonic-transforms-on-ring-based-pixelizations/}.

\bibitem[Xie et~al.(2022)Xie, Takikawa, Saito, Litany, Yan, Khan, Tombari,
  Tompkin, Sitzmann, and Sridhar]{xie2022neural_fields_beyond}
Y.~Xie, T.~Takikawa, S.~Saito, O.~Litany, S.~Yan, N.~Khan, F.~Tombari,
  J.~Tompkin, V.~Sitzmann, and S.~Sridhar.
\newblock Neural fields in visual computing and beyond.
\newblock \emph{Computer Graphics Forum}, 2022.

\bibitem[Zhao et~al.(2020)Zhao, Di, Liang, Li, Tao, Bessac, Chen, and
  Cappello]{sdrbench_2020}
K.~Zhao, S.~Di, X.~Liang, S.~Li, D.~Tao, J.~Bessac, Z.~Chen, and F.~Cappello.
\newblock Sdrbench: Scientific data reduction benchmark for lossy compressors.
\newblock \emph{International Workshop on Big Data Reduction}, 2020.

\bibitem[Zhao et~al.(2021)Zhao, Dmitriev, Tonellot, Chen, and
  Cappello]{sz3_algo}
S.~Zhao, Kaiand~Di, M.~Dmitriev, T.-L.~D. Tonellot, Z.~Chen, and F.~Cappello.
\newblock Optimizing error-bounded lossy compression for scientiﬁc data by
  dynamic spline interpolation.
\newblock \emph{IEEE International Conference on Data Engineering}, 2021.

\bibitem[Zängl et~al.(2015)Zängl, Reinert, Rípodas, and
  Baldauf]{zangl2015icon}
G.~Zängl, D.~Reinert, P.~Rípodas, and M.~Baldauf.
\newblock The icon (icosahedral non-hydrostatic) modelling framework of dwd and
  mpi-m: Description of the non-hydrostatic dynamical core.
\newblock \emph{Quarterly Journal of the Royal Meteorological Society},
  141\penalty0 (687):\penalty0 563--579, 2015.
\newblock \doi{https://doi.org/10.1002/qj.2378}.
\newblock URL
  \url{https://rmets.onlinelibrary.wiley.com/doi/abs/10.1002/qj.2378}.

\end{thebibliography}

\newpage
\appendix
\appendixpage

The following Appendices detail the HEALPix projection and its overset grid version (Appendix \ref{sx:appendix:overset_healpix}), the use of spherical harmonics for re-projection and power spectral density estimation (Appendix \ref{sx:appendix:spherical_harmonics}), details about the VQ-VAE, VQ-GAN and Hyper Prior models (Appendix \ref{sx:appendix:models}), reproducing the neural compression baseline from \citep{huang2022compressing} (Appendix \ref{sx:appendix:huangiclr2023}), and an analysis of compression results that investigates whether reconstruction artefacts can be attributed to data artefacts (Appendix \ref{sx:appendix:analysis}).

\section{Overset HEALPix}
\label{sx:appendix:overset_healpix}
\subsection{HEALPix projection}

Throughout the paper, we rely on a specific instance of the HEALPix projection that projects the sphere onto 12 diamond-shaped tiles of equal areas: 4 tiles joining at the North Pole and touching the Equator on their opposite corners, 4 equatorial tiles, and 4 tiles joining at the South Pole and touching the Equator. The HEALPix projection is illustrated on Figure \ref{fig:system}.

Each tile can be projected onto a square and subdivided into $2^n \times 2^n$ pixels. In this paper, we choose $2^n = 256$ so that one HEALPix pixel covers an area roughly equivalent to a pixel on the Equator of a lat/lon grid of 721 latitudes by 1440 longitudes.
The area of a single HEALPix pixel on a sphere of radius $r$ is $4 \pi r^2 / (12 \times 2^n \times 2^n) \approx 1.6 \times 10^{-5} \times r^2$ for $2^n = 256$, whereas the area of a single latitude/longitude cell on a grid of $N_{lat}$ latitudes by $N_{lon}$ longitudes can be approximated to $(2 \pi r)^2 / (N_{lon} \times 2 N_{lat}) \approx 1.9 \times 10^{-5} \times r^2$. While such a HEALPix mesh contains fewer pixels (only $12 \times 2^n \times 2^n = 786,432$) as opposed to the lat/lon grid at 0.25$^\circ$ ($N_{lon} \times N_{lat} = 1,038,240$), the latter cells cover increasingly smaller areas close to the poles.

The coordinates of all pixels in HEALPix tiles are computed using the {\tt astropy\_healpix}\footnote{\url{https://github.com/astropy/astropy-healpix}} Python software library. Projection from the lat/lon grid onto pixel coordinates of individual HEALPix tiles is done using bilinear interpolation over a grid $(\cos(\theta), \phi)$ of cosines of latitudes $\theta$ and longitudes $\phi$.

For reprojecting back from HEALPix to lat/lon coordinates, and while we could have used bilinear interpolation on individual HEALPix tiles' square grids, or inverse distance interpolation on triangles from the 3D mesh of HEALPix coordinates, we use spherical harmonics synthesis instead, as described in Appendix \ref{sx:appendix:spherical_harmonics}.

\subsection{Stitching artefacts due to combining different HEALPix reconstructions}

\begin{figure}
    \centering
    \includegraphics[width=0.495\textwidth]{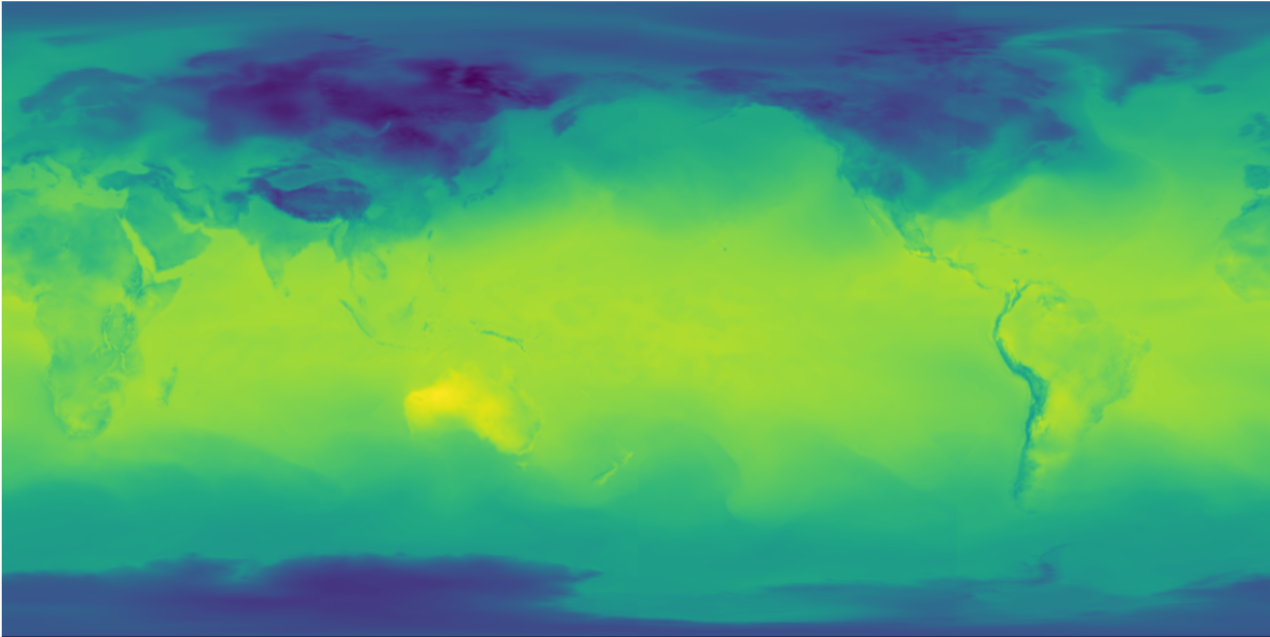}
    \includegraphics[width=0.495\textwidth]{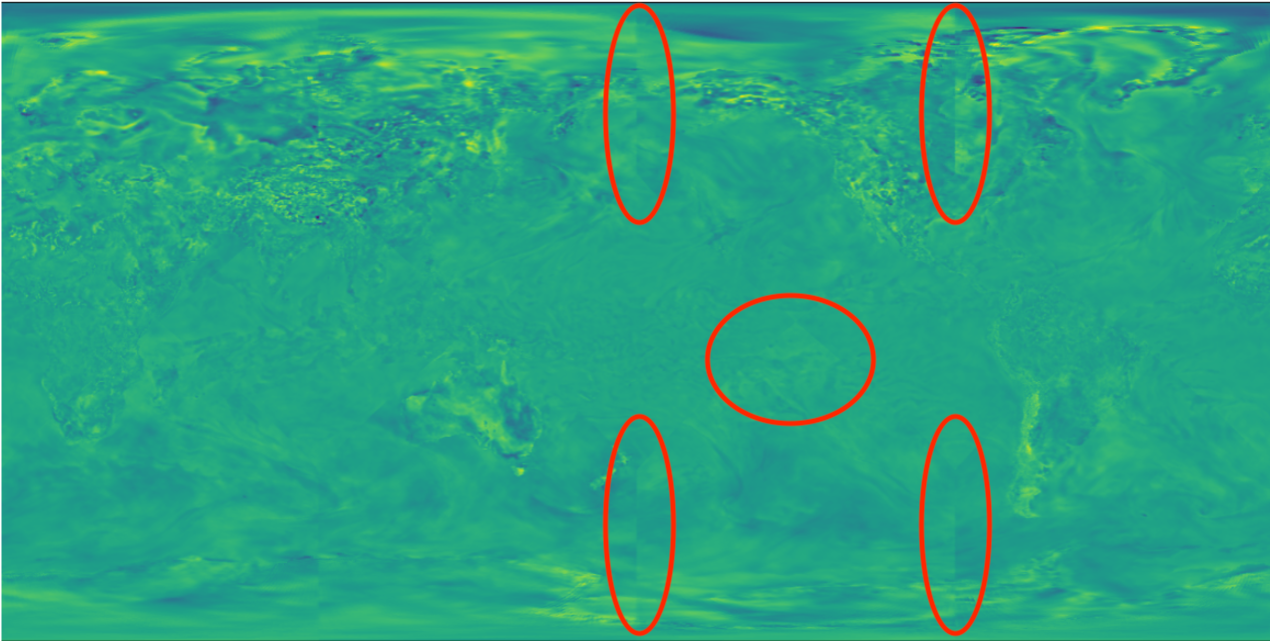}
    \hfill
    \caption{Example of ``stitching'' artefact due to reconstructing HEALPix tiles independently. The left image shows temperature at 2m after reconstruction in HEALPix projection space and reprojection back to lat/lon coordinates. The right image shows the difference between reconstruction and ground truth, making apparent that artefacts appear along the ``seamlines'' of different HEALPix tiles (highlighted in red ovals on the figure).}
    \label{fig:stitching}
\end{figure}

Our image compression models are trained to compress and reconstruct HEALPix tiles independently. As a consequence, small discontinuities can appear at the edges of each reconstructed HEALPix tile, and become apparent when neighboring HEALPix tiles are ``stitched'' together on the reprojected lat/lon image, as exemplified on Figure \ref{fig:stitching}. We hypothesize that discontinuities appear because at the edges of each HEALPix tile, the model misses contextual information from beyond each HEALPix tile.

\subsection{HEALPix tiles extended with padding}

To address the problem of discontinuites and stitching artefacts, we propose to learn image compression and reconstruction using overset HEALPix meshes and in extended image space, by going from $288 \times 288$ input images ($288 = 256 + 16 + 16$, i.e., 256 plus 16 padding borders on each side of the tile) to $288 \times 288$ output images. These resulting input, target and output images correspond to \emph{padded} HEALPix tiles.

Padding HEALPix tiles is complicated by the fact that some corners of HEALPix tiles are at the intersection of 4 tiles (the junctions at the Equator, and at the North and South Pole), whereas other corners are at the intersection of 3 tiles (the junction at $\pm 45^\circ$ latitude). \citet{karlbauer2023advancing} discuss padding HEALPix tiles by copying pixel values from neighbouring tiles (see Figure A2 in their paper) but their proposal relies on mirror-padding the polar tiles at the 3-tile junctions. Unlike their approach, ours preserves local angles by relying on interpolation instead and does not involve mirror-padding.

\begin{figure}
    \centering
    \includegraphics[width=0.24\textwidth]{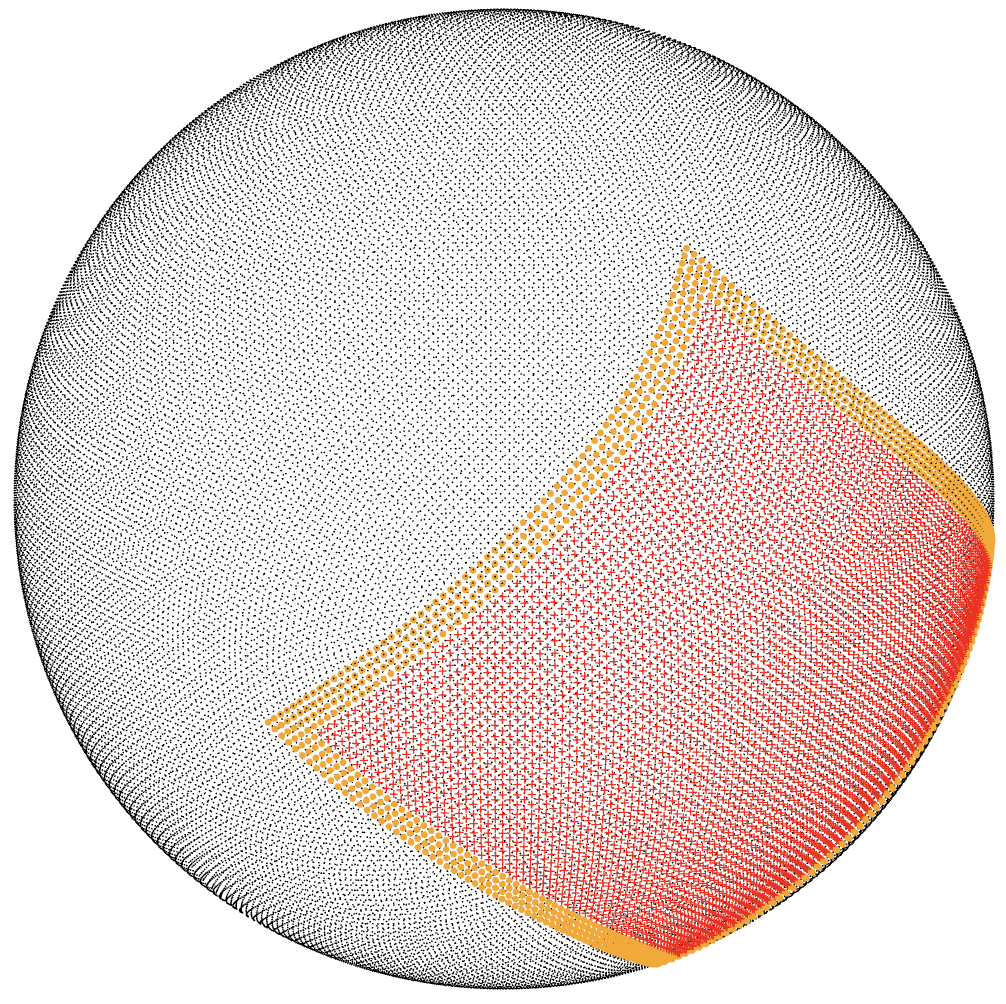}
    \includegraphics[width=0.24\textwidth]{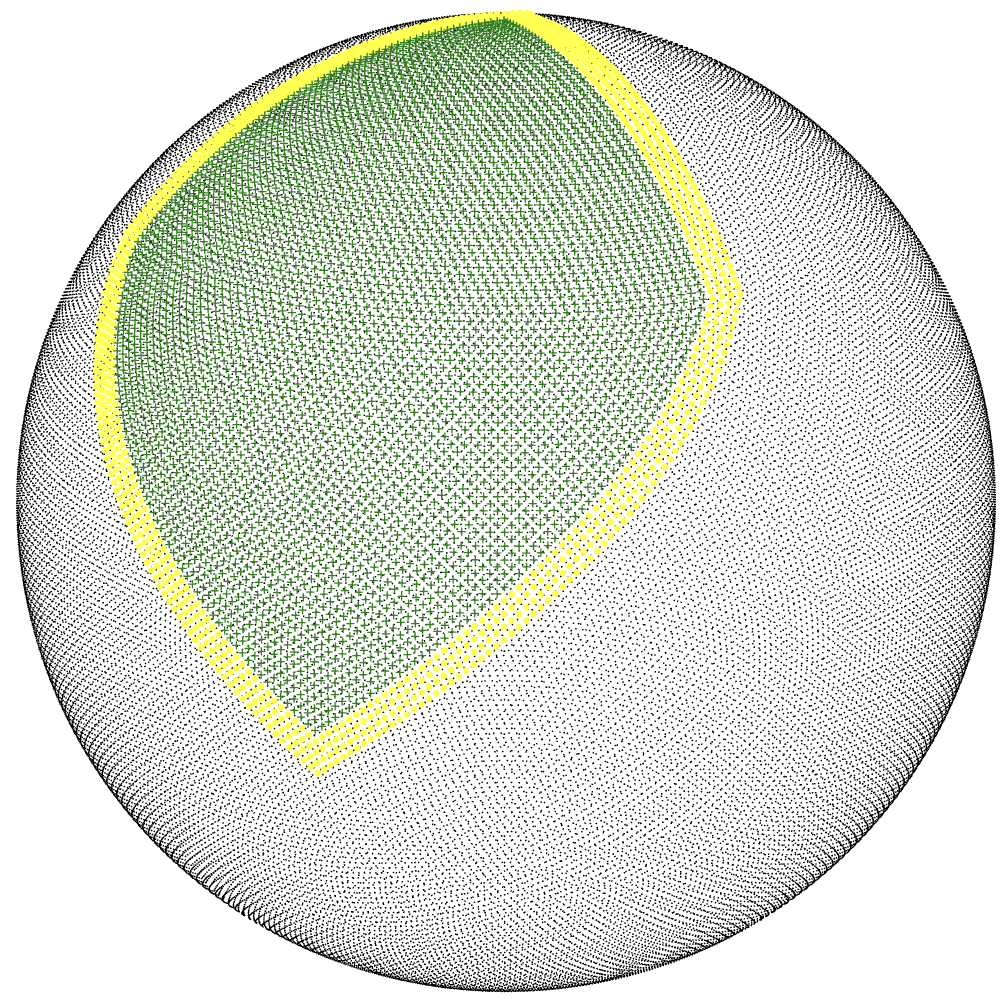}
    \includegraphics[width=0.24\textwidth]{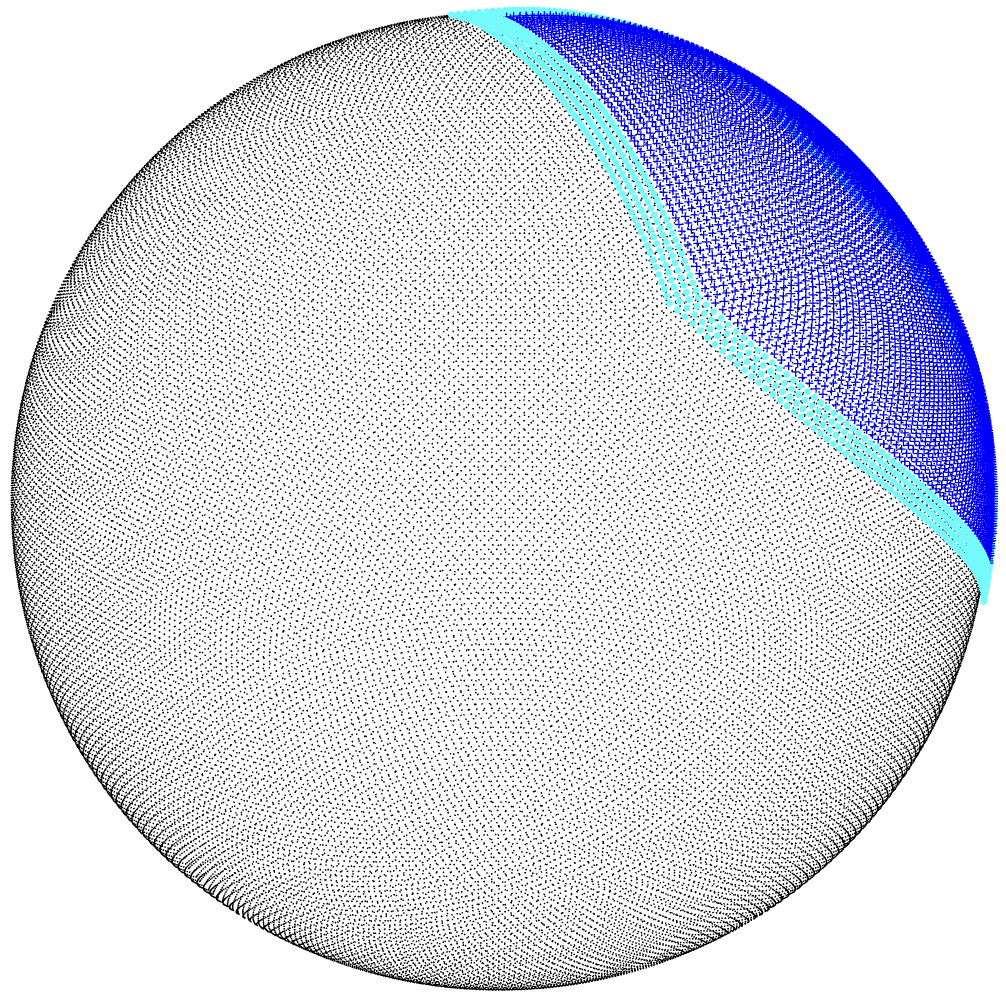}
    \includegraphics[width=0.24\textwidth]{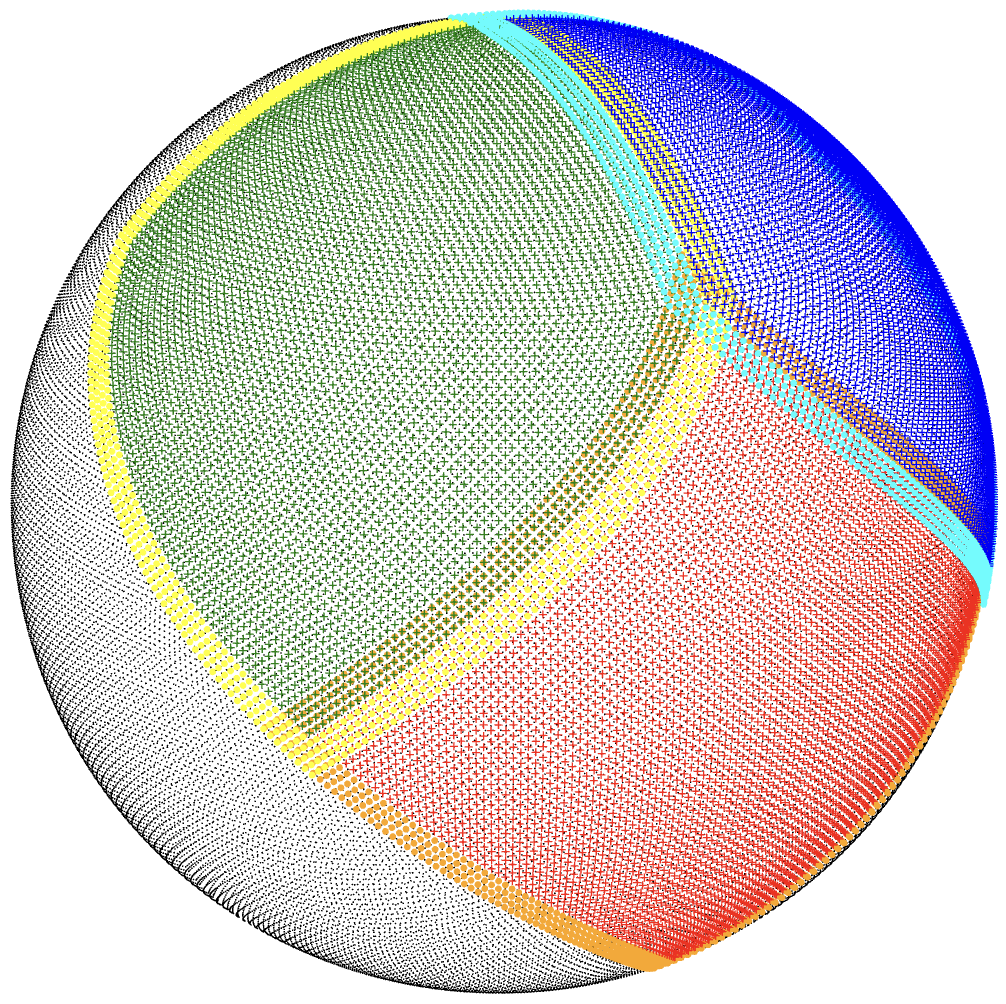}
    \hfill
    \caption{Illustration of HEALPix tiles 7 (red), 10 (green) and 11 (blue) with overset (so-called \emph{padding}) pixels on each side---respectively in orange, yellow and cyan for squares 7, 10, and 11---and composite figures with squares 7, 10 and 11 along with their padding superimposed onto a single globe. For clarity, we plot $64 \times 64$ HEALPix tiles with 2-pixel padding instead of $256 \times 256$ tiles with 16-pixel padding.}
    \label{fig:overset}
\end{figure}

Figure \ref{fig:overset} illustrates how we extend the $256 \times 256$ HEALPix tiles with extra padding of 16 pixels on each side. Note that unlike HEALPix coordinates, there is no closed form calculation for this padding; instead, we use extrapolation of HEALPix corners and interpolation between two such extrapolated corners to obtain a new $288 \times 288$ grid.

In order to blend the HEALPix tiles, we first compute the weight of the padding pixels, which decays linearly from 1 to 0 on each border.
Second, the coordinates of the padding borders of each HEALPix tile are computed in the Cartesian space of the sphere. Third, we then interpolate the effect of each HEALPix tile's overlap (say, red tile 7 on Fig. \ref{fig:overset}) upon each of its neighbors by computing the coordinates of the external tile in the target tile space (say, green tile 10 on Fig. \ref{fig:overset}) by projecting the Cartesian coordinates of tile 7 onto the HEALPix pixels of tile 10. The blended contribution of each HEALPix tile to its neighbours is computed using weighted sums. This results in new $256 \times 256$ HEALPix data, on which we compute spherical harmonics for re-projection onto the latitude/longitude grid.
 
\section{Spherical harmonics implementation}
\label{sx:appendix:spherical_harmonics}
Spherical harmonics are set of orthogonal functions; they form an orthonormal basis that allows to represent a 2D field defined on a sphere. Spherical harmonics functions are defined from degree 0 up to a maximum degree $l_{max}$, with orders $m$ such that $-l \leq m \leq l$ and $0 \leq l \leq l_{max}$.

At a given longitude (also called \emph{azimuthal} angle) $\phi$ and colatitude (also called \emph{polar} angle) $\theta$ (both expressed in radians), spherical harmonics function $Y_m^l(\theta, \phi)$ with degree $l$ and order $m$, and which is defined over complex range (with $i^2=-1$), is calculated as follows:

\begin{equation}
Y_m^l(\theta, \phi) \equiv \sqrt{\frac{2l+1}{4\pi} \frac{(l-m)!}{(l+m)!}}
 P_m^l(\cos(\theta))e^{im\phi}
\label{eq:sph_harm_fun}
\end{equation}

In Equation \ref{eq:sph_harm_fun}, $P_m^l(\cos(\theta))$ is the associated Legendre polynomial of order $m$, degree $l$ and defined over colatitudes $\theta$. We can separate the polar angles $\theta$ from the azimuthal angles $\phi$ by expressing Eq. \ref{eq:sph_harm_fun} as $Y_m^l(\theta, \phi) \equiv \lambda_m^l(\cos(\theta))e^{im\phi}$, where:

\begin{equation}
\lambda_m^l(\cos(\theta)) \equiv \sqrt{\frac{2l+1}{4\pi} \frac{(l-m)!}{(l+m)!}} P_m^l(\cos(\theta))
\label{eq:sph_harm_lambda}
\end{equation}

\subsection{Normalized associated Legendre polynomials}

Following \citet{willmert2020notes,willmert2019constraining}, we calculate the normalized associated Legendre polynomials using the following recurrence relations on $\lambda_m^l(\cos(\theta))$ in order to ensure numerical stability:

\begin{align}
\lambda_0^0(x) &=  \sqrt{\frac{1}{4\pi}} \\
\lambda_{l+1}^{l+1}(x) &= -\mu_{l+1} \sqrt{1-x^2}\lambda_l^l(x) \\
\lambda_l^{l+1}(x) &= \nu_{l+1} x \lambda_l^l(x) \\
\lambda_m^l(x) &= \alpha_m^l x \lambda_m^{l-1}(x) - \beta_m^l\lambda_m^{l-2}(x)
\end{align}

Where terms $\mu_l$, $\nu_l$, $\alpha_m^l$ and $\beta_m^l$ are defined as:

\begin{align}
\mu_l &= \sqrt{1 + \frac{1}{2l}} \\
\nu_l &= \sqrt{1+2l} \\
\alpha_m^l &= \sqrt{\frac{2l+1}{2l-3}\frac{4(l-1)^2-1}{l^2 - m^2}} \\
\beta_m^l &= \sqrt{\frac{2l+1}{2l-3}\frac{(l-1)^2-m^2}{l^2 - m^2}}
\end{align}

Note that for a given maximum order $l_{max}$ and a fixed set of polar angles $\theta$ (i.e., the set of colatitudes in the latitude/longitude data, or the unique set of colatitudes corresponding to the diagonals of HEALPix mesh coordinates), these polynomials need to be computed and stored only once. When developing the code for the Lagrange polynomial-based spherical harmonics transform, we matched our results to the standard SciPy implementation provided by {\tt scipy.special.sph\_harm}\footnote{\url{https://docs.scipy.org/doc/scipy/reference/generated/scipy.special.sph\_harm.html}}.

\subsection{Spherical harmonic transform (analysis)}

Spherical harmonic coefficients $a_{lm}$ are defined as
$a_{lm} = \int_{S} f(\theta, \phi) {Y_m^l}^*(\theta, \phi) d\Omega$ for a continuous field $f$. When using a set of $N$ pairs of longitude angles and colatitude angles $\{(\phi_i, \theta_i)\}_i$ covering a grid, their approximation becomes:

\begin{equation}
\hat{a}_{lm} = \sum_{i=1}^N f(\theta_i, \phi_i) \bar{Y}_{lm}(\theta_i, \phi_i) \sin(\theta_i)\Delta\theta_i\Delta\phi_i
\label{eq:sph_harm_forward}
\end{equation}

N.B. that if the signal $f$ is real, then: $a_{l(-m)} = (-1)^m \bar{a}_{lm}$, which entails that we only need to store the lower triangular $\mathbb{C} ^{(l+1) \times (l+1)}$ matrix of spherical harmonics coefficients.

Equation \ref{eq:sph_harm_forward} can be decomposed as a sum of sums, the first summation being over unique colatitudes $\{\theta_j\}_j$ and for each colatitude $\theta_j$, the nested summations over longitudes of the grid points at that constant colatitude (i.e., on a colatitude ring~\citep{willmert2022ring}). In the case of latitude/longitude grids, we sample $f(\theta, \phi)$ at $N_{lat}=721$ colatitudes with degree values (0, 0.25, ..., 179.75, 180) and at $N_{lon}=1440$ longitudes with degree values (0, 0.25, ..., 359.5, 359.75). In the case of HEALPix, we sample $f(\theta, \phi)$ at 1024 colatitude rings associated to the 512 off-diagonals of the northern polar HEALPix tiles and to the 512 off-diagonals of the southern polar HEALPix tiles (see \citet{willmert2022ring}).

If the $N_{lon}$ samples of $f(\theta,\phi)$ are regularly spaced on each colatitude ring, then the $N_{lat} \times N_{lon} \times {l_{max}+1} \times {l_{max}+1}$ multiplications in Equation \ref{eq:sph_harm_forward} can be replaced by $N_{lat} \times {l_{max}+1} \times {l_{max}+1}$ multiplications by using the Fast Fourier Transform of each ring.

\subsection{Reverse spherical harmonic transform (synthesis)}

The reverse spherical harmonic transform is:

\begin{equation}
\hat f(\theta, \phi) = \sum_{l=0}^{l_{max}}\sum_{m=-l}^l a_{lm}Y_m^l(\theta, \phi)
\label{eq:sph_harm_backward}
\end{equation}

From Equation \ref{eq:sph_harm_fun}, we can deduct that $Y_{-m}^l(\theta, \phi) = (-1)^m \bar{Y}_m^l(\theta, \phi)$, hence the real-valued synthesis $\hat f(\theta, \phi)$ can be written as:

\begin{equation}
\hat f(\theta, \phi) = {\tt Re} \left [ \sum_{l=0}^{l_{max}}\sum_{m=0}^l (2 - \delta_{m0}) a_{lm}Y_m^l(\theta, \phi) \right ]
\label{eq:sph_harm_backward_real}
\end{equation}

We benchmarked our forward and backward implementations by performing a round-trip analysis and synthesis of latitude/longitude data (and of HEALPix projections of latitude/longitude data) against the latitude/longitude spherical harmonics transform implementation in Python/JAX\footnote{\url{https://github.com/google-research/dinosaur}} that accompanies \citep{hoyer2023neural}.

\subsection{Power spectral density of spherical data using spherical harmonics}

Once we have spherical harmonics coefficients $\{a_{lm}\}_{l,m}$, we compute the power spectral density as:

\begin{equation}
I_l = \sum _{m=0}^{l-1} \frac{1}{2l + 1} | a_{lm} |^2
\end{equation}

\subsection{Reprojection of HEALPix data into latitude/longitude coordinates}

Spherical harmonics provide a simple mechanism for reprojecting HEALPix data into equirectangular latitude/longitude coordinates: namely, after analysing a HEALPix representation $f_{HEALPix}(\theta, \phi)$ into a set of spherical harmonics coefficients $\{a_{lm}\}_{l,m}$, we synthesize points $f(\theta,\phi)$ on a latitude/longitude grid simply by looping over $N_{lat}$ coordinates and calculating inverse FFT on $N_{lon}$-point equilateral and regularly spaced rings. 
\section{Detailed model diagrams}
\label{sx:appendix:models}
\begin{figure}
    \centering
    \begin{subfigure}[t]{.6\textwidth}
        \centering
        \includegraphics[height=2.5in]{figures/networks/vqgan.pdf}
        \caption{Overview\label{fig:vq:vqgan}}
    \end{subfigure}
    \hfill
    \begin{subfigure}[t]{.29\textwidth}
        \centering
        \includegraphics[height=2.5in]{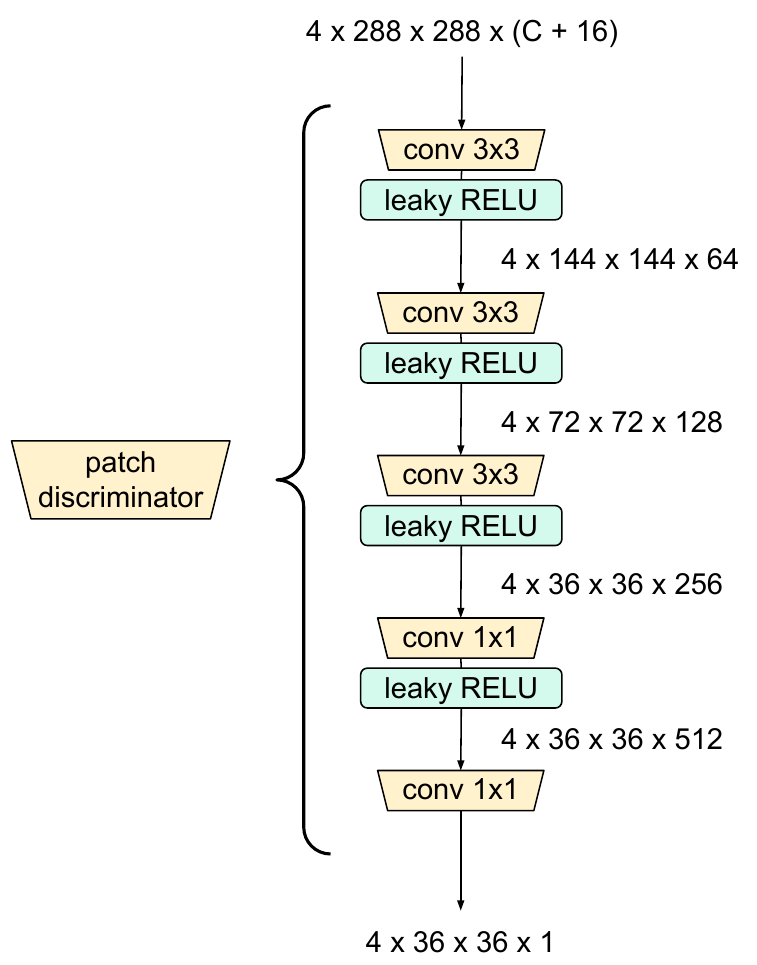}
        \caption{Patch discriminator\label{fig:vq:discriminator}}
    \end{subfigure}
    \par\bigskip \begin{subfigure}[c]{.3\textwidth}
        \centering
        \includegraphics[height=2in]{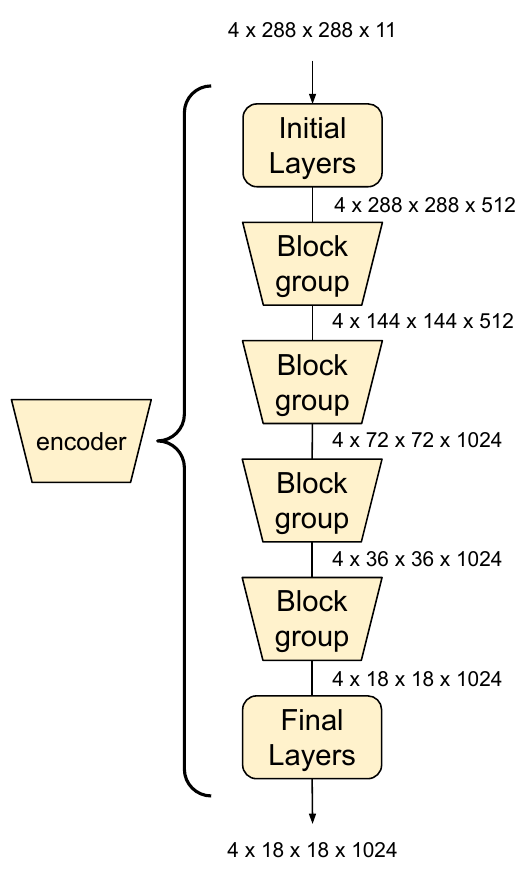}
        \caption{Encoder \label{fig:vq:enc}}
    \end{subfigure}
    \hfill
    \begin{subfigure}[c]{.3\textwidth}
        \centering
        \includegraphics[height=2in]{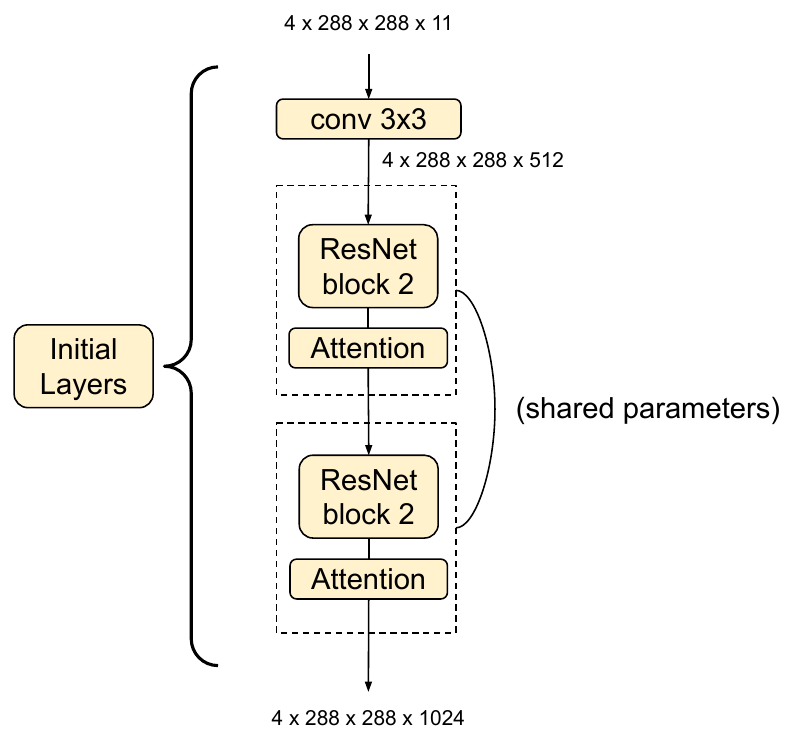}
        \caption{Initial layers \label{fig:vq:enc:initial_layers}}
    \end{subfigure}
    \hfill
    \begin{subfigure}[c]{.3\textwidth}
        \centering
        \includegraphics[height=2in]{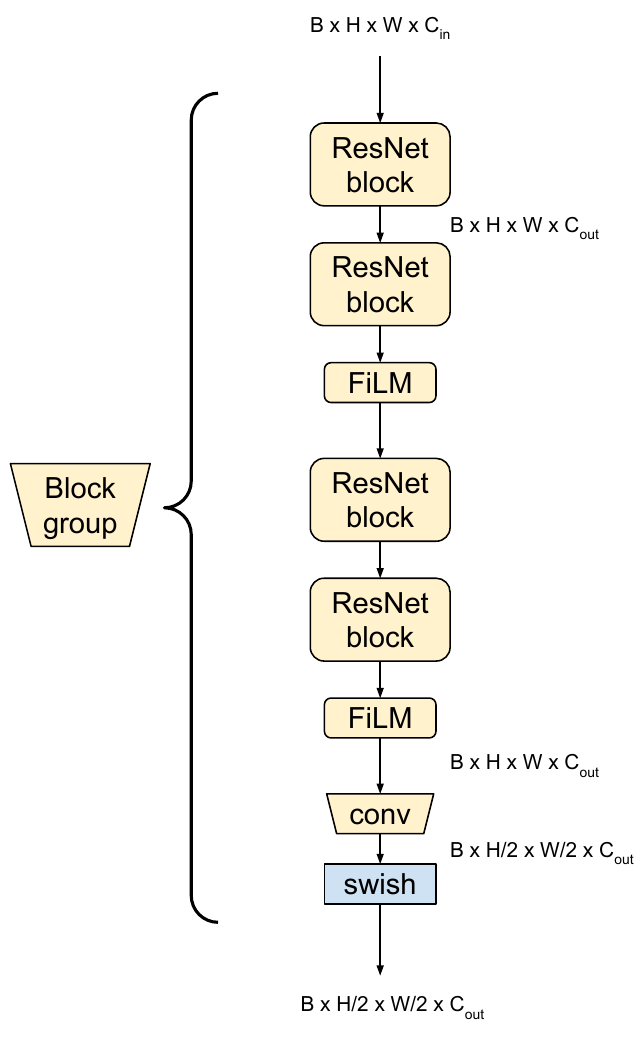}
        \caption{Block group \label{fig:vq:enc:block_group}}
    \end{subfigure}
\par\bigskip \begin{subfigure}[c]{.3\textwidth}
        \centering
        \includegraphics[height=2in]{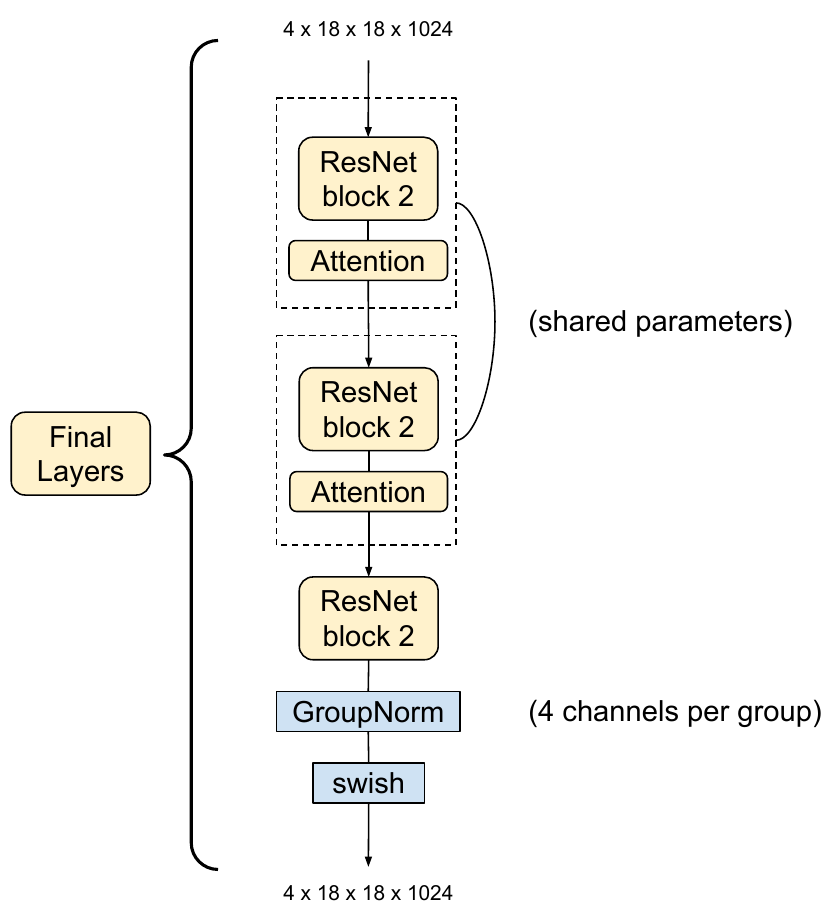}
        \caption{Final layers \label{fig:vq:enc:final_layers}}
    \end{subfigure}
    \hfill
    \begin{subfigure}[c]{0.3\textwidth}
        \centering
        \includegraphics[height=2in]{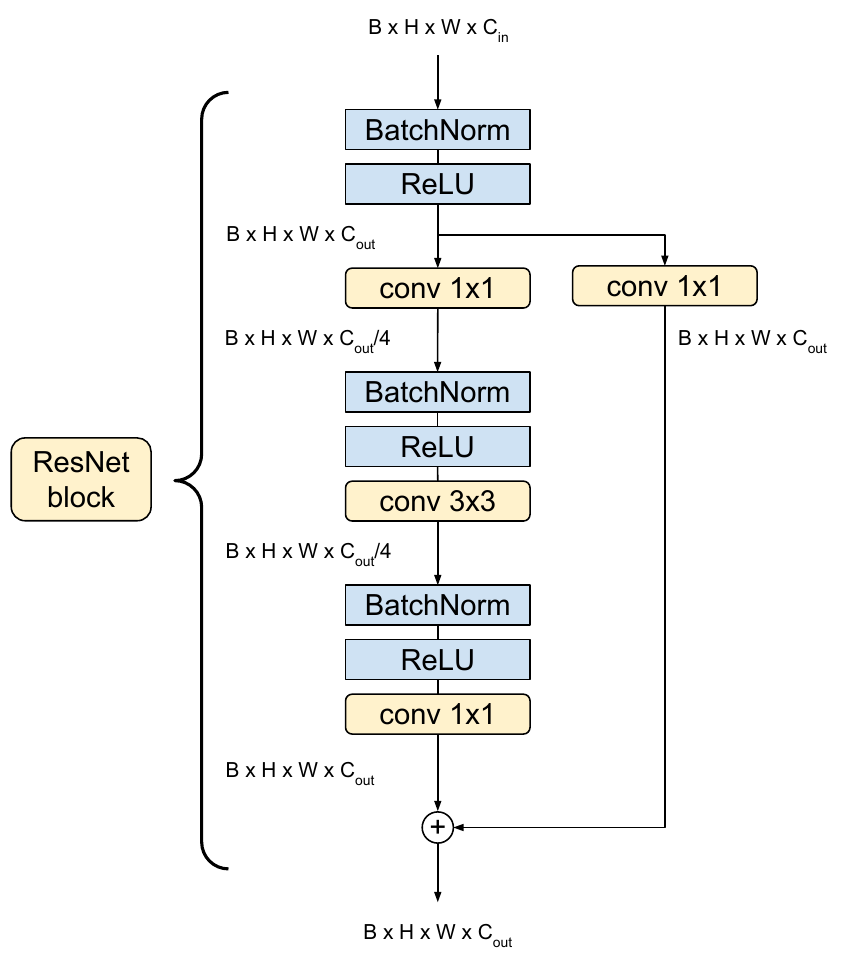}
        \caption{ResNet block (for mid layers)\label{fig:vq:enc:mid:resnet_block}}
    \end{subfigure}
    \hfill \begin{subfigure}[c]{0.3\textwidth}
        \centering
        \includegraphics[height=2in]{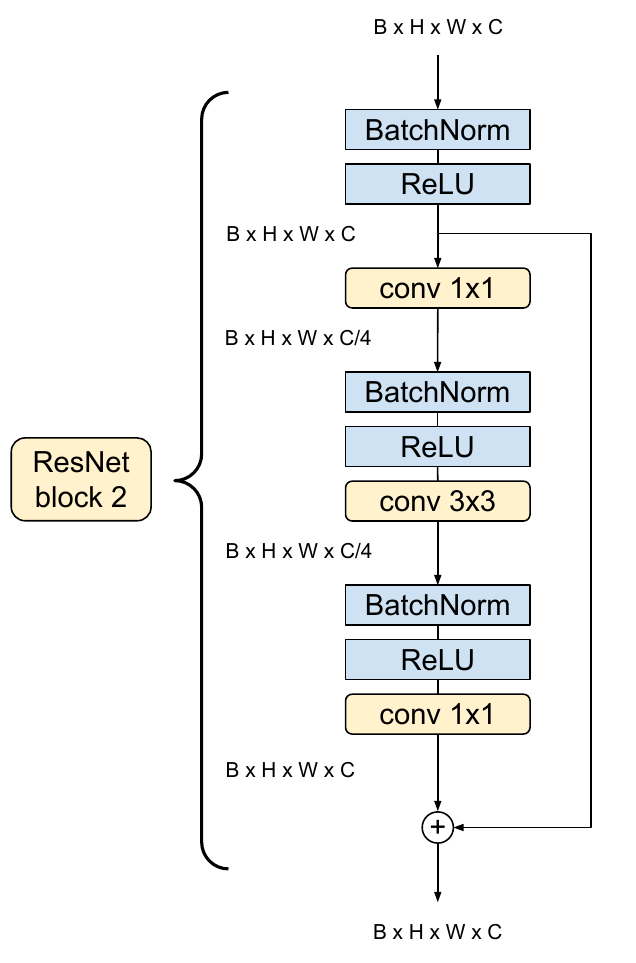}
        \caption{ResNet block (for initial and final layers)\label{fig:vq:enc:final:resnet_block}}
    \end{subfigure}
\caption{The VQ-GAN network, showing the encoder pathway in detail as well as the GAN discriminator. For the decoder, the upsampling convolution in fig.~\ref{fig:vq:enc:block_group} first upsamples by $2\times$ using nearest-neighbor sampling, then applies a stride-1 conventional convolution, not a transposed convolution. For the decoder, the initial (fig~\ref{fig:vq:enc:initial_layers}) and final (fig~\ref{fig:vq:enc:final_layers}) layers loop through the ResNet-attention block (dotted rectangle) three times, not twice as shown for the encoder. The ``Attention'' module denotes an all-to-all attention layer. 
    \label{fig:vq}}
\end{figure}

Figure~\ref{fig:vq} shows the detailed network diagram of the VQ-GAN model, alongside details of the convolutional networks used for encoding and decoding and the patch discriminator of the GAN.

\begin{figure}
    \centering
\begin{subfigure}[t]{\textwidth}
        \centering
        \includegraphics[width=.6\textwidth]{figures/networks/hyperprior_horizontal.pdf}
        \caption{The hyperprior network, which contains a factorized prior network (dashed rectangle). C, the number of input channels, varies by dataset and conditioning. The surface dataset has 4 channels, the vertical level dataset has 5. Conditioning adds 6 channels. N, the codebook size, varies from 512 to 8192 depending on the experiment.}
    \end{subfigure}
\par\bigskip \hfill
    \begin{subfigure}[b]{.3\textwidth}
        \centering
        \includegraphics[height=3in]{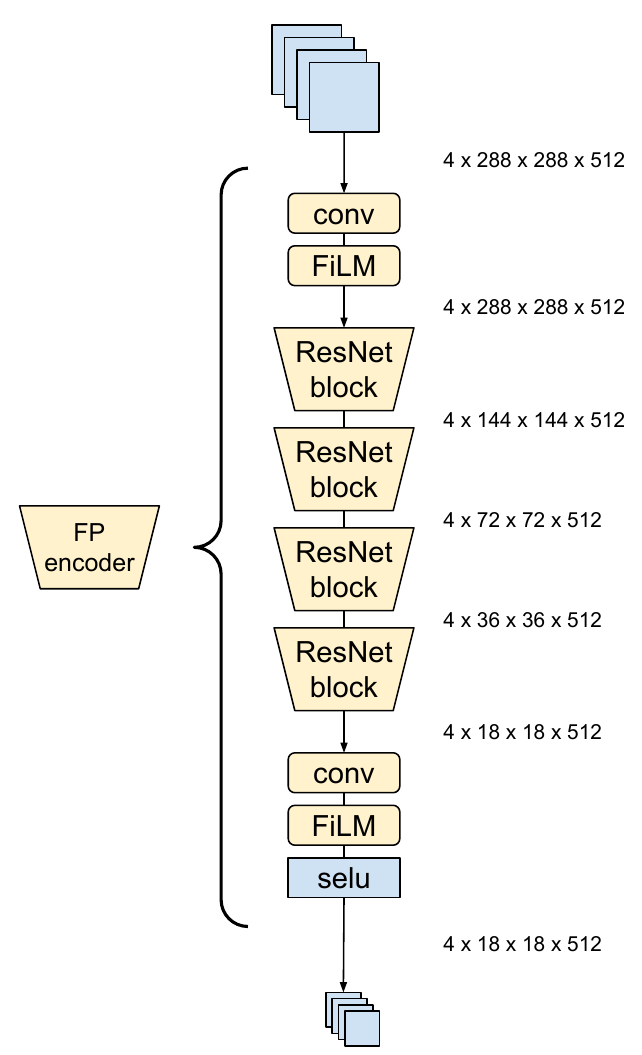}
        \caption{FP encoder\label{fig:balle:fp_encoder}}
    \end{subfigure}
    \hfill
    \begin{subfigure}[b]{0.3\textwidth}
        \centering
        \includegraphics[height=3in]{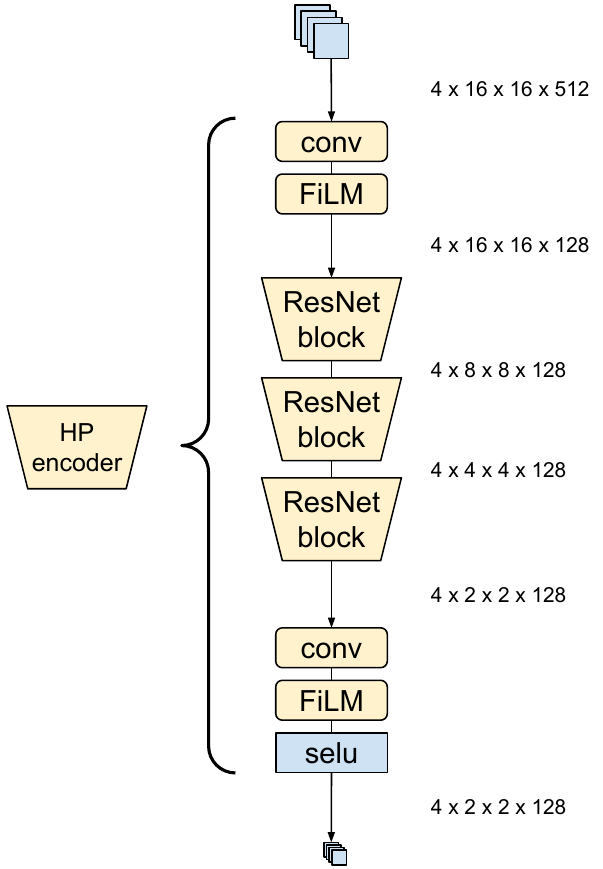}
        \caption{HP encoder\label{fig:balle:hp_encoder}}
    \end{subfigure}
    \hfill
    \begin{subfigure}[b]{0.3\textwidth}
        \centering
        \includegraphics[height=2in]{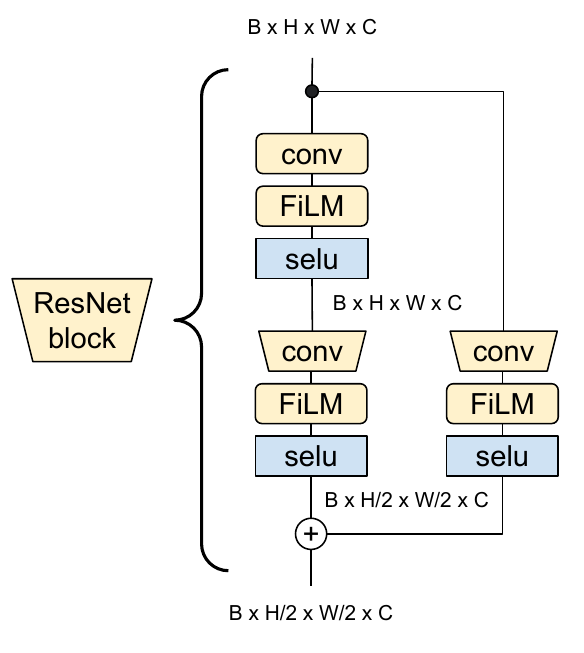}
        \vspace{.4in}
        \caption{Encoder ResNet block\label{fig:balle:resnet}}
    \end{subfigure}
    \hfill
    \caption{Hyperprior network architecture. Yellow: parameterized (learned) operations. Trapezoids: resizing operations. Fig~\ref{fig:balle:fp_encoder}~to~\ref{fig:balle:resnet} show the encoder sub-networks. To get their decoder analogues, replace conv with deconv operations, and reverse the arrows, and in fig~\ref{fig:balle:resnet}, swap the split~($\bullet$) and  sum~($\oplus$) nodes. Convolutions and deconvolutions are 2D, using kernels with spatial dimensions of 3x3. Downsampling convolutions use a stride length of 2. Upsampling deconvolutions upscale the input by a factor of 2 using nearest-neighbor sampling, then deconvolve with a stride of 1. The FiLM operation is conditioned on elevation level, but this conditioning input is omitted for visual clarity.\label{fig:balle}}
\end{figure}

Figure~\ref{fig:balle} shows the network diagram of the hyperprior model, alongside details of the convolutional networks used for encoding and decoding.
 
\section{\cite{huang2022compressing} reproduction and comparison}
\label{sx:appendix:huangiclr2023}
\begin{figure}[h]
\centering
\includegraphics[width=0.9\textwidth]{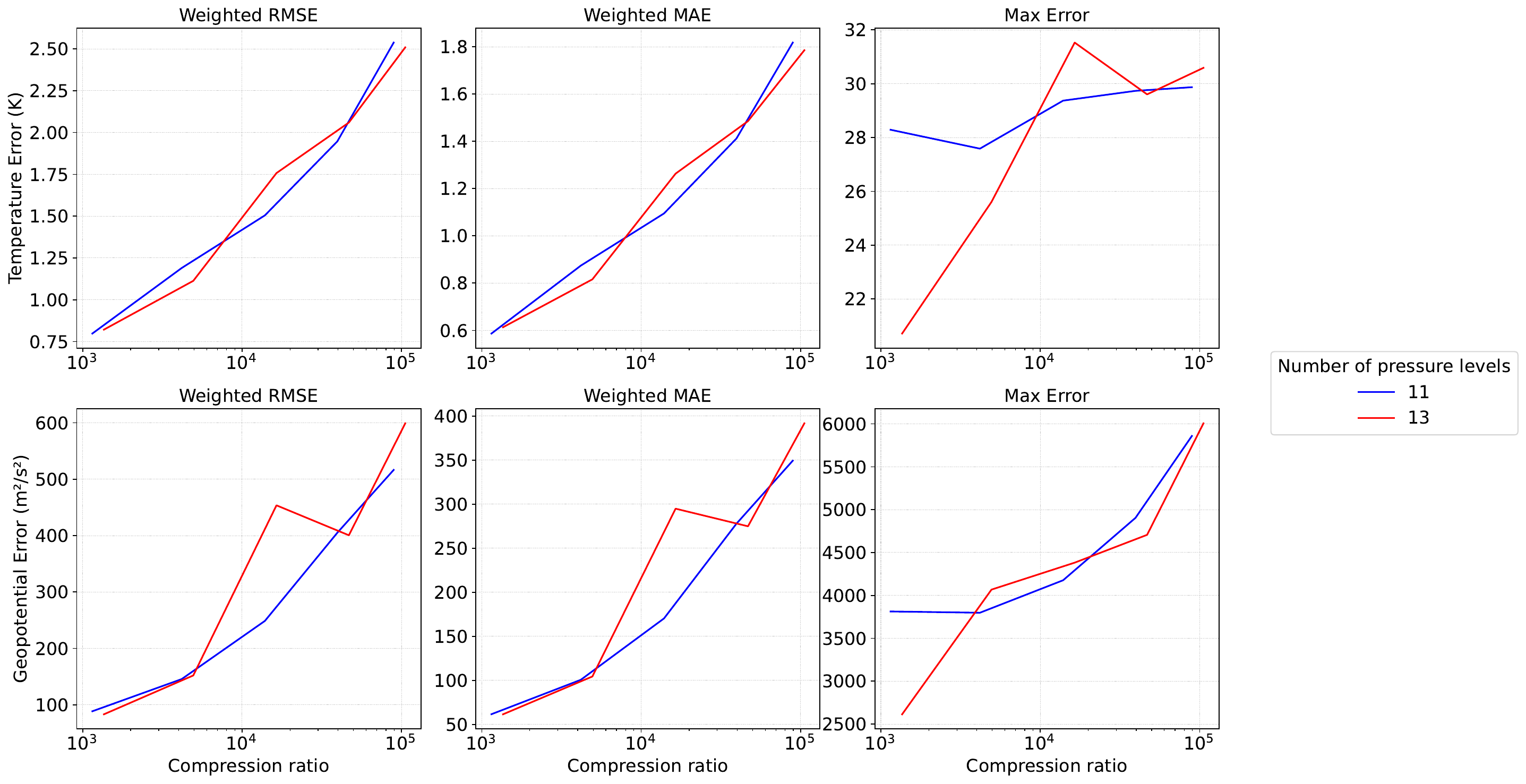}
\caption{Comparison of reconstruction performance for models trained on 11 and 13 pressure levels, with varying network widths on data from year 2016 with 24 hour temporal resolution and 0.25 degrees spatial resolution.}
\label{fig:huangiclr2023_global_11_13_levels_comparison}
\end{figure}

While ~\cite{huang2022compressing} train on 11 pressure levels (10, 50, 100, 200, 300, 400, 500, 700, 850, 925, 1000 hPa), we train on 13 pressure levels (50, 100, 150, 200, 250, 300, 400, 500, 600, 700, 850, 925, 1000 hPa).
To assess the impact of the change in number and level of the pressure levels on reconstruction accuracy, we compare models trained on the 11 pressure  levels used by \cite{huang2022compressing} with models trained on our chosen 13 pressure levels for one of the four datasets, namely "Dataset 2", which contains daily data at the spatial resolution of 0.25 degrees for the year 2016.
We chose to compare on this dataset specifically as it has the highest spatial resolution.
We adapted the models proposed by 
~\cite{huang2022compressing} to accommodate our 13 pressure levels, varying the width parameter, which corresponds to different compression ratios.
As the number of model parameters does not depend on the number of levels, encoding a larger number of levels for a given architecture results in a higher compression ratio. 
Figure \ref{fig:huangiclr2023_global_11_13_levels_comparison} demonstrates that their model can effectively handle the additional pressure levels while maintaining comparable reconstruction performance across different width values. A slight degradation in performance is observed for the width=128 model when trained on 13 levels. It is important to note that the results displayed are from our own implementation of their model, not the original results presented in their paper, although they exhibit close agreement.

\begin{figure}[h]
\centering
\includegraphics[width=0.9\textwidth]{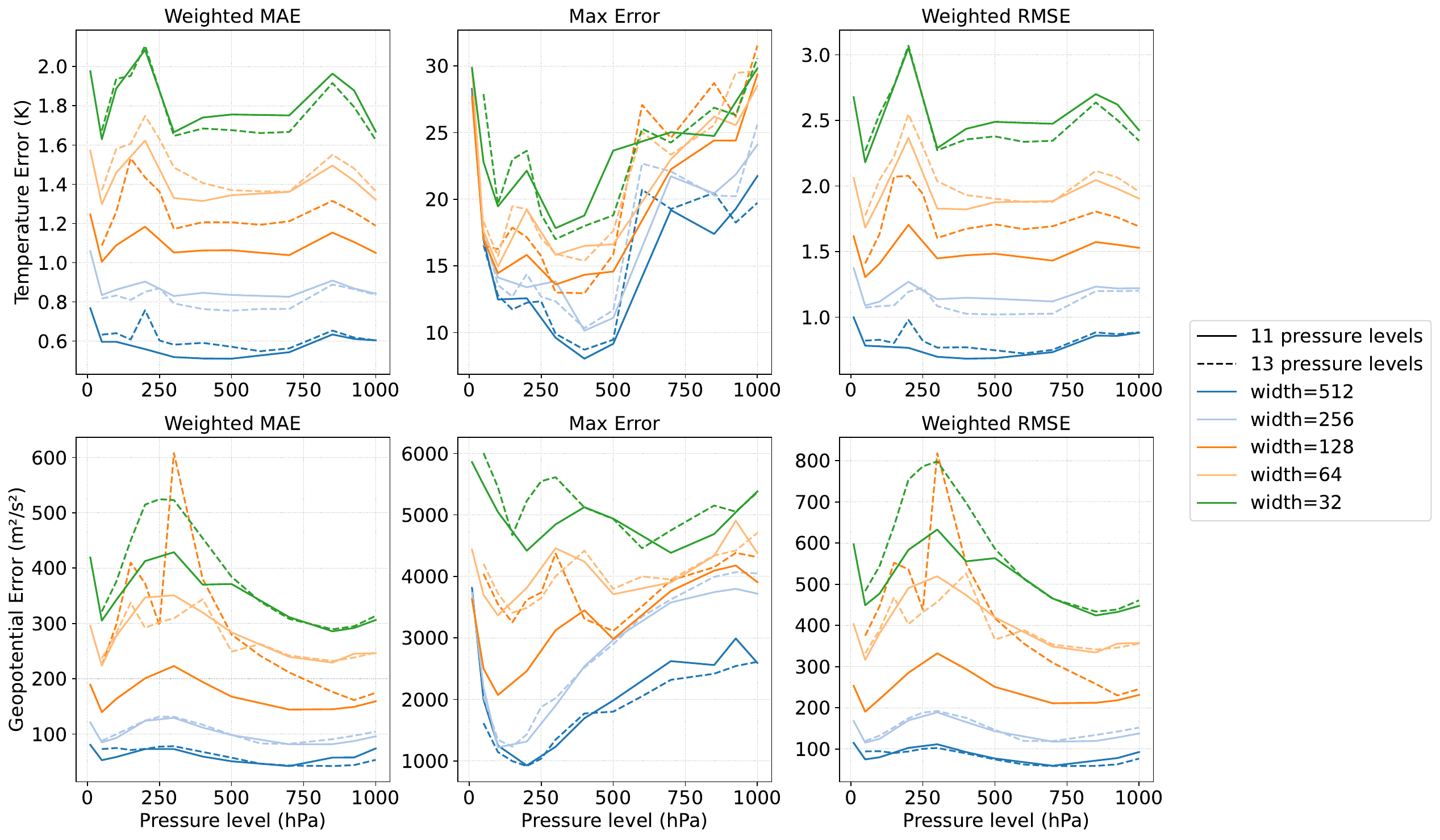}
\caption{Comparison of per-pressure level reconstruction accuracy for models trained on 11 and 13 pressure levels, with varying network widths.}
\label{fig:huangiclr2023_per_level_11_13_levels_comparison}
\end{figure}

We further investigated the reconstruction accuracy of their models at each pressure level when trained on their 11 levels and our selected 13 levels. As shown in Figure \ref{fig:huangiclr2023_per_level_11_13_levels_comparison}, similar performance was achieved for a given pressure level across different width values, with a small degradation when training on 13 levels for a width of 128. Here, width corresponds to the number of neurons in the fully connected layers of the network.

Given these observations, for comparison with ~\cite{huang2022compressing} we compare reconstructions for geopotential and temperature for 13 vertical levels  (50, 100, 150, 200, 250, 300, 400, 500, 600, 700, 850, 925, 1000 hPa) for data from the year 2016 sampled every 24 hours reprojected on (721, 1440) latitude/longitude grids. The corresponding data shape is therefore (366, 13, 721, 1440) for each variable. 
\section{Detailed analysis}
\label{sx:appendix:analysis}
\subsection{Supplemental results for VQ-VAE}

\begin{figure}
    \centering
    \includegraphics[width=\textwidth, trim = 6cm 2.5cm 5cm 4.5cm, clip]{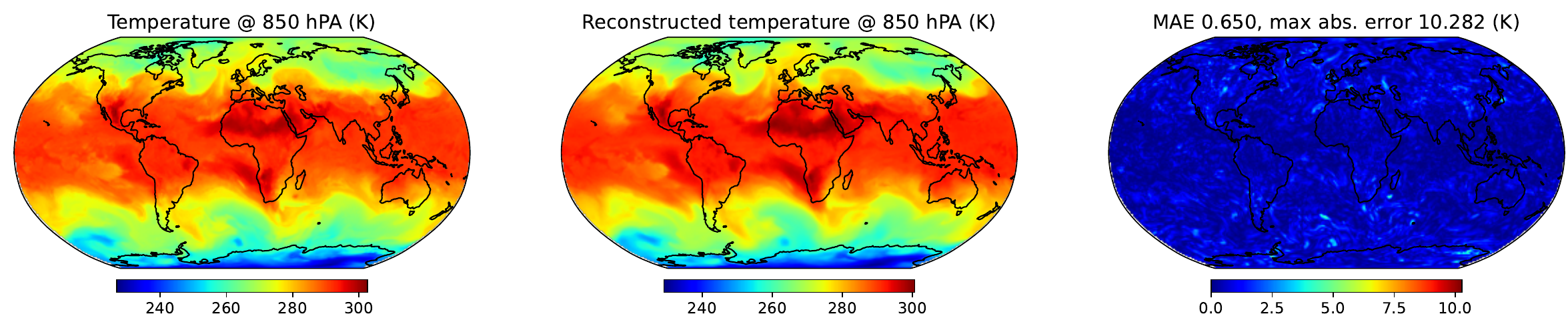}
    \includegraphics[width=\textwidth, trim = 6cm 2.5cm 5cm 4.5cm, clip]{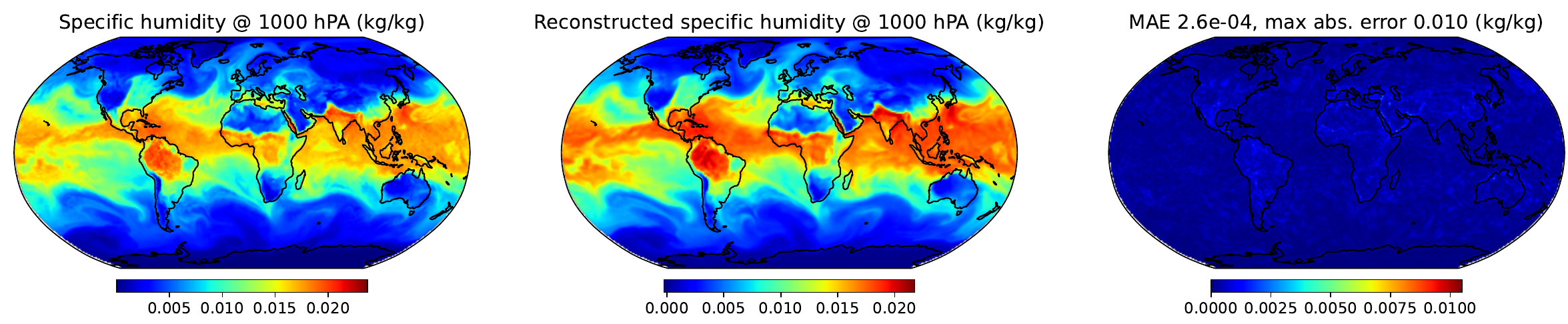}
    \includegraphics[width=\textwidth, trim = 6cm 2.5cm 5cm 4.5cm, clip]{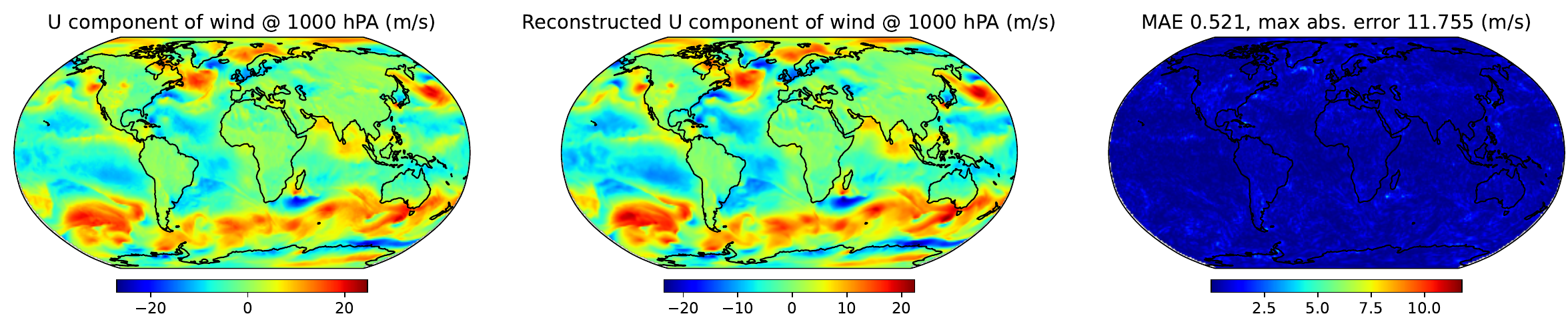}
    \includegraphics[width=\textwidth, trim = 6cm 4cm 5cm 3cm, clip]{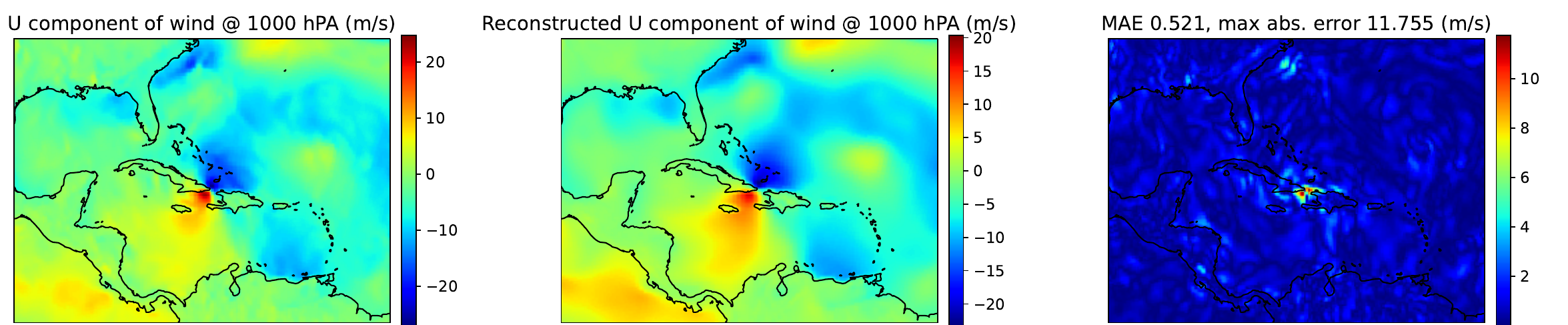}
    \includegraphics[width=\textwidth, trim = 6cm 2.5cm 5cm 4.5cm, clip]{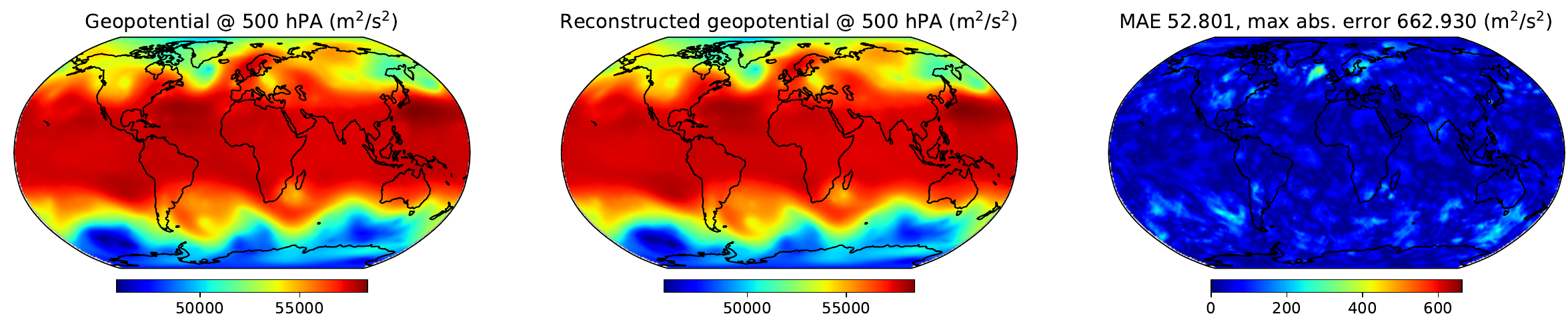}
    \includegraphics[width=\textwidth, trim = 6cm 4cm 5cm 3cm, clip]{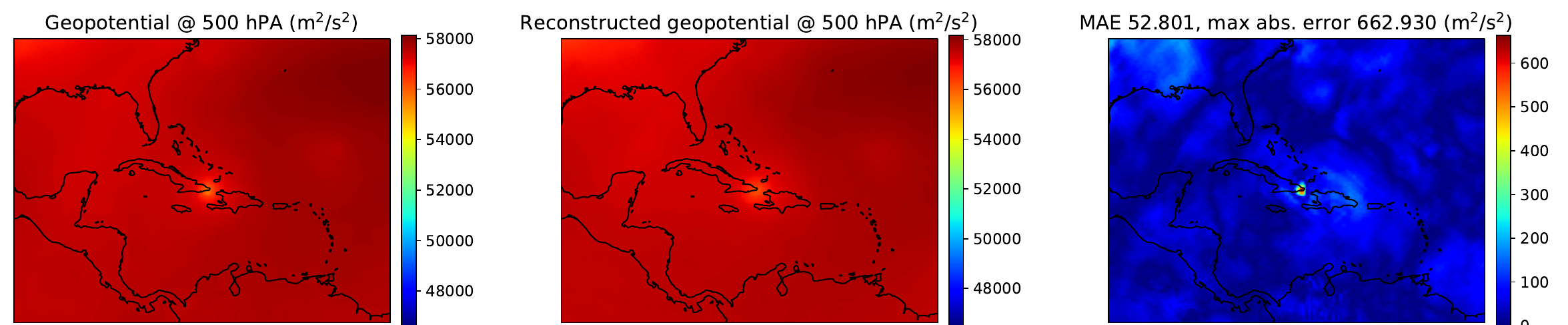}
    \hfill
    \caption{Examples of 3-layer VQ-VAE reconstructions of a global frame on 2016/10/5 at 0 UTC, compressed $\approx 1000\times$. From top to bottom, plots show temperature at 850 hPa, specific humidity at 1000 hPa, the zonal component of wind at 1000 hPa and geopotential at 500 hPa. While errors are higher than the hyperprior model's (see Fig. \ref{fig:hurricane_matthew_hyperprior}), Hurricane Matthew's reconstruction remains visible.}
    \label{fig:hurricane_matthew_vqvae}
\end{figure}

Figure \ref{fig:hurricane_matthew_vqvae} shows the example of reconstruction of Hurricane Matthew on 5 October 2016 using the 3-layer VQ-VAE with $1100\times$ compression rate. It can be contrasted with Fig. \ref{fig:hurricane_matthew_hyperprior} for the hyperprior model. Specifically on the frame corresponding to 5 October 2016 at 0 UTC, the VQ-VAE achieves $\text{MAE}=52.8~\text{m}^2/\text{s}^2$ on geopotential at 500 hPa, $\text{MAE}=0.65^\circ$~K on temperature at 850 hPa, $\text{MAE}=2.1 \times 10^-4$ for specific humidity at 1000 hPa, $\text{MAE}=0.52~\text{m}/\text{s}$ for zonal wind speed at 1000 hPa; whereas the Hyper Prio achieves respectively $\text{MAE}=32.8~\text{m}^2/\text{s}^2$, $\text{MAE}=0.5^\circ$~K, $\text{MAE}=2.1\times10^-4$ and $\text{MAE}=0.39~\text{m}/\text{s}$. In line with the overall performance of these models on the whole dataset (see Fig. \ref{fig:error_vs_cr}) and on the hourly 2016 baseline dataset (see Fig. \ref{fig:reprojection_error_vs_cr}), the hyperprior model performs slightly better than the VQ-VAE in reconstructing this specific hurricane.

\subsection{Reconstruction results in HEALPix projection}

\begin{figure}
    \centering
    \includegraphics[width=0.8\textwidth, trim = 3cm 0cm 3cm 0cm, clip]{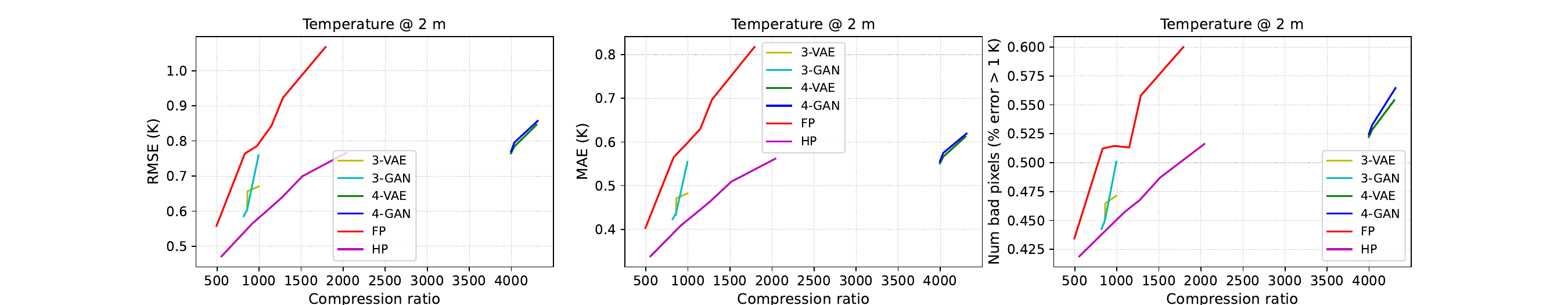}
    \includegraphics[width=0.8\textwidth, trim = 3cm 0cm 3cm 0cm, clip]{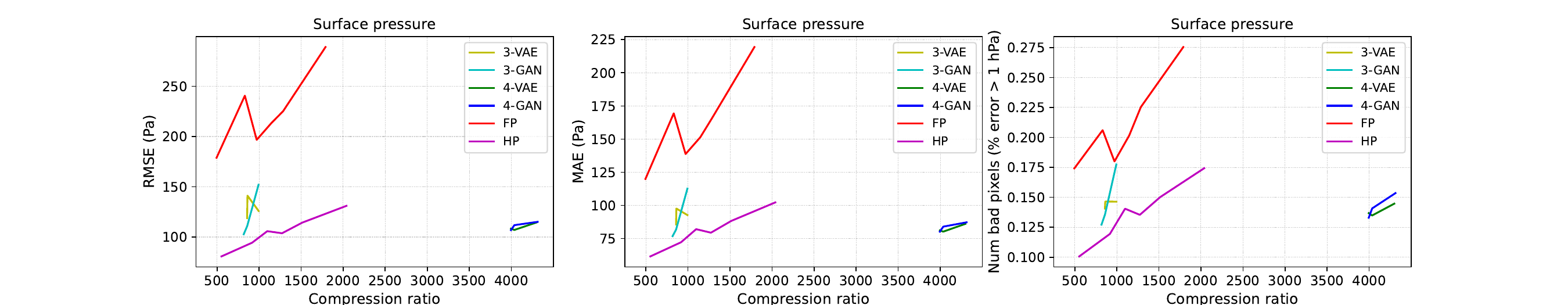}
    \includegraphics[width=0.8\textwidth, trim = 3cm 0cm 3cm 0cm, clip]{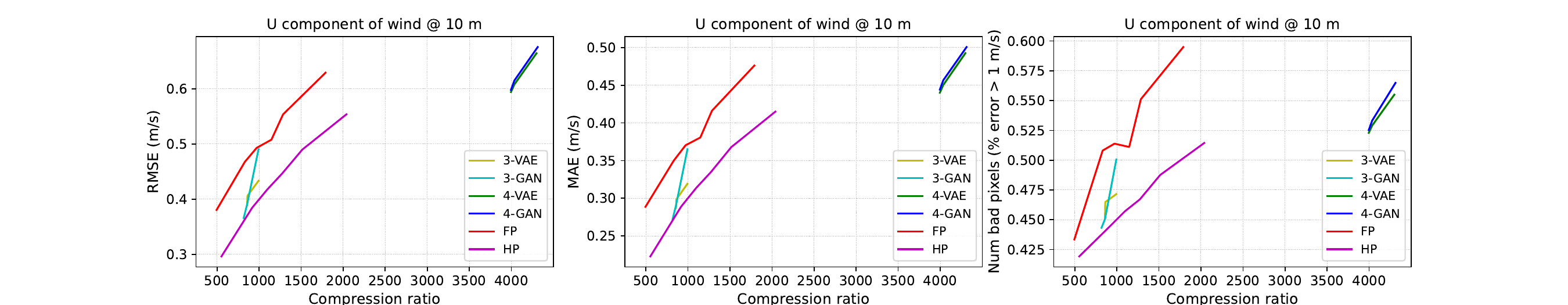}
    \includegraphics[width=0.8\textwidth, trim = 3cm 0cm 3cm 0cm, clip]{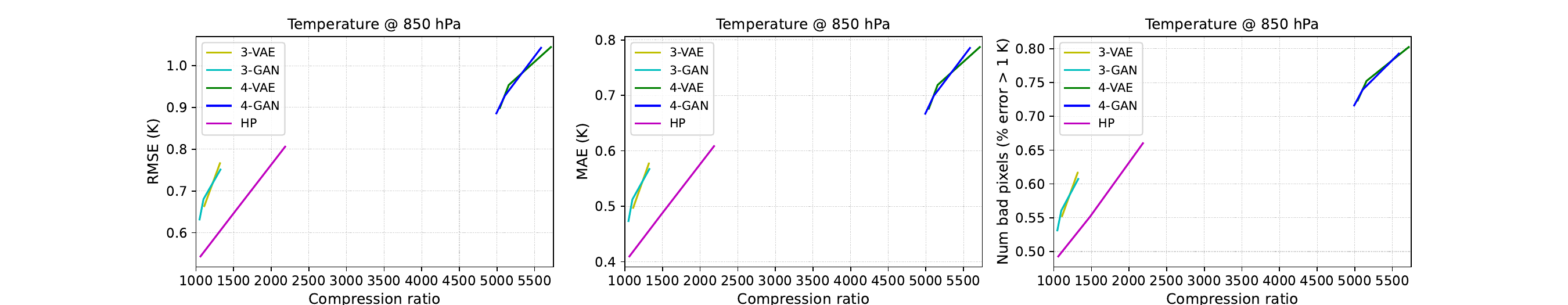}
    \includegraphics[width=0.8\textwidth, trim = 3cm 0cm 3cm 0cm, clip]{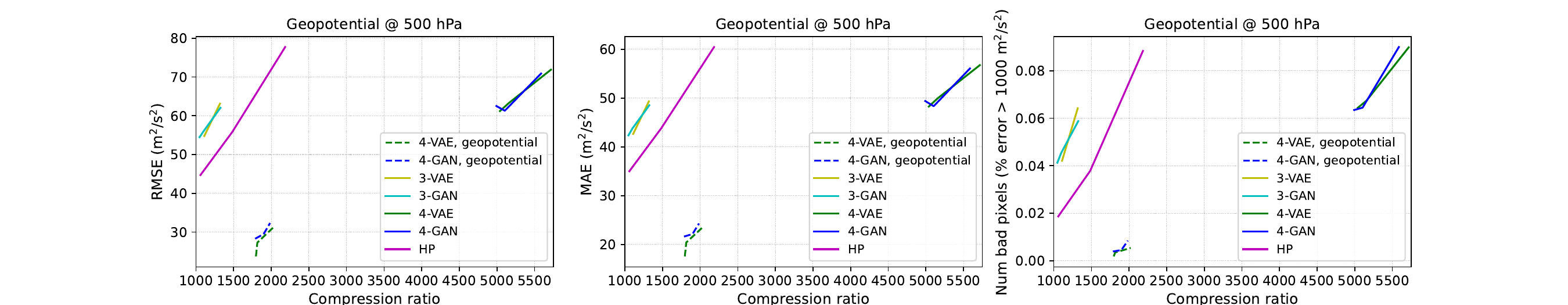}
    \includegraphics[width=0.8\textwidth, trim = 3cm 0cm 3cm 0cm, clip]{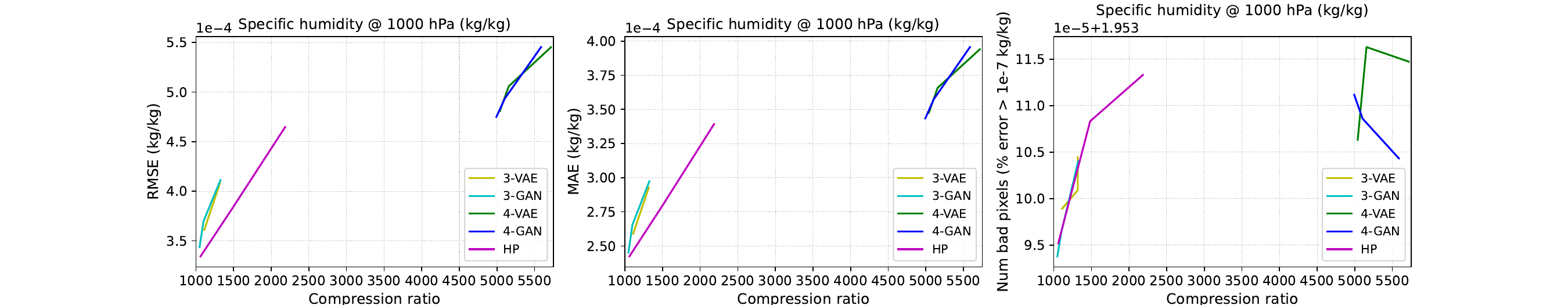}
    \includegraphics[width=0.8\textwidth, trim = 3cm 0cm 3cm 0cm, clip]{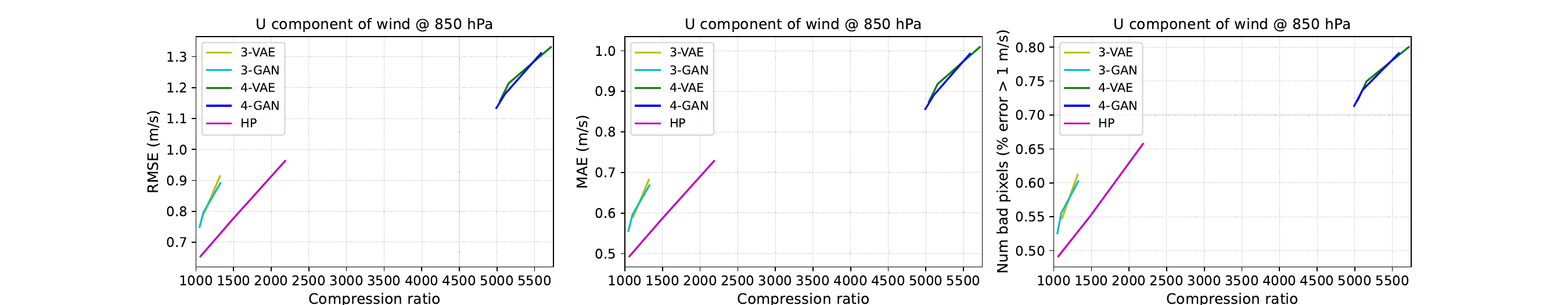}
    \hfill
    \caption{Reconstruction error vs. compression ratio on both surface and vertical data, in HEALPix projection. Colors correspond to models: respectively 3-block VQ-VAE and VQ-GAN, 4-block VQ-VAE and VQ-GAN, factorized prior and hyperprior. Rows: atmospheric variables: temperature at 2 m, surface pressure, zonal component of wind at 10 m, temperature at 850 hPa, geopotential at 500 hPa, specific humidity at 1000 hPa and zonal component of wind at 850 hPa. Columns: error metrics. Errors are computed over all hourly frames in the evaluation set (2010-2023). MAE is the mean absolute error. ``Num bad pixels'' is the percent of pixels exceeding the threshold shown on the y-axis.}
    \label{fig:error_vs_cr}
\end{figure}

Figure \ref{fig:error_vs_cr} compares all models on the reconstruction task in HEALPix projection on both surface and vertical data, using the following metrics: RMSE, MAE and number of pixels beyond a given error threshold. These results are reported in HEALPix pixel space, i.e. without reprojection to latitude/longitude coordinates and subsequent additional reprojection errors, and confirm the choice of the hyperprior model as the least erroneous compression method.

\subsection{Ablations}

We verify that the MAE of a VQ-VAE model trained on $288 \times 288$ squares (which correspond to a HEALPix tile padded with 16 pixels on each side, see Appendix \ref{sx:appendix:overset_healpix}), and when evaluated in the central $256 \times 256$ area, is similar to the MAE of a VQ-VAE model directly trained on on $256 \times 256$ HEALPix squares. The purpose of this comparison is to understand if the models are penalised when we input and reconstruct overset HEALPix grids. We evaluate on the central $256 \times 256$ crop of the HEALPix tile, because these are the non-overlapping data that cover the sphere and are used in reprojection.

\begin{figure}
    \centering
        \includegraphics[width=0.497\textwidth, trim = 2cm 0cm 3cm 0cm, clip]{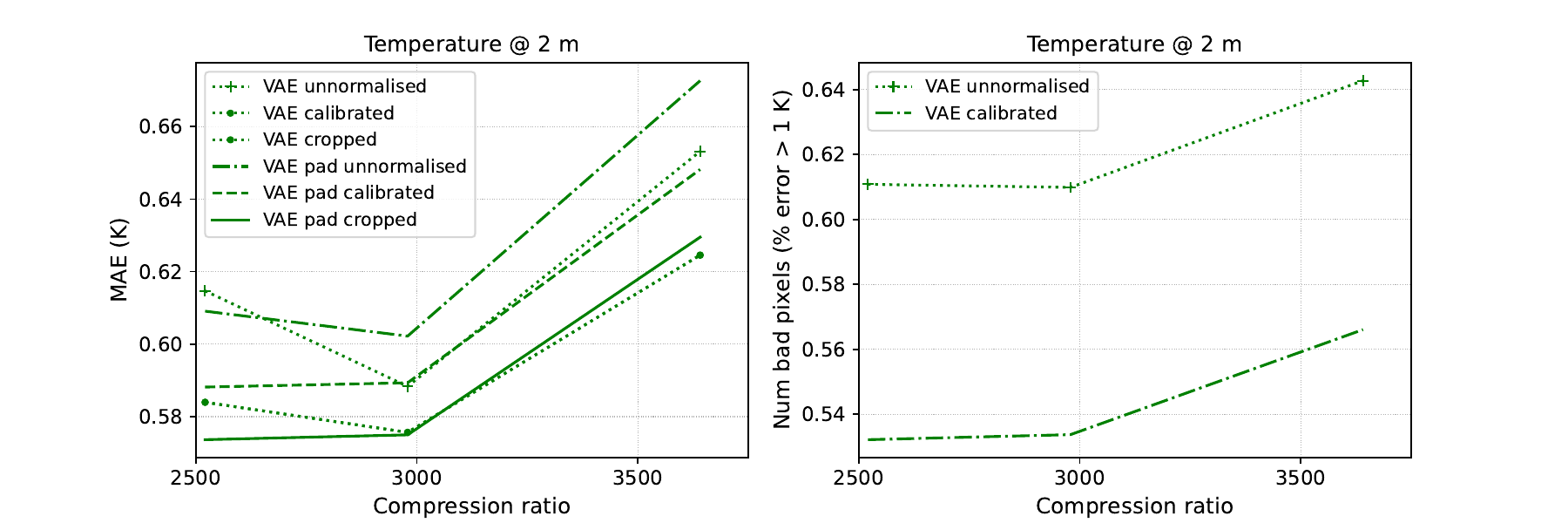}
        \includegraphics[width=0.497\textwidth, trim = 2cm 0cm 3cm 0cm, clip]{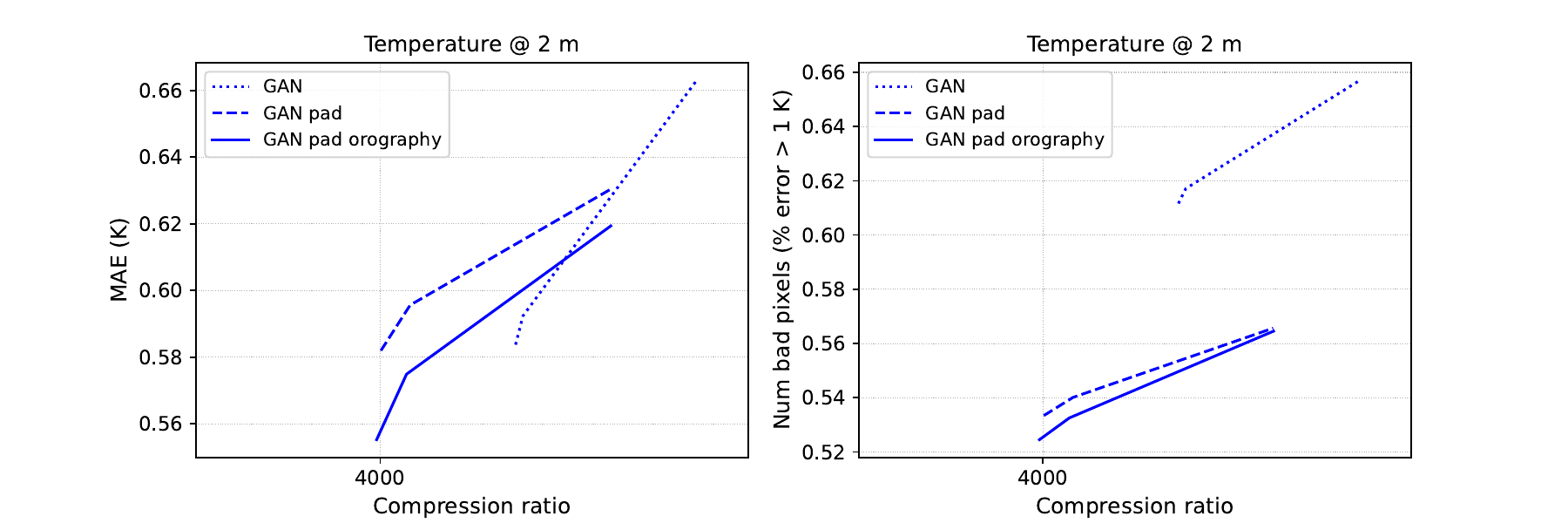}
        \includegraphics[width=0.497\textwidth, trim = 2cm 0cm 3cm 0cm, clip]{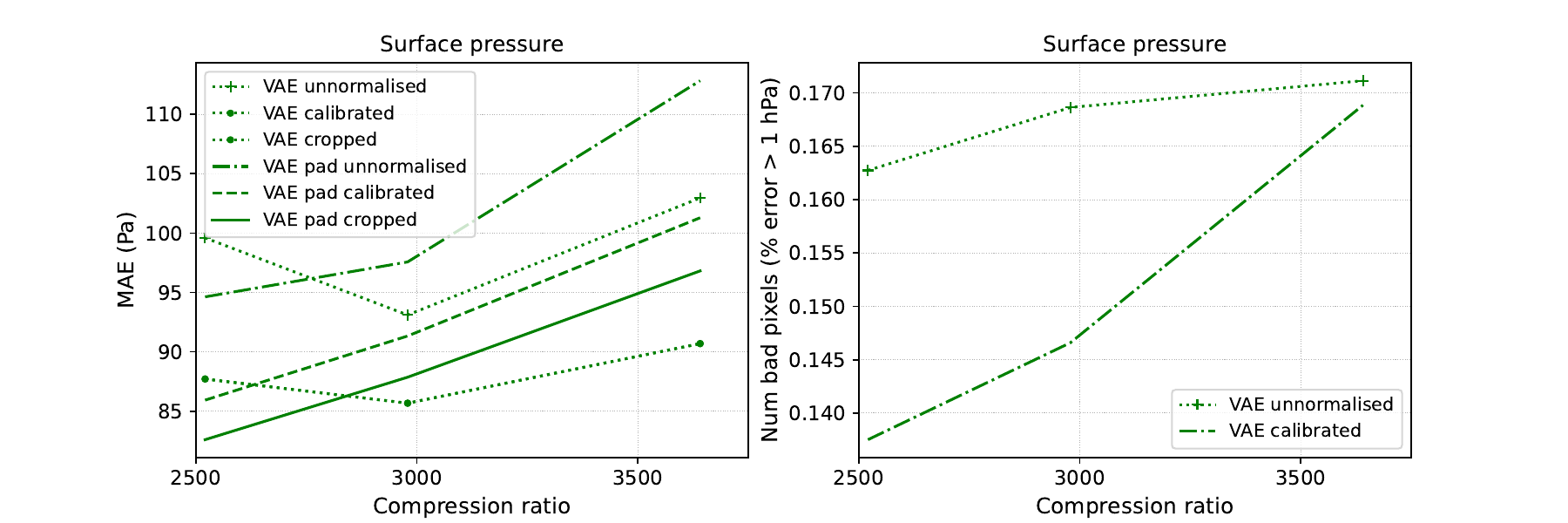}
        \includegraphics[width=0.497\textwidth, trim = 2cm 0cm 3cm 0cm, clip]{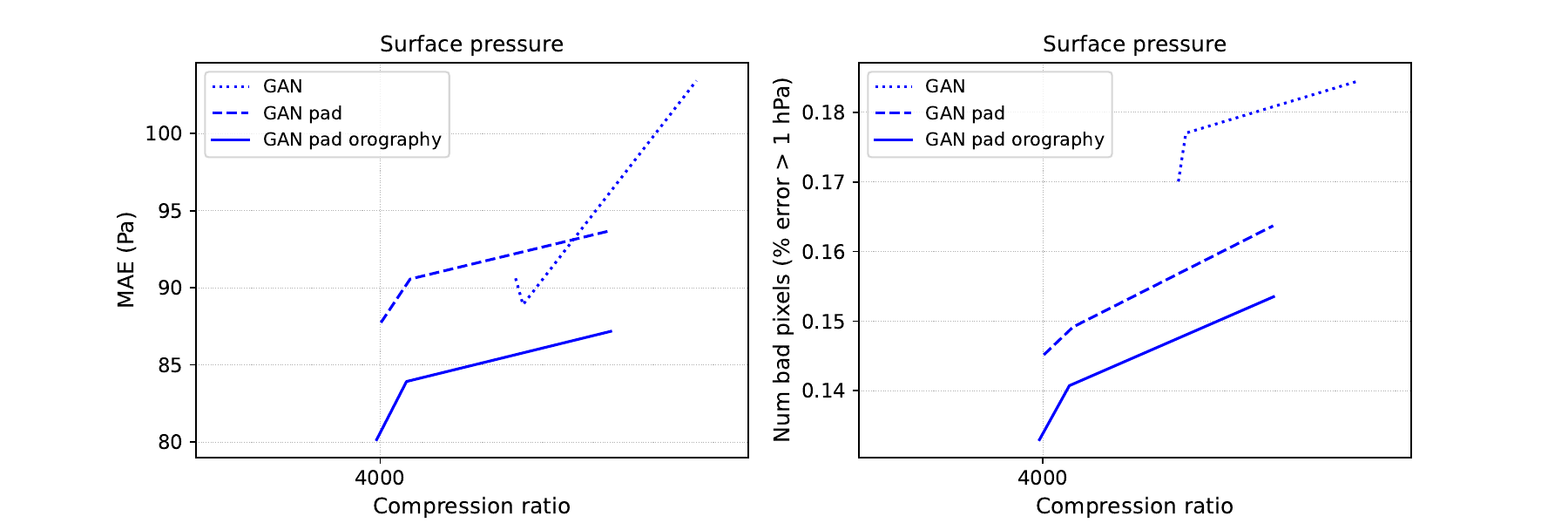}
    \caption{Reconstruction error vs compression ratio on surface data in HEALPix projection. Rows: surface variables: respectively temperature at 2 m and surface pressure. Columns: error metrics: mean absolute error (MAE) and ``number of bad pixels'' which is the percentage of pixels exceeding the 1$^\circ$ K or 1 hPa error threshold. The curves compare two models (4-block VQ-VAEs and VQ-GANs) and three post-processing schemes, to ablate their effects on error. The two models are equivalent, except that one takes padded inputs as described in section~\ref{sx:appendix:overset_healpix}. For most compression ratios, the padded models outperform the unpadded models. Additionally, we plot three postprocessing methods. In \textbf{unnormalized}, the reconstruction is scaled by the standard deviation of that variable, and shifted by the mean, where these statistics are computed over the entire training set. In \textbf{calibrated}, the reconstruction is scaled and shifted by the statistics of that variable as computed from the specific frame and HEALpix square being reconstructed. In \textbf{cropped}, the reconstruction metric is averaged only over the central $256 \times 256$ crop (as opposed to the full HEALPix area). Models with \textbf{orography} use additional conditioning on orography features.}
\label{fig:ablate_pad_and_postprocess}
\end{figure}

The first result that Figure \ref{fig:ablate_pad_and_postprocess} shows, is that calibrating the reconstructions by making each reconstructed square have the same mean and standard deviation as the input square, reduces both MAE and the percentage of pixels beyond the error thresholds. This is visible when comparing {\tt VAE unormalised} to {\tt VAE calibrated} and comparing {\tt VAE pad unormalised} to {\tt VAE pad calibrated}. This result is intuitive because we ensure that the reconstruction has the same first and second order moment statistics as the target. 

The second result is that {\tt VAE cropped} (a model trained to predict $256 \times 256$ outputs and ``cropped'' to the central $256 \times 256$ area (i.e. without any change) has comparable performance to {\tt VAE pad cropped} (a model trained to predict $288 \times 288$ outputs but actually cropped to the central $256 \times 256$ area).

The third results is that including orography inputs drives down the reconstruction MAE and the percentage of bad pixels (beyond the error threshold) (compare {\tt VAE pad} with {\tt VAE pad orography}, and {\tt GAN pad} with {\tt GAN pad orography}).

These ablations confirm the intuitive design choices we adopted in our study.

\subsection{Histogram of errors vs. ground truth values}
\label{sx:appendix:hist2d-err-value}

In order to understand the distribution of the errors vs. the ground truth target values for the hyperprior model, we compute two-dimensional histograms over the entire test set. To make the computation tractable, we compute the histogram by binning the target values and the reconstruction errors. In the case of temperature and wind speed, we bin the targets to the closest integer value, namely from $171^\circ~\text{K}$ to $326^\circ~\text{K}$ for temperature target values, and from -140 m/s to 127 m/s for zonal wind; We bin the errors in a similar way. In the case of geopotential, we bin the targets and errors to the closest 100 $\text{m}^2/\text{s}^2$ value.

Errors are defined as the difference between reconstruction and target, meaning that positive errors correspond to overestimation, and negative errors correspond to underestimation.

\begin{figure}
    \centering
    \includegraphics[width=\textwidth, trim = 4.5cm 6cm 4.5cm 6cm, clip]{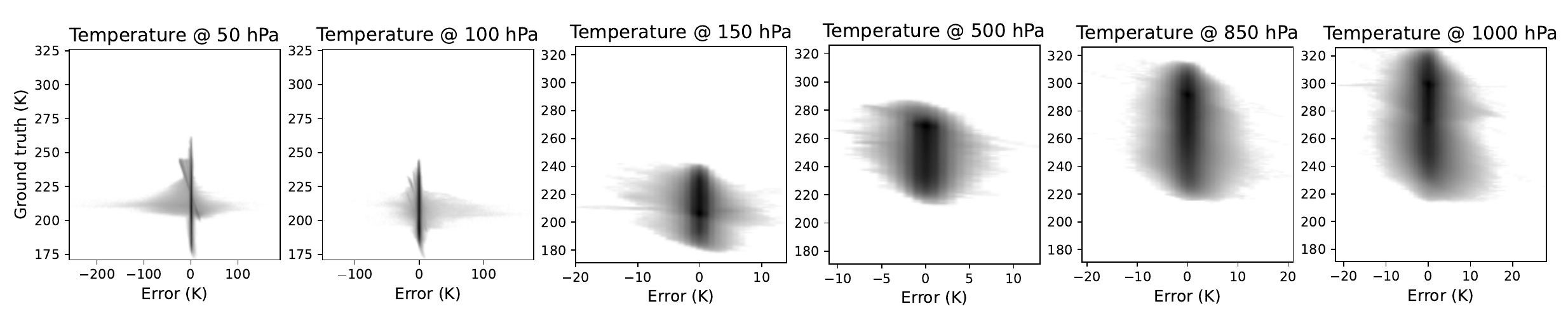}
    \includegraphics[width=\textwidth, trim = 4.5cm 6cm 4.5cm 6cm, clip]{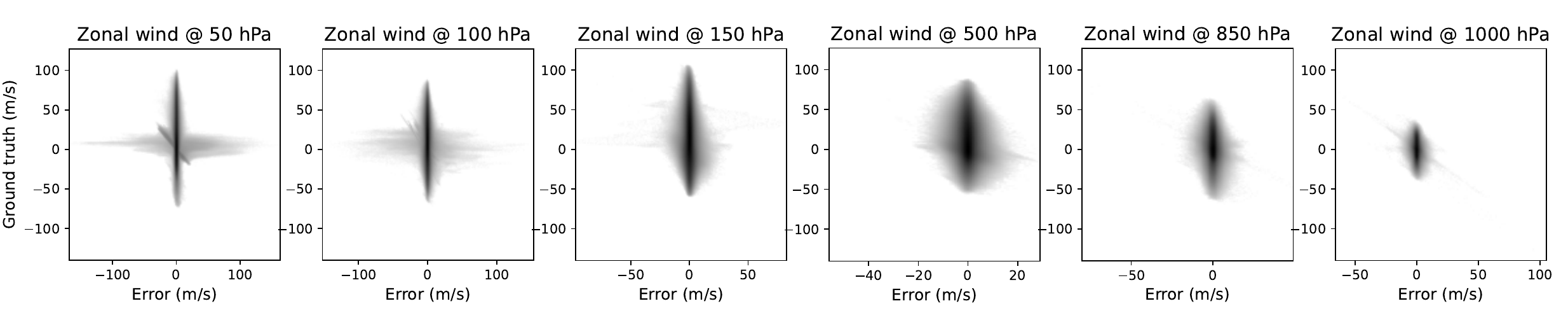}
    \includegraphics[width=\textwidth, trim = 4.5cm 6cm 4.5cm 6cm, clip]{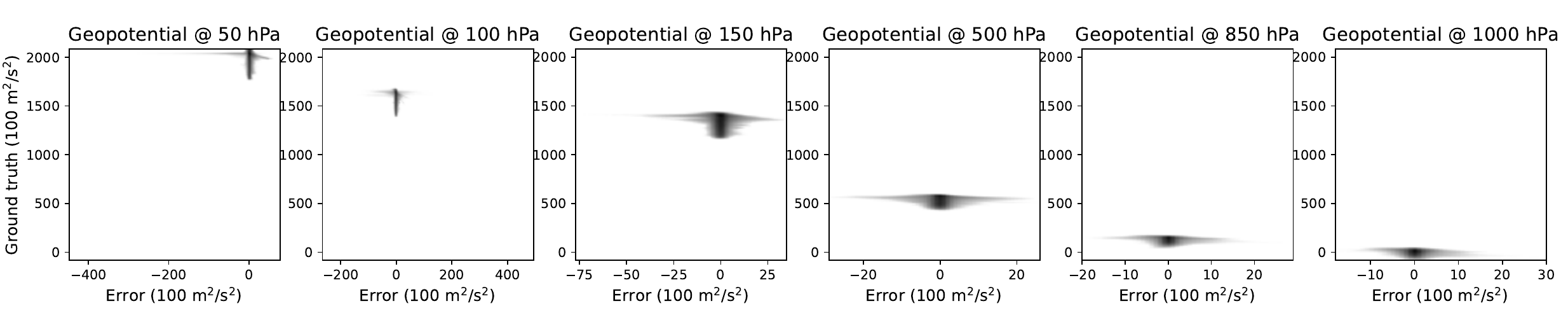}
    \hfill
    \caption{Two-dimensional histograms of reconstruction errors vs. ground truth values for the hyperprior model with $1000\times$ compression rate. Rows correspond to temperature, zonal wind and geopotential, columns correspond to different pressure levels: 50 hPa, 100 hPa, 150 hPa, 500 hPa, 850 hPa and 1000 hPa. For visibility, we plot $\log(\text{hist}+1)$ to highlight low counts.}
    \label{fig:hist2d-err-value}
\end{figure}

Figure \ref{fig:hist2d-err-value} shows the distribution of reconstruction errors made by the hyperprior model at $1000\times$ compression rate, for temperature, zonal wind and geopotential, and for different pressure levels, which is plotted with errors on the abscissa and target values on the ordinate. We observe that when the target value is higher, there are more negative errors, whereas when the target value is lower, there are more positive errors. We also observe that there are diagonal lines with negative slope in the histogram plots. These two observations suggest that the model slightly underestimates higher temperature or wind speed values, and overestimates lower temperature or wind.

\subsection{ERA5 data anomalies at low pressure levels}
\label{sx:appendix:artefacts}

\begin{figure}
    \centering
    \includegraphics[width=\textwidth, trim = 6cm 0cm 5cm 0cm, clip]{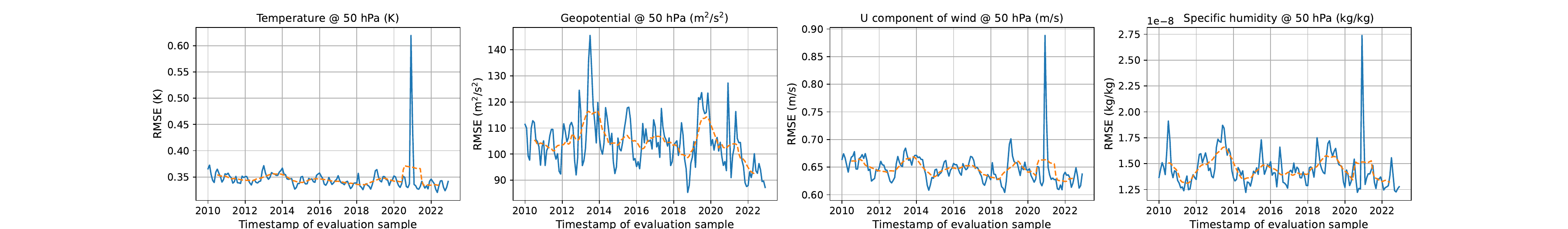}
    \hfill
    \caption{Plot of the average RMSE over 13 years of the test set, from 1/1/2010 to 31/12/2022, using the hyperprior with $1000\times$ compression rate, and on the 50 hPa pressure level. Columns show temperature, geopotential, zonal wind speed and specific humidity. The blue curve shows the monthly average; the red dashed curve shows the yearly rolling average. We notice how there is a spike in RMSE in December 2020 for temperature, wind speed and specific humidity at 50 hPa, confirming the observation about specific humidity anomalies made in Section \ref{sx:discuss:limitation-artefacts}.}
    \label{fig:generalisation-time-50hPa}
\end{figure}

Figure \ref{fig:generalisation-time-50hPa}, shows the RMSE (averaged over HEALPix squares, hours of the days and days of the month, as well as a yearly rolling average) for 4 variables, of the hyperprior model with $1000\times$ compression ratio, evaluated on the 50 hPa pressure level. We notice that there is a spike in RMSE for temperature, zonal wind speed and specific humidity at 50 hPa in December 2020, which seem to confirm the observations about data artefacts at 50 hPa that are discussed in Section \ref{sx:discuss:limitation-artefacts}.

\begin{figure}
    \centering
    \includegraphics[width=\textwidth, trim = 6cm 2cm 5cm 4.5cm, clip]{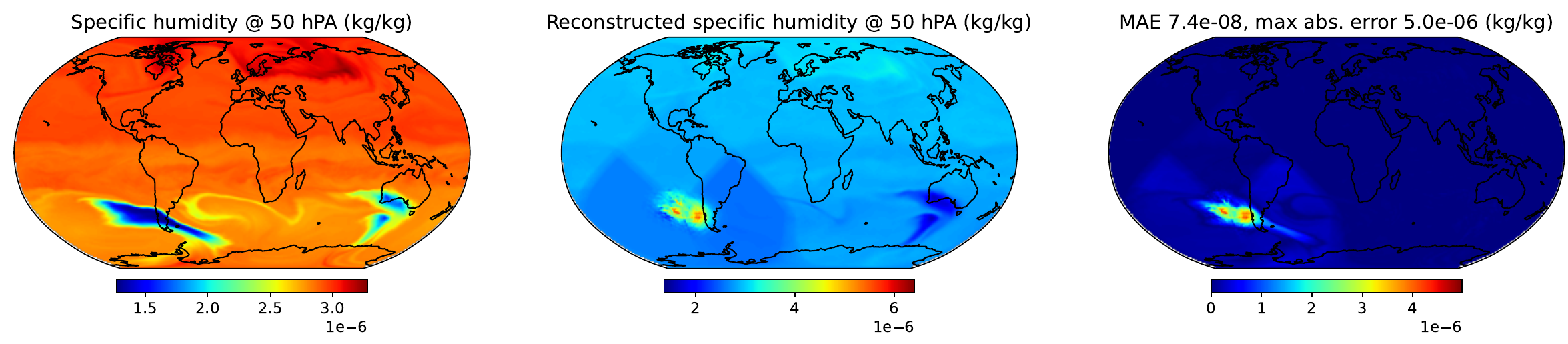}
    \includegraphics[width=\textwidth, trim = 6cm 3.5cm 5cm 3cm, clip]{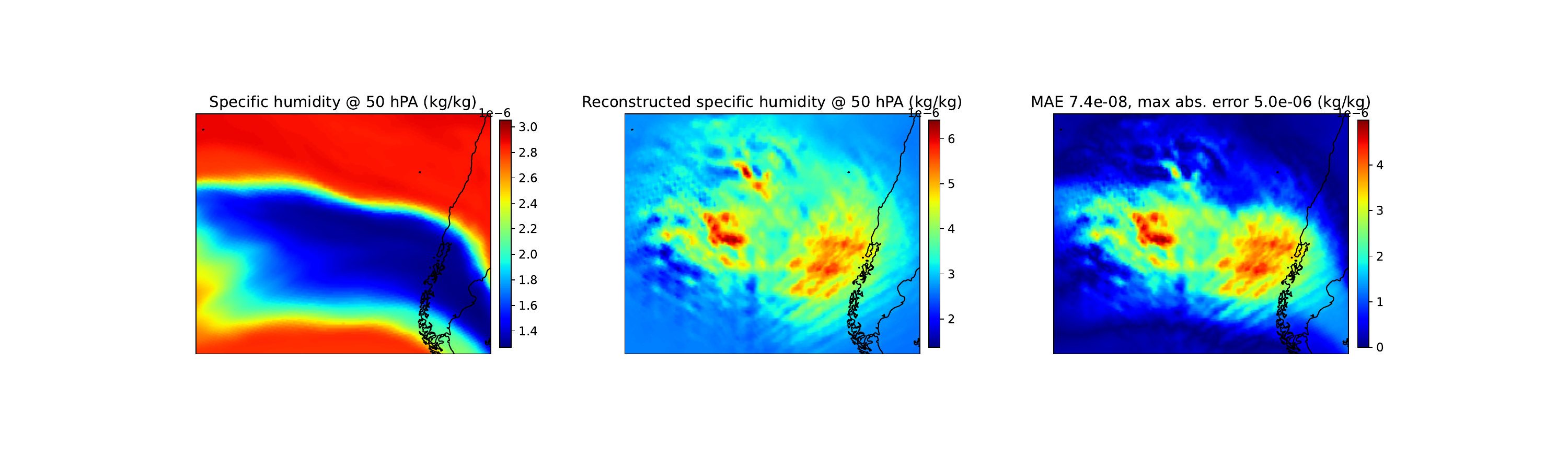}
    \includegraphics[width=\textwidth, trim = 6cm 2.5cm 5cm 4.5cm, clip]{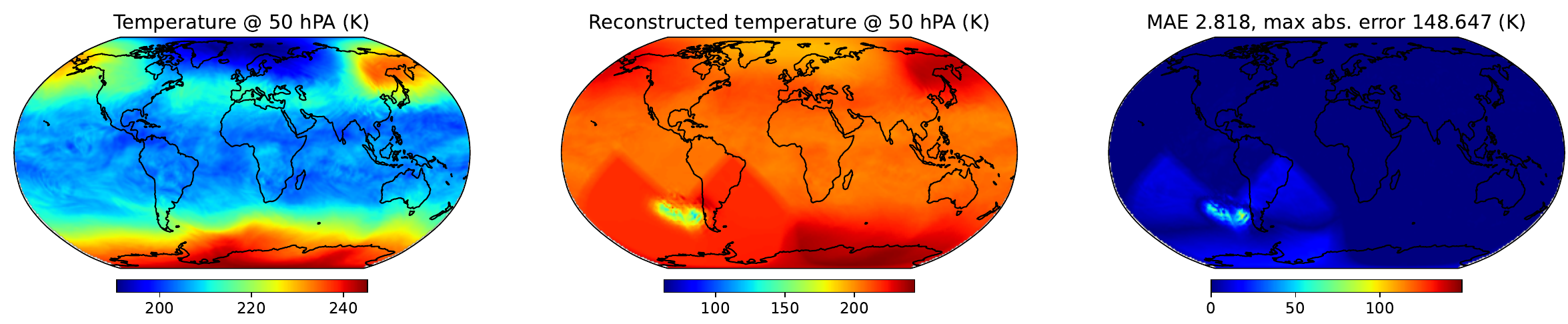}
    \includegraphics[width=\textwidth, trim = 6cm 3.5cm 5cm 3cm, clip]{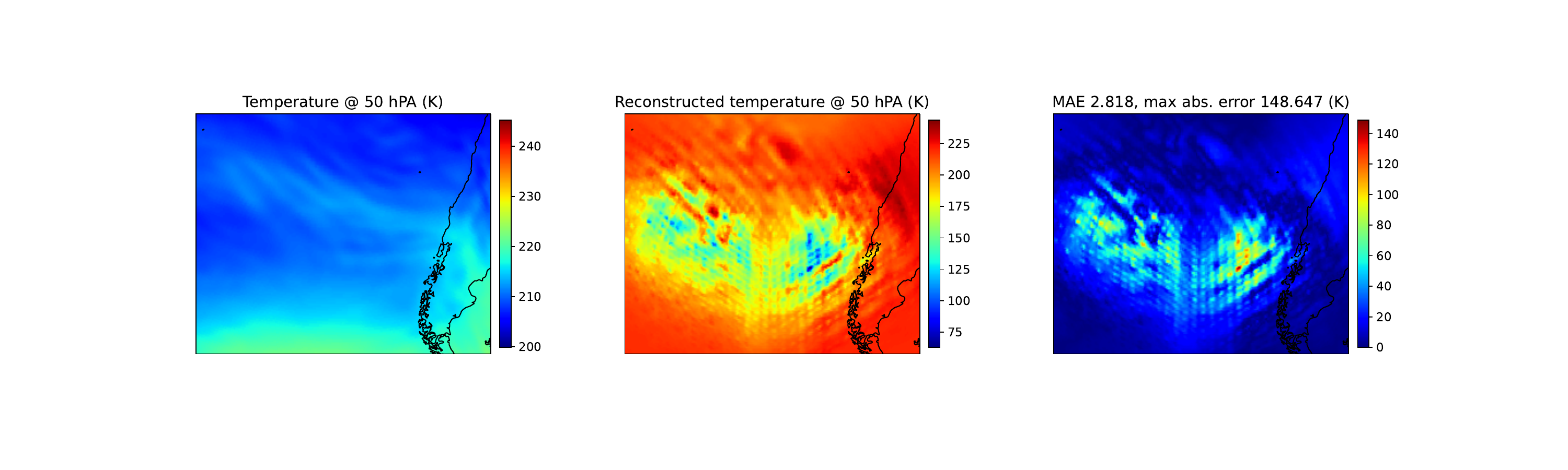}
    \includegraphics[width=\textwidth, trim = 6cm 2cm 5cm 4.5cm, clip]{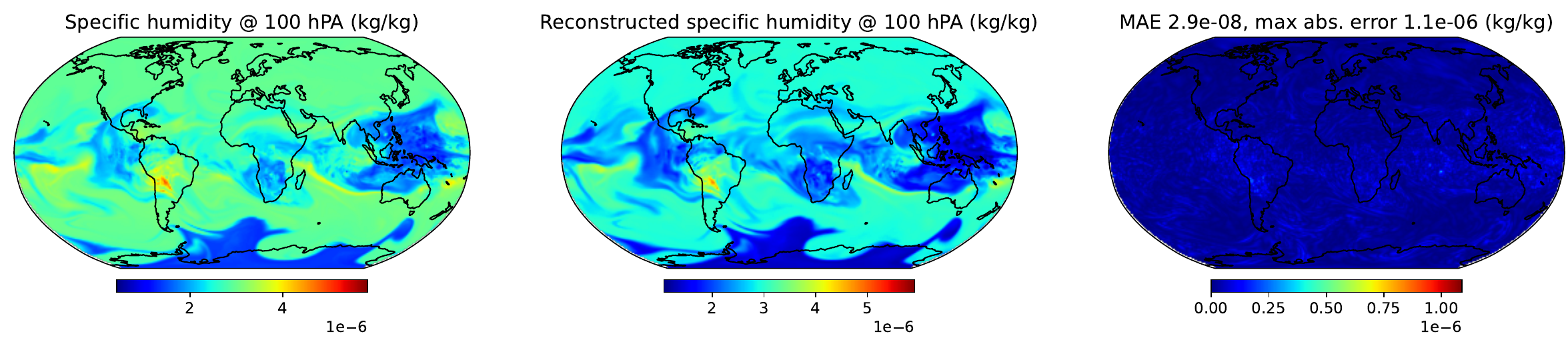}
    \includegraphics[width=\textwidth, trim = 6cm 3.5cm 5cm 3cm, clip]{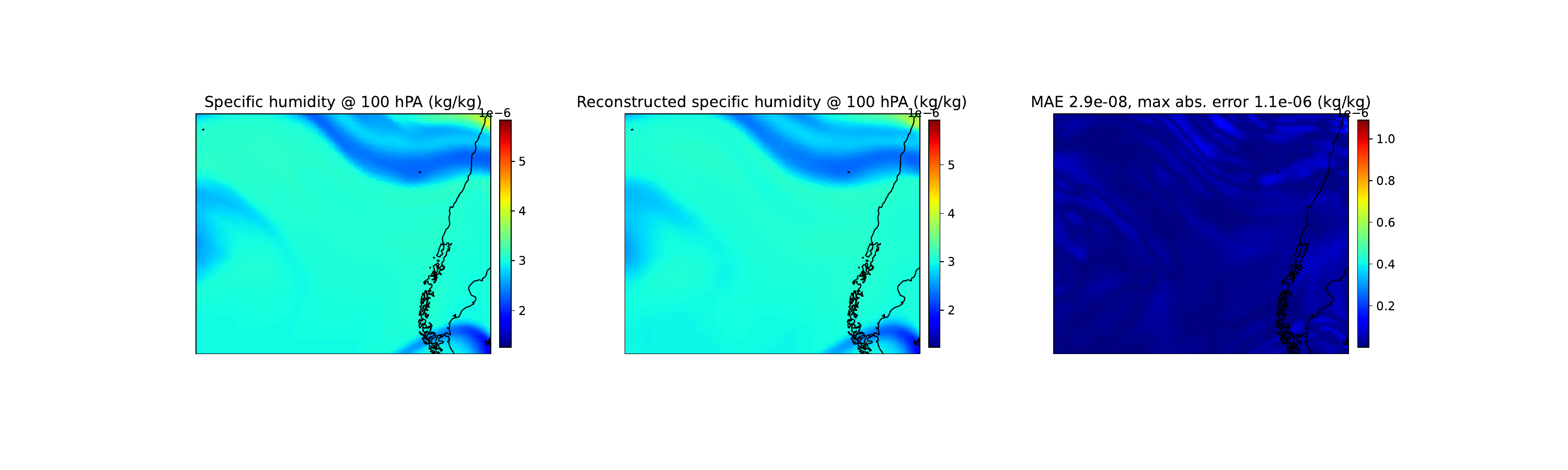}
    \hspace*{-2em}
    \caption{Analysis of the error maxima for hyperprior reconstructions on 2020/12/31 at 00:00 UTC. Top 2 rows show plots of specific humidity at 50hPa (global plot and zoom over the area with a large artefact of low specific humidity, over the southern Pacific near the coast of Chile). Middle 2 rows show plots of temperature at 50 hPa (global and zoom over the artefact area and reconstruction errors). Bottom two rows show reconstructions, without artefacts, of specific humidity at 100 hPa.}
    \label{fig:max_err_specific_humidity_hyperprior}
\end{figure}

\begin{figure}
    \centering
    \includegraphics[width=\textwidth, trim = 6cm 2.0cm 5cm 4.5cm, clip]{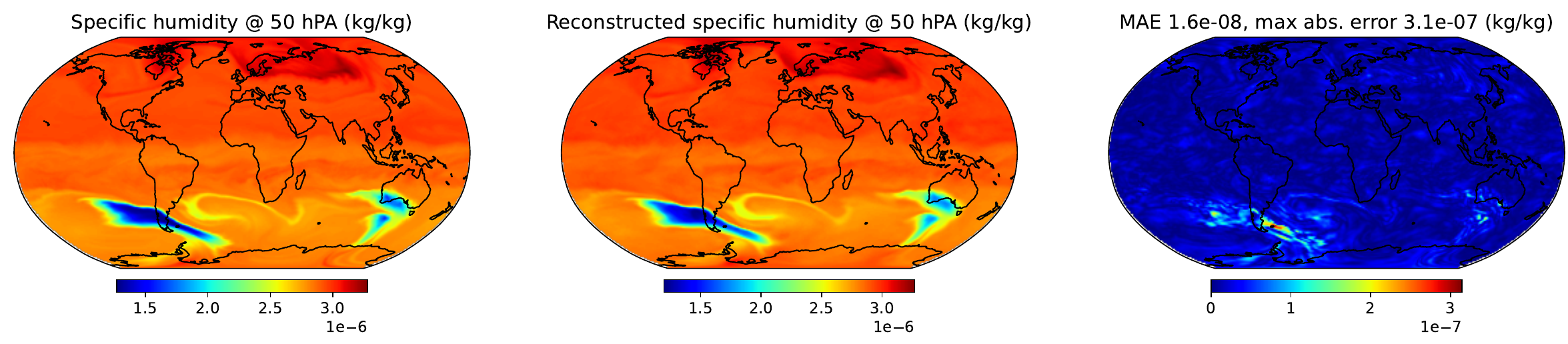}
    \includegraphics[width=\textwidth, trim = 6cm 3.5cm 5cm 3.0cm, clip]{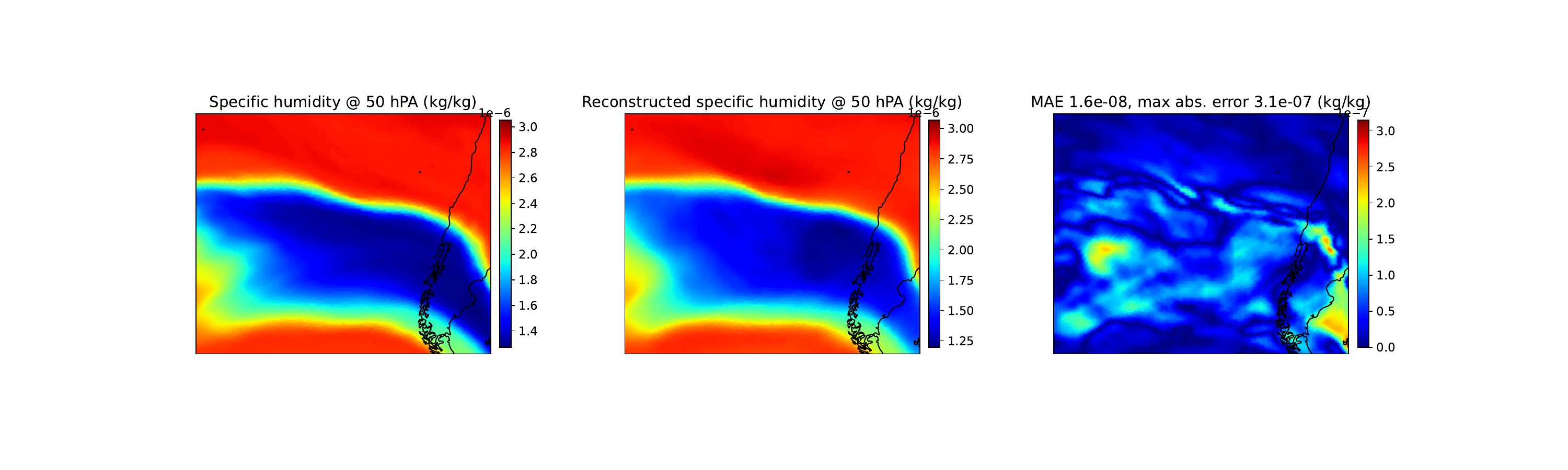}
    \includegraphics[width=\textwidth, trim = 6cm 2.0cm 5cm 4.5cm, clip]{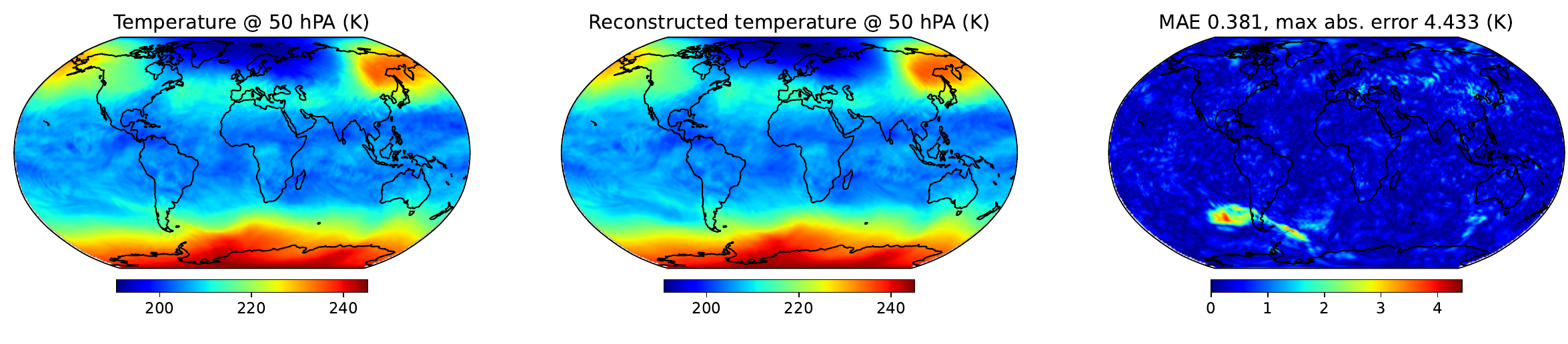}
    \includegraphics[width=\textwidth, trim = 6cm 3.5cm 5cm 3.0cm, clip]{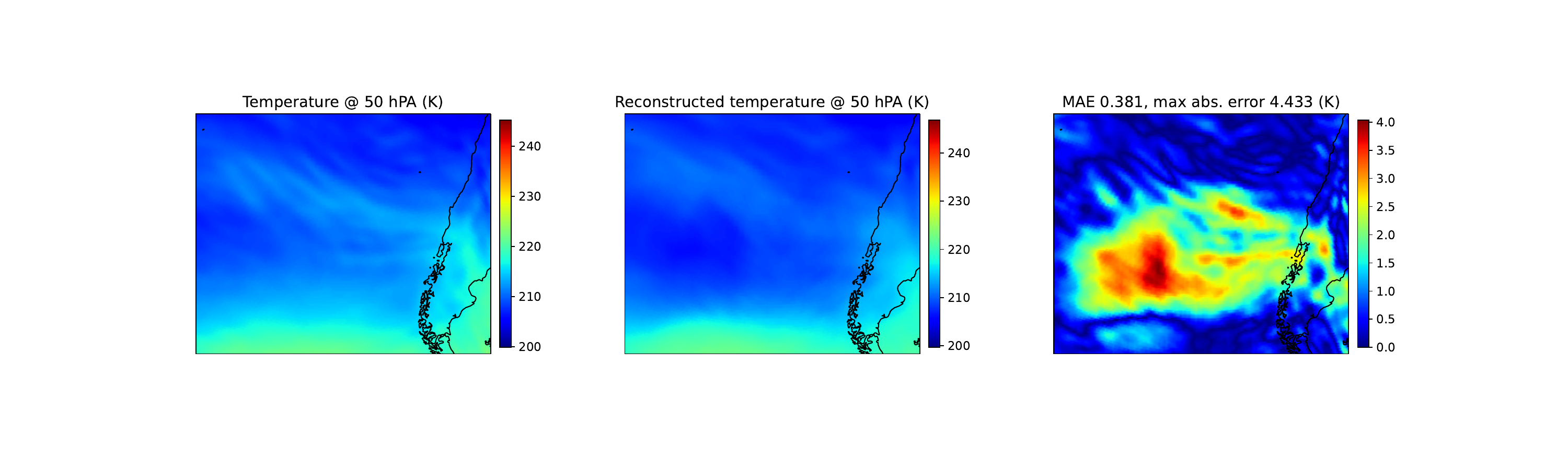}
    \hfill
    \caption{Analysis of the 3-block VQ-VAE reconstructions of 2020/12/31 at 0 UTC data (same as on Fig. \ref{fig:max_err_specific_humidity_hyperprior}). Top two rows show plots of specific humidity at 50hPa, bottom two rows show plots of temperature at 50 hPa. We show both global plots and plots localised over the southern Pacific near the coast of Chile, highlighting how the artefacts in the specific humidity are ``well'' reconstructed by VQ-VAE, while there are no major reconstruction artefacts of temperature ($\text{MAE}=0.381^\circ$~K).}
    \label{fig:max_err_specific_humidity_vqvae}
\end{figure}

We sort the reconstruction errors by value in order to investigate which timestamps and levels present the highest reconstruction errors, and select 31 December 2020 at 0 UTC as one of the worst reconstructions. We notice that at that timestamp, specific humidity at 50 hPa presents something that looks like low-value anomalies over the southern hemisphere (top two rows, left column of Figs. \ref{fig:max_err_specific_humidity_hyperprior} and \ref{fig:max_err_specific_humidity_vqvae}). Elsewhere in the literature it has been confirmed that ERA5 presents records with negative specific humidity\footnote{\url{https://forum.ecmwf.int/t/era5-lat-lon-pressure-negative-specific-humidity/2028}} and the GraphCast model evaluated on ERA5 performed worse at the 50 hPa level \citep{lam2023graphcast}. The hyperprior model does not manage to reproduce that low-value anomaly (see top two rows, center and right column of Fig. \ref{fig:max_err_specific_humidity_hyperprior}), generating instead high value artefacts, while the VQ-VAE can (see top two rows, centre and right columns of Fig. \ref{fig:max_err_specific_humidity_vqvae}). Furthermore, we see that the hyperprior fails to reconstruct temperature at that same timestamp of 31 December 2020 at 0 UTC and at pressure level 50 hPa (see how the middle two rows of Fig. \ref{fig:max_err_specific_humidity_hyperprior} show a localised reconstruction artefact appearing at the same location as the reconstruction artefact that appeared in reconstructions of specific humidity), whereas the VQ-VAE manages to reconstruct these data (see equivalent rows on Fig. \ref{fig:max_err_specific_humidity_vqvae}). 

Finally, the bottom two rows of Figure \ref{fig:max_err_specific_humidity_hyperprior} show that for the same timestamp (31 December 2020 at 0 UTC), but a different pressure level (100 hPa), there is no low-value anomaly in specific humidity, and the weighted MAE is lower ($2.9 \times 10^{-8}$ vs. $7.4 \times 10^{-8}$).

\begin{figure}
    \centering
    \includegraphics[width=\textwidth, trim = 6cm 2.0cm 5cm 4.5cm, clip]{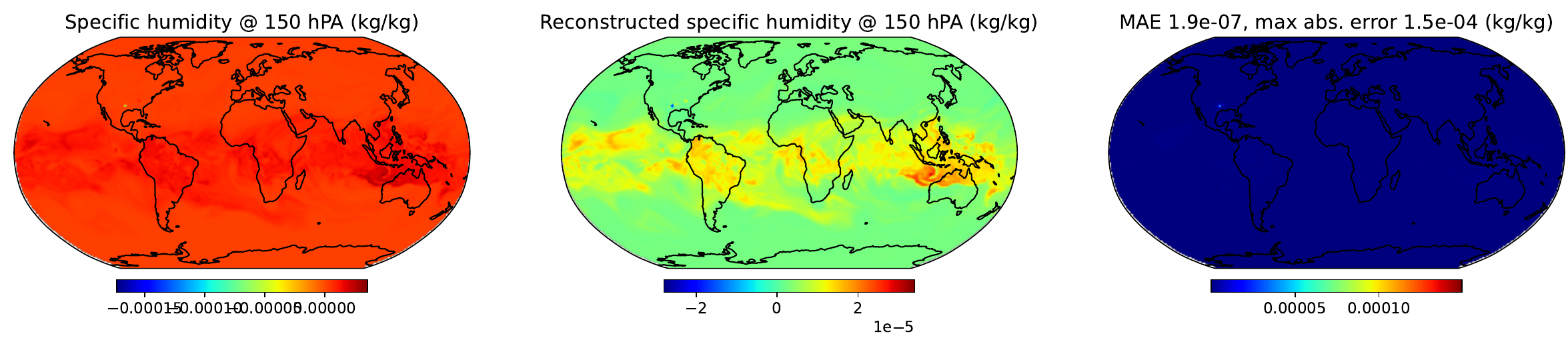}
    \includegraphics[width=\textwidth, trim = 6cm 3.5cm 5cm 3.0cm, clip]{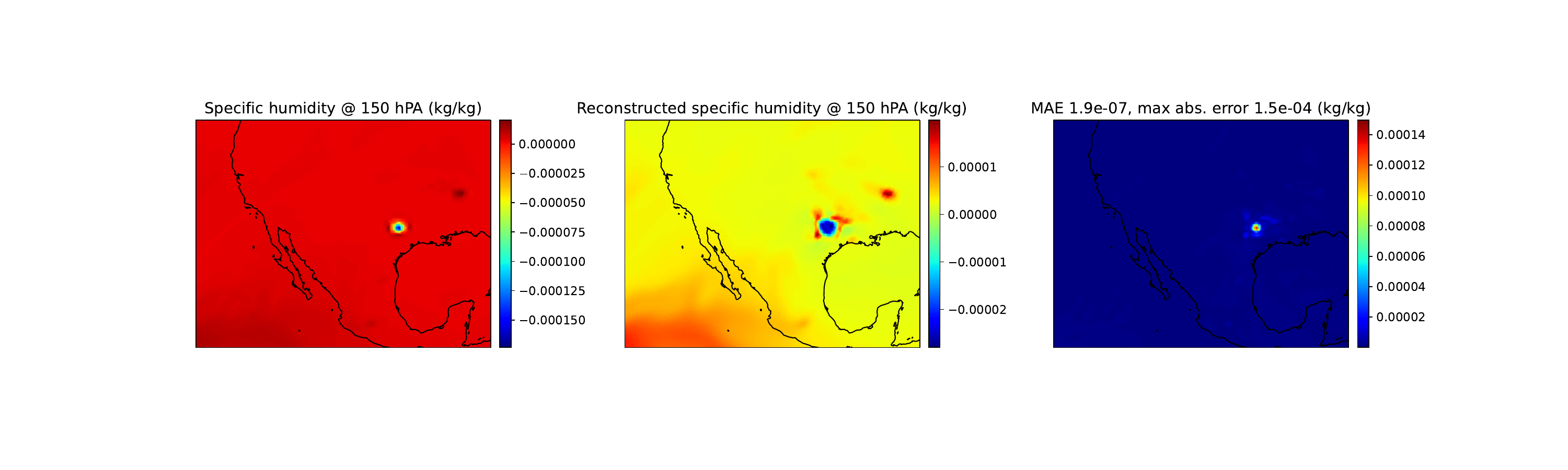}
    \includegraphics[width=\textwidth, trim = 6cm 3.5cm 5cm 3.0cm, clip]{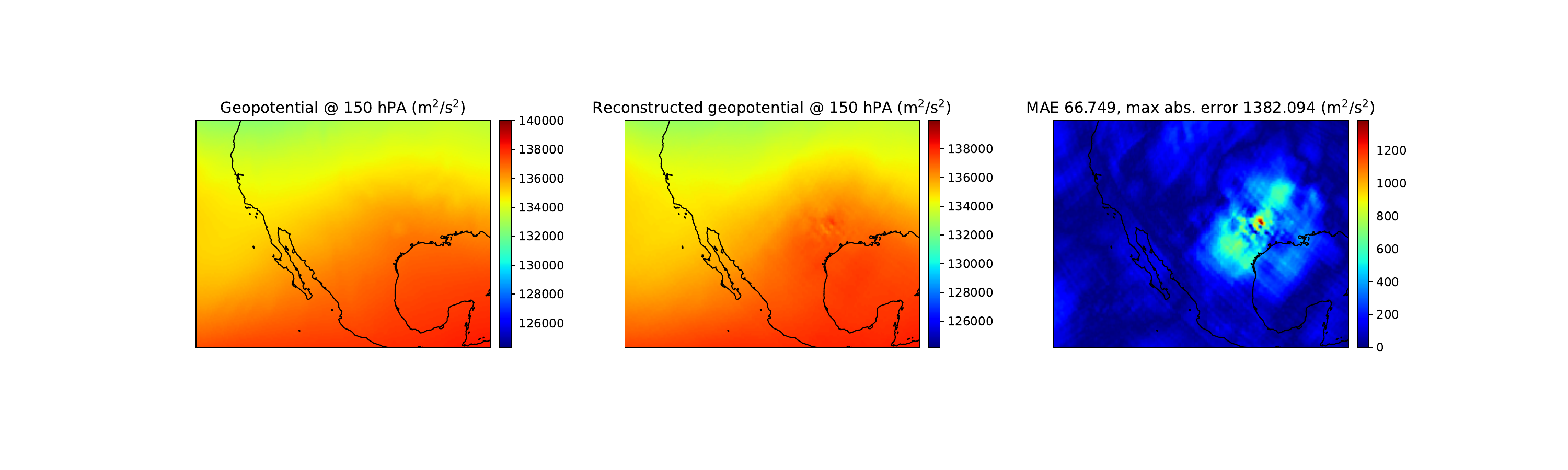}
    \includegraphics[width=\textwidth, trim = 6cm 2.0cm 5cm 4.5cm, clip]{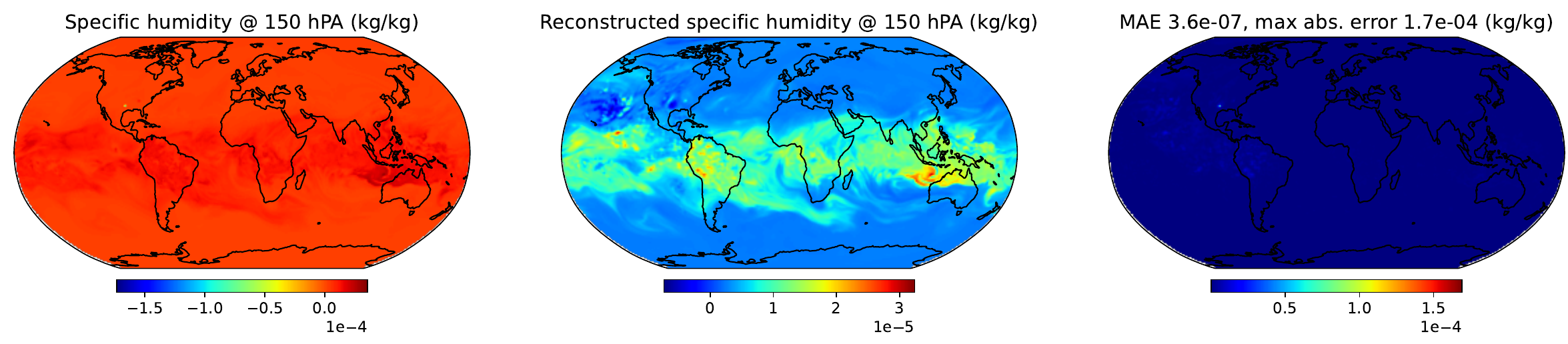}
    \includegraphics[width=\textwidth, trim = 6cm 3.5cm 5cm 3.0cm, clip]{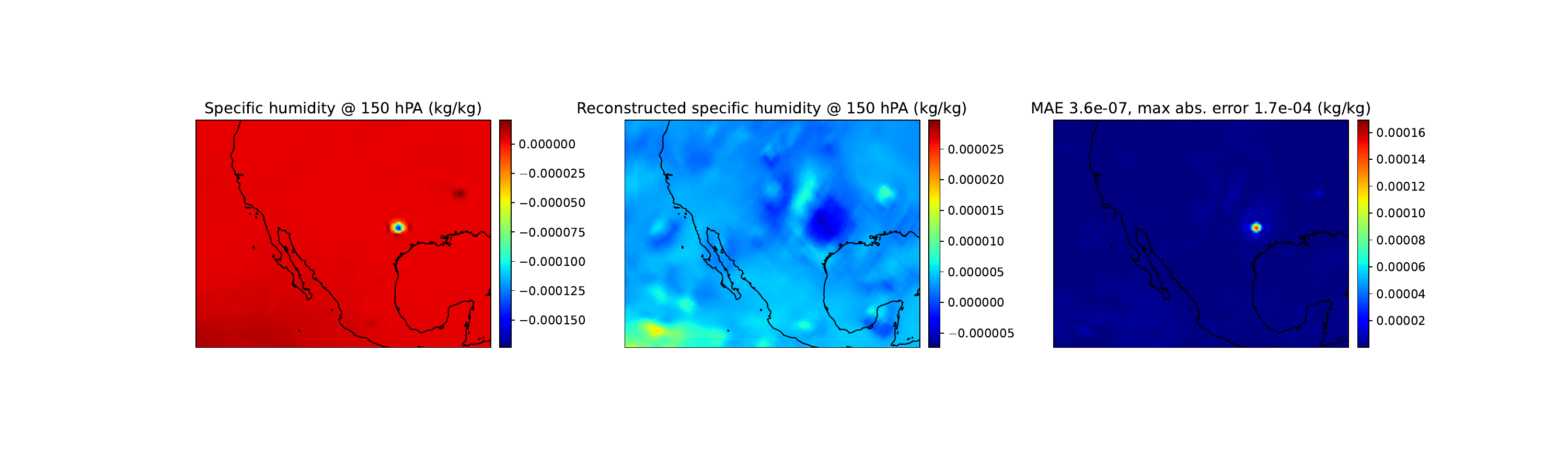}
    \includegraphics[width=\textwidth, trim = 6cm 3.5cm 5cm 3.0cm, clip]{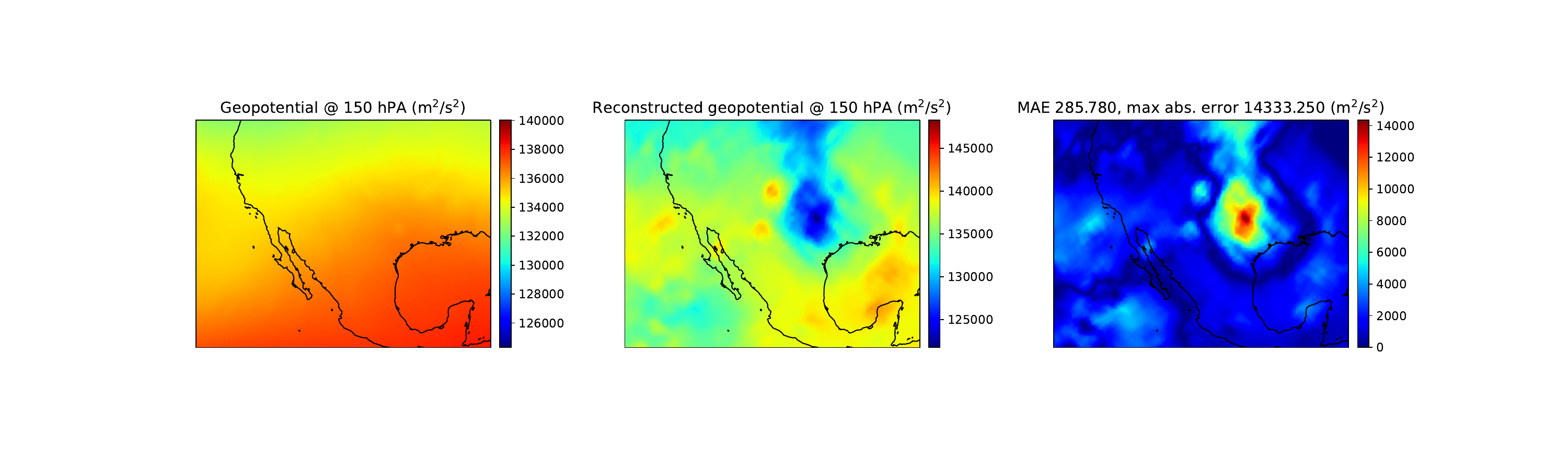}
    \hfill
    \caption{Reconstructions of data from 2012/3/14 at 22 UTC with an artefact in specific humidity at 150 hPa, localised over Texas. Top three rows show reconstructions using a hyperprior model, bottom three rows show reconstructions using 3-block VQ-VAEs.}
    \label{fig:max_err_specific_humidity_alt}
\end{figure}

Data artefacts or anomalies in the ERA5 dataset do not appear solely at 50 hPa. In a previous analysis of another model, we identified another data artefact appearing on specific humidity on 14 March 2012 at 22 UTC and at the 150 hPa pressure level, namely a localised blob over Texas with negative specific humidity values. Figure \ref{fig:max_err_specific_humidity_alt} demonstrates that this time around, it is the 3-block VQ-VAE model that suffers from worse reconstruction than the hyperprior model, and that the hyperprior manages to reconstruct another correlated variable at that same timestamp and pressure level: geopotential (with $MAE = 66.7^\circ$~K for hyperprior, as opposed to 285.8 for VQ-VAE).

\end{document}